\documentclass{article}

\usepackage{arxiv}

\usepackage[utf8]{inputenc} 
\usepackage[T1]{fontenc}    
\usepackage{hyperref}       
\usepackage{url}            
\usepackage{booktabs}       
\usepackage{amsfonts}       
\usepackage{nicefrac}       
\usepackage{microtype}      
\usepackage{lipsum}
\usepackage{algorithmic}
\usepackage{algorithm}
\usepackage{hyperref}
\hypersetup{hidelinks}
\usepackage{multirow}
\usepackage{tikz}
\usepackage{subfigure}
\usepackage{amssymb}
\usepackage{amsmath}
\usepackage{amsthm}
\title{OD-SGD: One-step Delay Stochastic Gradient Descent for Distributed Training}

\author{
  Yemao Xu\\
  Department of Computer Science\\
  National University of Defense Technology\\
  \texttt{xuyemaovip@nudt.edu.cn} \\
   \And
 Dezun Dong\thanks{Corresponding Author} \\
  Department of Computer Science\\
  National University of Defense Technology\\
  \texttt{dong@nudt.edu.cn} \\
  \And
  Weixia Xu \\
  Department of Computer Science\\
  National University of Defense Technology\\
  \texttt{xuweixia@nudt.edu.cn} \\
  \And
  Xiangke Liao \\
  Department of Computer Science\\
  National University of Defense Technology\\
  \texttt{xkliao@nudt.edu.cn} \\
}

\begin{document}
\maketitle

\begin{abstract}
The training of modern deep learning neural network calls for large amounts of computation, which is often provided by GPUs or other specific accelerators. To scale out to achieve faster training speed, two update algorithms are mainly applied in the distributed training process, i.e. the Synchronous SGD algorithm (SSGD) and Asynchronous SGD algorithm (ASGD). SSGD obtains good convergence point while the training speed is slowed down by the synchronous barrier. ASGD has faster training speed but the convergence point is lower when compared to SSGD. 
To sufficiently utilize the advantages of SSGD and ASGD, we propose a novel technology named One-step Delay SGD (OD-SGD) to combine their strengths in the training process. Therefore, we can achieve similar convergence point and training speed as SSGD and ASGD separately. 

To the best of our knowledge, we make the first attempt to combine the features of SSGD and ASGD to improve distributed training performance. Each iteration of OD-SGD contains a global update in the parameter server node and local updates in the worker nodes, the local update is introduced to update and compensate the delayed local weights. We evaluate our proposed algorithm on MNIST, CIFAR-10 and ImageNet datasets. Experimental results show that OD-SGD can obtain similar or even slightly better accuracy than SSGD, while its training speed is much faster, which even exceeds the training speed of ASGD. 

\end{abstract}


\section{Introduction}

Deep learning has recently experienced success on various areas, 
ranging from computer vision to natural language processing and 
so on. Many prominent results in these areas are achieved through 
the utilization of deep learning networks \cite{mpcasgd}. The resurgence 
of deep learning mainly results from the appearance of general and specific 
hardware accelerators (GPU \cite{GPU}, NPU \cite{chen2016diannao},  
TPU \cite{jouppi2017datacenter} etc. ), which brings 
about tremendous advances in computing power, and the public 
availability of versatile training datasets like ImageNet \cite{deng2009imagenet} 
and CIFAR \cite{CIFAR-10} 
also contributes its development. However, with the fast-growing sizes 
of deep neural networks and datasets, the computing capability for 
training makes the bottleneck, it requires days or  weeks to complete 
the training (For instance, the training of ResNet-50 \cite{he2016deep} for 90 epochs 
takes 21 hours with 8 P100GPUs), which makes it impossible for real-time 
interaction in the development process. Under this circumstance, distributed 
training becomes the prevailing practice and it greatly improves the productivity 
of training deeper and larger neural networks \cite{chilimbi2014project, xing2015petuum, moritz2015sparknet, zinkevich2010parallelized}. 

Stochastic Gradient Descent (SGD) is a widely used optimization algorithm 
for distributed training. For the training data of the same size, the computation 
time used in the forward-backward process can be reduced dramatically by 
increasing the training nodes and making use of data parallelism \cite{lin2017deep}. 
One can choose the synchronous SGD algorithm (SSGD) or asynchronous SGD algorithm (ASGD) in the training phase. For SSGD, devices in each local worker will figure 
out the gradients firstly with their own mini-batches of data, and then the 
gradients are added to the global models. SSGD can obtain good convergence speed 
while its training speed deteriorates due to the synchronous overhead. 
To circumvent the shortcoming of SSGD, practitioners have also resorted to ASGD, which 
focus on training speed and uses the staled information for computation. Although
ASGD achieves better training performance by alleviating the waiting 
overhead in SSGD, it often suffers from poorer convergence accuracy 
because of the delayed gradients \cite{chen2016revisiting}. 

\begin{figure}[t]
	\centering
	\subfigure[SSGD training mode]{
		\label{fig:sgd:a} 
		\includegraphics[width=6cm, height=3.2cm]{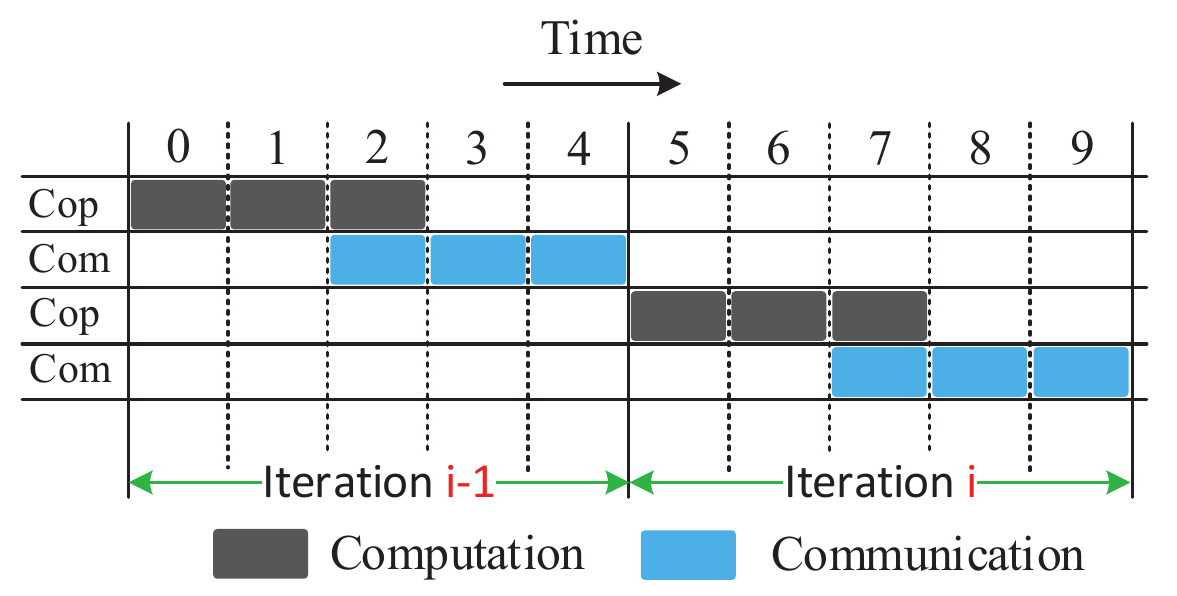}}
	\hspace{0.5in} 
	\subfigure[OD-SGD training mode]{
		\label{fig:sgd:b} 
		\includegraphics[width=6cm, height=3.2cm]{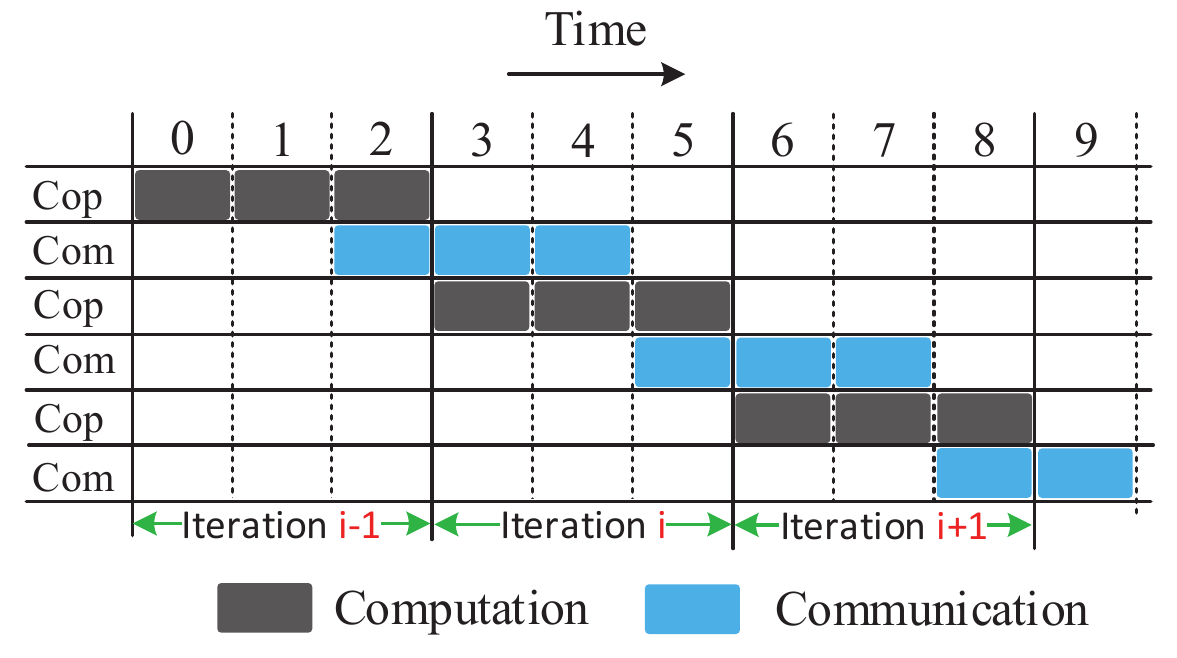}}
	\caption{Mechanisms of SSGD and OD-SGD. Supposing the training and communication overheads are 3 time units respectively. When training with SSGD, the next iteration task will not start until the communication process of current iteration ends, and the overhead per iteration is 5 time units. When training with OD-SGD, the next training task can be launched when the computation process of current iteration ends, and the overhead for one iteration is 3 time units.}
	\label{fig:sgd} 
\end{figure}

Based on the conclusions that SSGD achieves faster convergence 
speed while ASGD experiences faster training speed, the combination 
of these features would probably lead to better distributed training performance. 
Therefore, we propose a novel method named One-step Delay SGD algorithm (OD-SGD) to 
take advantage of the merits in both sides. The intrinsical motivation 
of this strategy is to start the next iteration training as soon as possible 
when current iteration ends. As illustrated in Fig.~\ref{fig:sgd}, the computation and communication processes take three time units separately. The overhead for one iteration under SSGD is five time units. By launching the next training task immediately when the current iteration ends, the overhead under OD-SGD takes only three time units for one iteration.

OD-SGD is similar to SSGD in the sense that to ensure the 
convergence speed and training result, the synchronous mechanism is still 
contained. It differs from SSGD in that no workers need to wait for others, 
the weights used for the next iteration are delivered while the current
iteration task is running. The similarity between OD-SGD and ASGD is that 
each local worker will update the pulled weights with their local gradients. 
From Fig.~\ref{fig:sgd} we can notice OD-SGD has a higher overlap ratio, the communication 
overhead is overlapped by the computation overheads of two adjacent iterations. 
This new proposed method includes three main stages, namely the warm-up 
stage, the switching stage and the training stage with one-step delay mechanism. 
The warm-up stage here means training with SSGD in the beginning, which is different from that mentioned in \cite{goyal2017accurate}. To facilitate the distinction between these two warm-up operations, the warm-up in OD-SGD is marked as warm-up while the latter is marked as WP. Warm-up stage is introduced in OD-SGD to make the global weights stable before the application of one-step delay mechanism, thus achieving a better convergence point. The switching stage is the last two iterations in the warm-up stage. In the penultimate iteration, the local workers will make a copy of the pulled weights. In the last iteration, each local worker will update the copied weights with the newly calculated gradients and launch the next training task immediately. In the meanwhile, the gradients are transmitted to the parameter servers. In the third stage, the parameter servers will update weights with the received gradients under synchronous mechanism, 
while the local workers will update the copied weights and deliver the gradients. It is noteworthy that OD-SGD is applicable to both Peer-to-Peer and Parameter Server structures.

We implement OD-SGD in MXNet \cite{chen2015mxnet}, an efficient and flexible 
framework designed for neural network training. MXNet is parameter server 
structure based and one of the pivotal designs is the dependency engine, 
which is a library that takes a sequence of operations and schedules them 
according to the dependency pattern, potentially in parallel. It supports 
parallel executions of irrelevant operations in the training process. Although 
OD-SGD is implemented in MXNet, it is definitely applicable 
to end-to-end based learning platforms like Pytorch \cite{pytorch}
and Caffe \cite{jia2014caffe}.

We emphasize the obstacles that need to be tackled for the sake of implementing 
the OD-SGD in MXNet as follows. Firstly, how the computation and 
communication operations for training are scheduled in MXNet. Secondly, 
the communication mechanism for transmitting the gradients and updated 
parameters.Thirdly, how does the dependency engine solve the data flow 
dependency problem, or a detailed understanding of the dependency engine. 
Finally, what operations should be added to implement the OD-SGD.

In order to settle the challenges above, we dive into the training 
mechanism of MXNet and solidly implement the OD-SGD, experimental 
results show the effectiveness of our OD-SGD. This paper mainly makes the 
following contributions. In the first place, we briefly introduce the essential training 
mechanism of MXNet, emphasizing on communication pattern and 
dependency engine. Secondly, we design the OD-SGD and implement it on MXNet platform. 
Thirdly, mathematical analysis is made on why OD-SGD works and its convergence property.
Fourthly, experimental results prove that OD-SGD delivers similar convergence accuracy as SSGD, or even slightly better, while it achieves faster training speed than ASGD. Finally, we make some analyses  about warm-up sensitivity and the performance improvement under different cluster scales and bandwidths.

The rest of this paper is structured as follows. We show the background in Section 2, then introduce  the design and implementation of OD-SGD and provide mathematical proof in Section3 and Section 4. Experiment results and related work are presented in Section 5 and Section 6. Finally, we conclude with insights we have learned through designing such an algorithm in Section 7.

\section{Background}
\subsection{Data-Parallelism SGD and Model-Parallelism}
\begin{figure}
	\centering
	\begin{minipage}{0.48\textwidth}
		\centering
		\includegraphics[width=7cm, height=3.2cm]{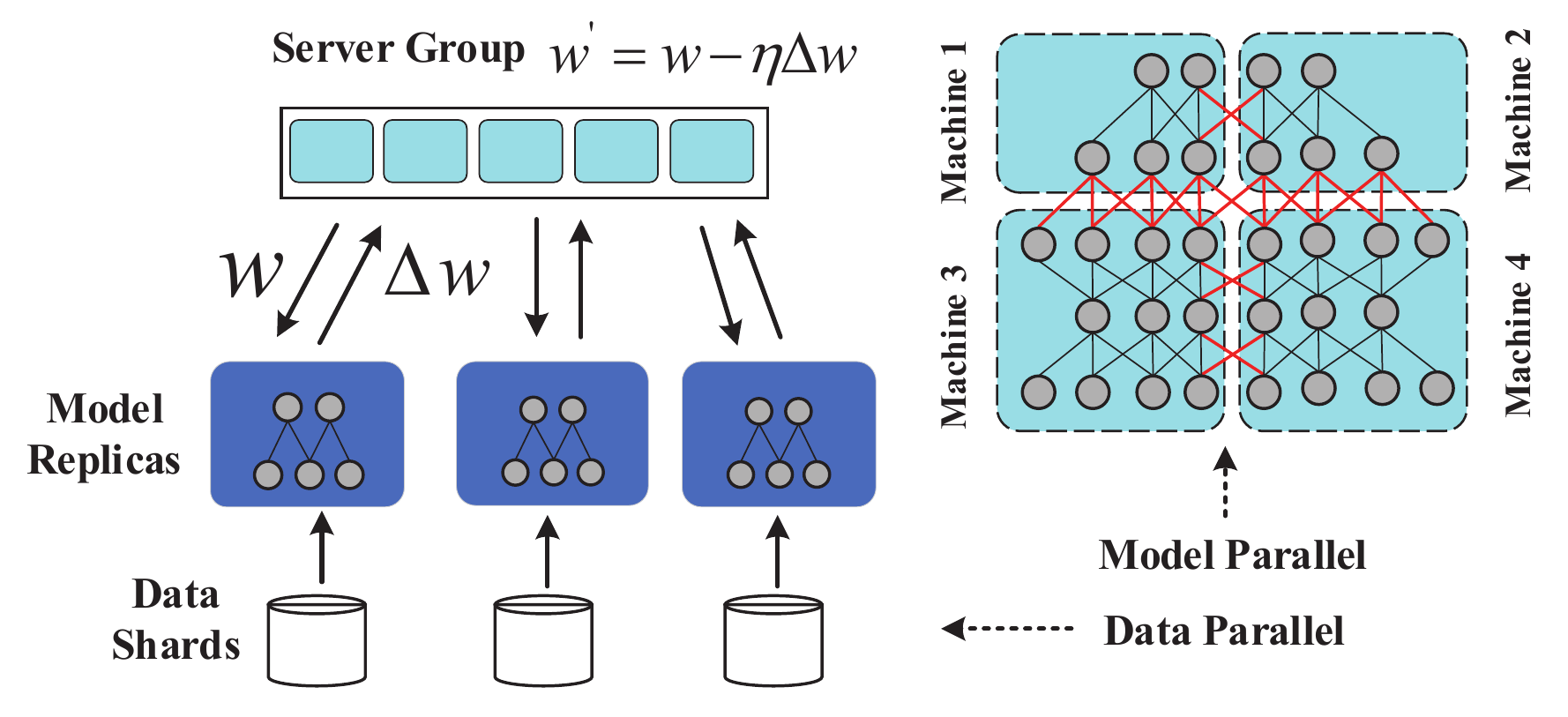}
		\caption{Data parallel and model parallel}
		\label{fig:data-model-p}
	\end{minipage}	
	\begin{minipage}{0.48\textwidth}
		\centering
		\includegraphics[width=7cm, height=3.2cm]{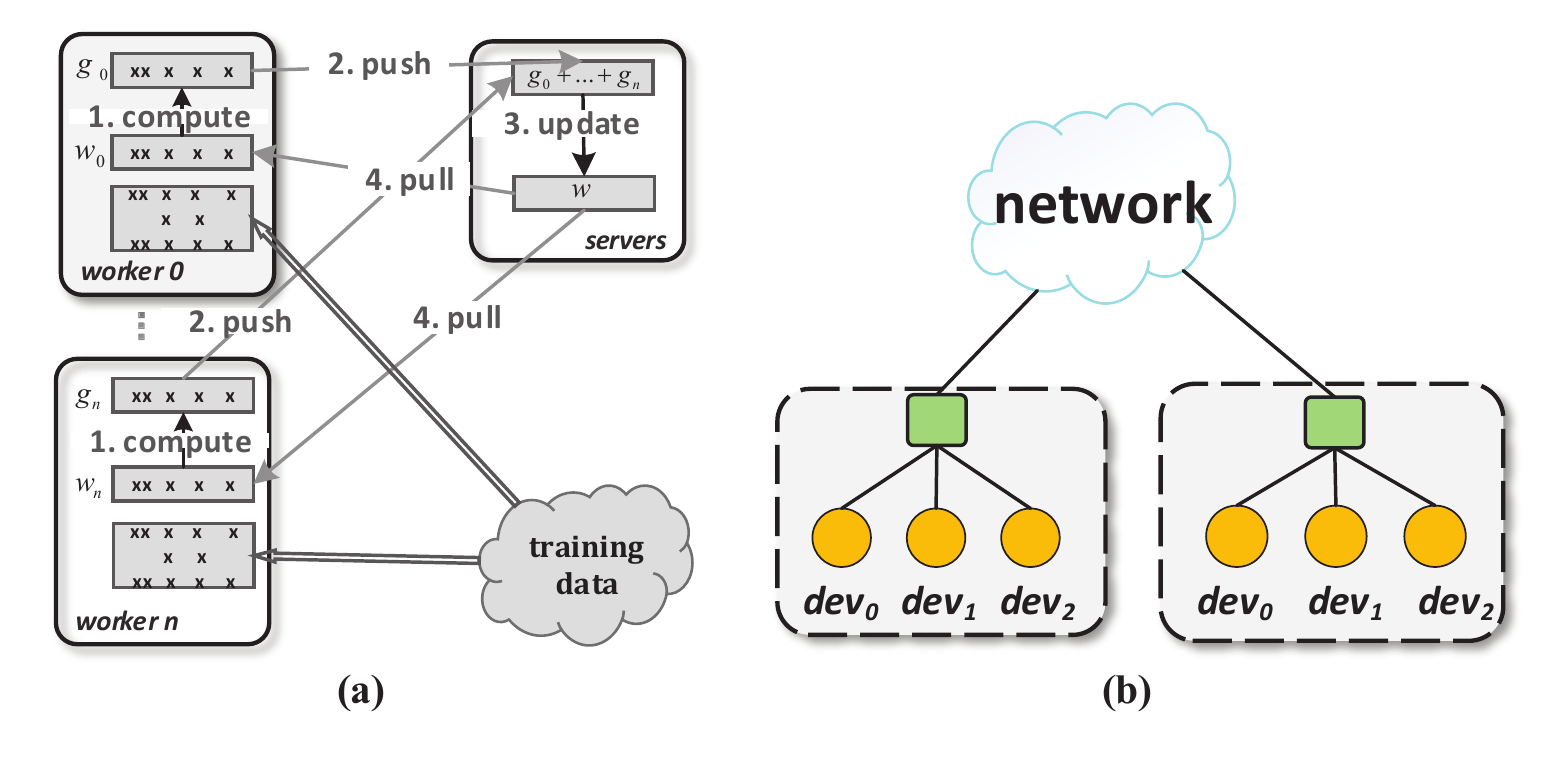}
		\caption{Communication mechanism of MXNet}
		\label{fig:mxnet-pp}
	\end{minipage}	
\end{figure}

Data parallelism method is widely used in distributed deep learning training, in which the dataset is partitioned according to the number of local workers. Supposing $\textbf{m}$ workers is involved, each worker is responsible for $1/M$ of the dataset in one epoch training task. In the data parallelism method, every worker makes a copy of the neural network and the weights ($\textbf{w}_{t}$). The training process contains three parts: computing and delivering the local gradients, updating the global weights maintained in the master node and pulling back the updated weights. The data parallelism strategy can be divided into two modes: synchronous mode and asynchronous mode, corresponding to SSGD and ASGD, whose advantages and disadvantages are discussed above. 
Different from data parallelism, model parallelism method divides the neural network into several parts. Nevertheless, because of the limited input size (e.g. size of an image), the matrix operations are not large, the partition of a neural network does not need many machines \cite{you2018imagenet} 
and the batch size for one iteration is limited because of the limited memory. Therefore, the data parallelism is often the choice of many state-of-art methods, Fig.~{\ref{fig:data-model-p}} provides examples of data and model parallelism.

\subsection{Communication Mechanism of MXNet}
To update the global weights, the parameter servers need to get gradients from all other workers. The communication process, which consists of two phases: 1) intra-node reduction, 2) inter-node point to point message passing, is completed  through \textbf{\textit{Push/Pull}} operations. In the first phase, gradients calculated by different devices of the local worker will be aggregated on one single 
device, as shown in Fig.~\ref{fig:mxnet-pp}{\color{red}{(b)}}. In the second phase, each local worker will transmit the aggregated gradient to the server node via \textbf{\textit{Push/Pull}} operations ( Fig.~\ref{fig:mxnet-pp}{\color{red}{(a)}}). 
Every parameter corresponds to one pair of \textit{<key, value>}, and the gradients 
of parameters are delivered one by one. Supposing a layer has two parameters, 
two pair of \textbf{\textit{Push/Pull}} operations will be launched.

\subsection{Dependency Engine in MXNet}

\begin{figure}[b]
	\centering
	\begin{minipage}{0.4\textwidth}
		\includegraphics[width=5.4cm, height=3.2cm]{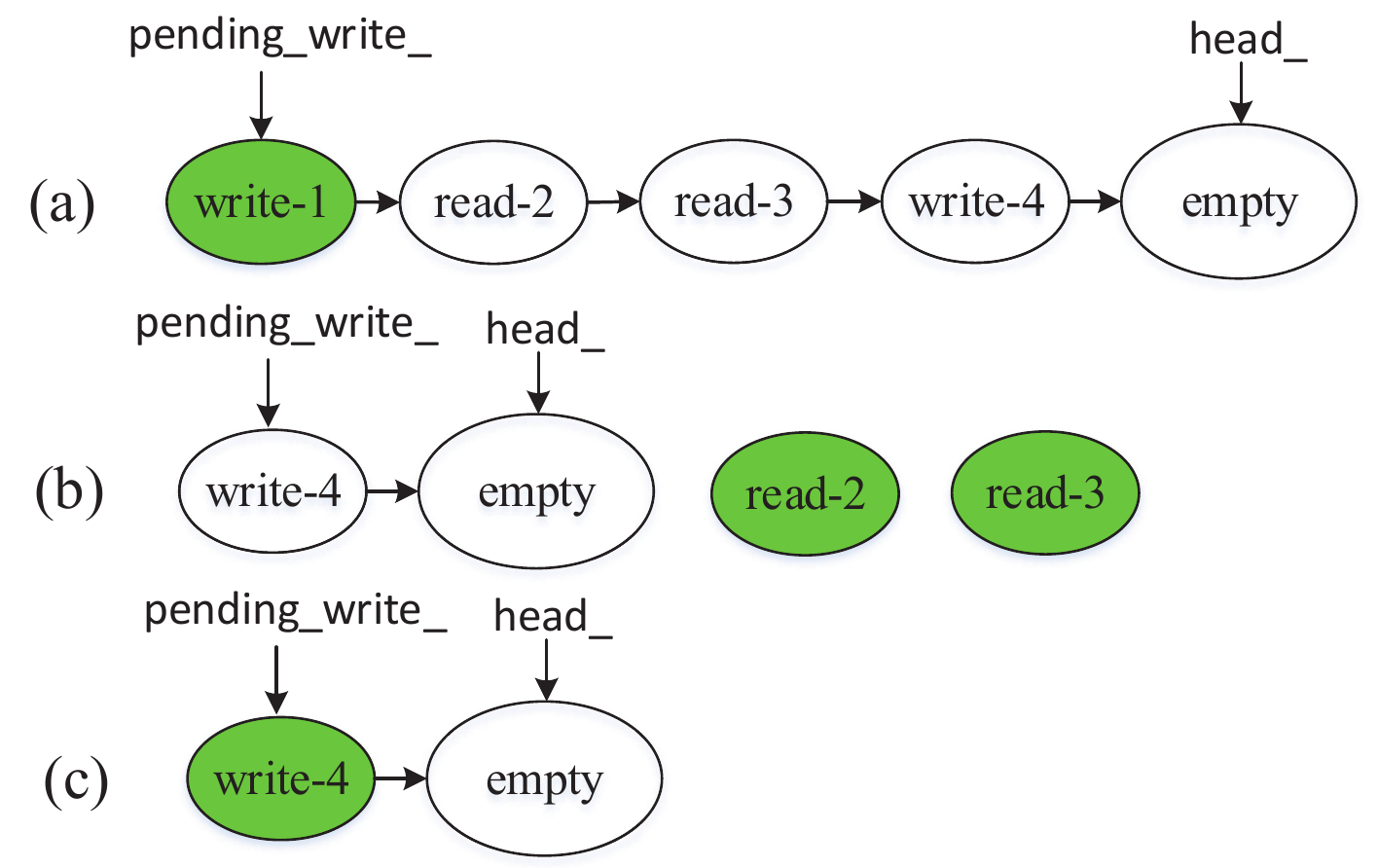}
		\caption{A maintained queue sample in dependency engine.}
		\label{fig:dp-sample} 
	\end{minipage}
	\begin{minipage}{0.56\textwidth}
		\includegraphics[width=8cm, height=3.5cm]{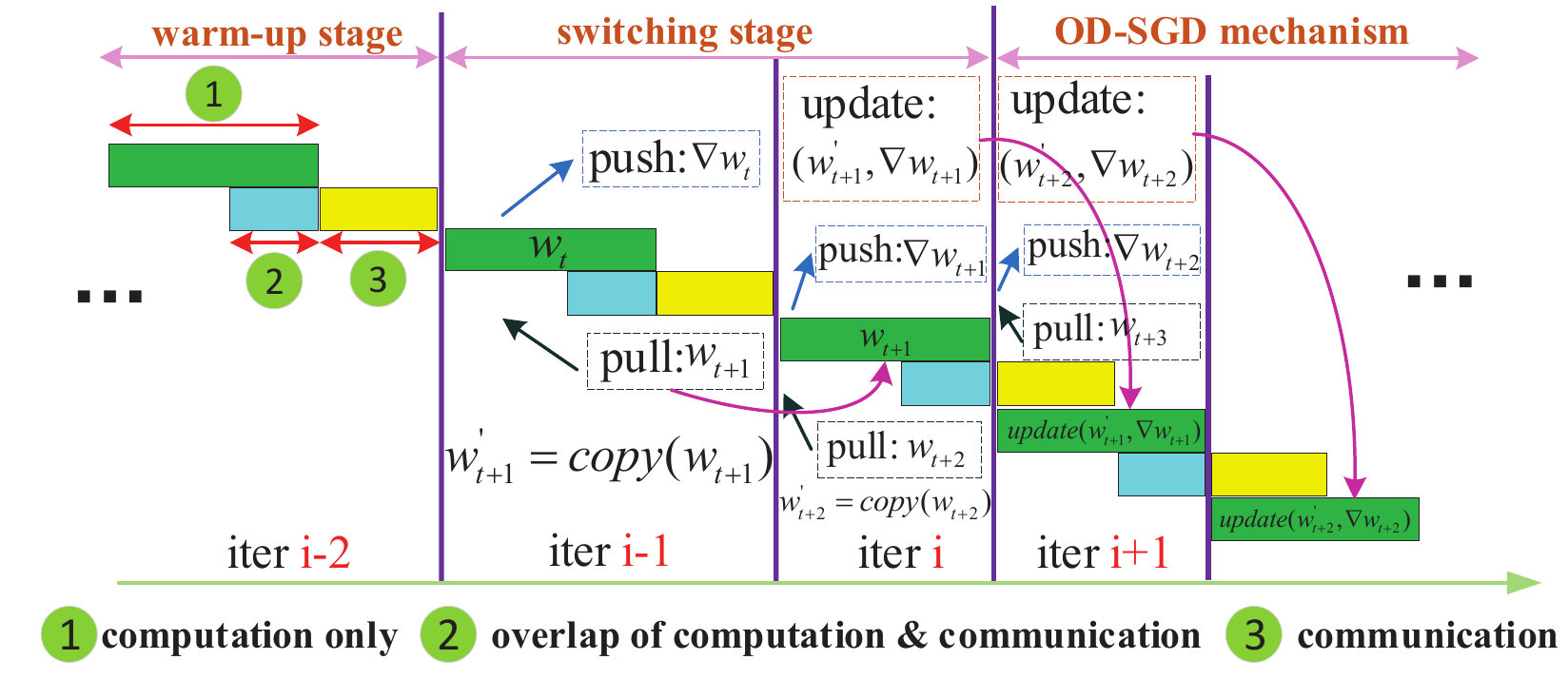}
		\caption{Overview of the three stages involved in OD-SGD. Weights ({\color{blue}{$w_{t+1}^{'}$}}) copied at iteration {\color{red}{\textit{i-1}}} is updated at iteration {\color{red}{\textit{i}}} ({\color{blue}{update: ($w_{t+1}^{'}$, $\triangledown{w_{t+1}}$)}}) and used for training at iteration {\color{red}{\textit{i+1}}} with the OD-SGD mechanism. Actually, the \textbf{pull} and \textbf{copy} operations in iteration {\color{red}{\textit{i}}} are conducted when the training task of iteration {\color{red}{\textit{i+1}}} is executing, they can be parallelized.}
		\label{fig:od-mech} 
	\end{minipage}
\end{figure}

A dependency engine is a library that schedules a series of operations according to their dependency relationships between each other. MXNet dependency engine is designed to make MXNet run faster and scale to larger datasets, with which we can parallelize computation and communication operations across devices. 
Both kinds of operations require reading or writing data from different variables. Each variable correspond to a maintained queue in the dependency engine and Fig.~\ref{fig:dp-sample} presents a sample which describes three different states of a queue. \textit{write-1}, \textit{read-2}, \textit{read-3} and \textit{write-4} are operations to access the same variable, independent operations (\textit{read-2} \& \textit{read-3}) can be parallelized to improve the execution efficiency. We come up with the OD-SGD by parallelizing more independent operations to improve the distributed training performance.

\section{One-Step Delay SGD: Algorithm Description and Implementation}

This design aims at improving the training performance of distributed deep learning systems, and accelerating the computation and communication pipeline in the training engine. The main idea is to increase the overlap ratio between computation and communication process. OD-SGD is divided into three primary stages, as shown in Fig.~\ref{fig:od-mech}, warm-up stage, switching stage, and training stage.

\begin{algorithm}[b] 
	\caption{OD-SGD: parameter server} 
	\label{alg:one-s} 
	\begin{algorithmic}[1] 
		\STATE \textbf{Initialize}: {$t$ = 0, $count$ = 0, $\textbf{w}_0$ is initialized with specified initializer, $m \in \{1, 2, ... , M\}$}
		\REPEAT
		\IF {receive $gradient_{m}^{t}$ from $worker_{m}$}
		\STATE $count \leftarrow count + 1$
		\IF{$count$ == $M$}
		\STATE Update $w_{t}$ with aggregated gradient $gradient_{t}$
		\STATE $t \leftarrow t + 1$
		\STATE $count  \leftarrow 0$
		\ENDIF
		\ELSIF{receive ''pull request'' from $\textbf{worker}_m$}
		\STATE Send $\textbf{w}_t$ back to $\textbf{worker}_m$
		\ENDIF
		\UNTIL{forever}		
	\end{algorithmic} 
\end{algorithm}

\subsection{Algorithm Description}
The flow of OD-SGD is shown in Algorithm~\ref{alg:one-s} and Algorithm~\ref{alg:one-w}. Here we implement OD-SGD by using the parameter server framework (it can be implemented in other frameworks). According to Algorithm~\ref{alg:one-s}, the work process of parameter 
server keeps unchanged when compared to SSGD. It receives gradients from 
all other workers and update the global weights using the synchronous mechanism. Then it 
sends the updated weights back to corresponding worker on receiving its \textit{pull request}. 
Algorithm~\ref{alg:one-w} illustrates the work process of each worker node. The warm-up 
phase of OD-SGD differs from that mentioned in \cite{goyal2017accurate}, which is increasing the learning rate from a relative small value to the target value. Warm-up stage here means the worker 
conducts training task under the SSGD in the beginning. It broadcasts the pulled weight to every local device to compute the gradient, then aggregate the gradients from local devices and push the aggregated one to the parameter server. In the second stage, the worker makes a copy of the pulled weight ($\textbf{w}^{t}_{bak}$) when the iteration number is (\textit{wp - 1}) and updates it in the next iteration. In the third stage, the worker directly broadcasts  $\textbf{w}^{t-1}_{bak}$ to its local devices without waiting for the completion of \textbf{\textit{Pull}} operation in the last iteration, because there is no data dependency between these two operations. For instance, the last \textbf{\textit{Pull}} operation in the \textit{While} loop and the first \textit{Broadcast} in the \textbf{repeat} part can be parallelized, so are the following adjacent \textbf{\textit{Pull}} and \textbf{Broadcast} operations in the \textbf{repeat} part. The parallel execution of \textbf{\textit{Pull}} and \textbf{Broadcast} is the key to increasing the overlap ratio, making it possible to launch the next iteration task as soon as the current iteration completes. 

It is noteworthy that OD-SGD combines the features of SSGD and ASGD. 
OD-SGD and SSGD are similar 
in that the parameter server works under synchronous update mechanism, 
and OD-SGD and ASGD are similar in that local workers can start the next iteration 
tasks quickly without waiting for the synchronous communication process. 
When compared to SSGD in implementation, no modification is required in the 
parameter server side. For the local worker side, OD-SGD calls for a copy of 
the local weights, which brings about extra memory usage. 
Therefore, we store the 
backup model weights ($\textbf{w}^{t}_{bak}$) in the CPU-side memory and will not 
occupy the memory of devices (GPUs, TPUs etc.) used for computation task. 
In addition, OD-SGD has extra computational requirement on the local workers to 
update $\textbf{w}^{t}_{bak}$ with newly calculated gradients. However, the additional computation only introduces a lightweight overhead and it can be partly overlapped by the 
\textbf{\textit{Push}} operation. Because both the \textbf{\textit{Push}} operation and the \textbf{\textit{local\_update}} operation have only \textit{read dependency} on the newly calculated gradients, and they can be parallelized according to Fig.~\ref{fig:dp-sample}. 
For the algorithm for \textbf{\textit{local\_update}}, we can utilize the existing SGD-based algorithms or custom algorithms defined by ourselves. In our experiments, we adopt the SGD, DC-ASGD-c and DC-ASGD-a \cite{dcasgd-2017} algorithms for different neural network and cluster scales.

\begin{algorithm}[t] 
	\caption{OD-SGD: worker \textit{m}} 
	\label{alg:one-w} 
	\begin{algorithmic}[1] 
		\STATE \textbf{Input:} {warm-up iteration number $wp$}
		\STATE \textbf{Initialize}: {Pull initialized $w_0$ from parameter server, $num$ = 0, $t$ = 0}
		\WHILE{$num$ <= $wp$}
		\STATE Broadcast $\textbf{w}_t$ to local devices ($\textbf{dev}^{i}_{loc}$)
		\STATE Computer gradient $\textbf{g}_m$
		\STATE $t \leftarrow t + 1$
		\STATE Push $\textbf{g}_m$ to the parameter server
		\IF{$num$ == $wp$} 
		\STATE Update $\textbf{w}^{t-1}_{bak}$ with $\textbf{g}_m$
		\ENDIF
		\STATE Pull $\textbf{w}_t$ from the parameter server
		\IF{$num$ == $wp - 1$}
		\STATE $\textbf{w}^{t}_{bak}$ = $\textbf{w}_t$
		\ENDIF
		\STATE $num \leftarrow num + 1$
		\ENDWHILE	
		\REPEAT
		\STATE Broadcast $\textbf{w}^{t-1}_{bak}$ to local devices ($\textbf{dev}^{i}_{loc}$)
		\STATE $\textbf{w}^{t}_{bak}$ = $\textbf{w}_t$
		\STATE Computer gradient $\textbf{g}_m$
		\STATE Push $\textbf{g}_m$ to the parameter server
		\STATE Update $\textbf{w}^{t}_{bak}$ with $\textbf{g}_m$
		\STATE $t \leftarrow t + 1$
		\STATE Pull $\textbf{w}_t$ from the parameter server
		\UNTIL{forever}		
	\end{algorithmic} 
\end{algorithm}

\begin{figure}[t]
	\centering
	\subfigure[$T_{com}$ < $T_{cop}$]{
		\label{fig:od-pberformance:a} 
		\includegraphics[width=5cm, height=2.3cm]{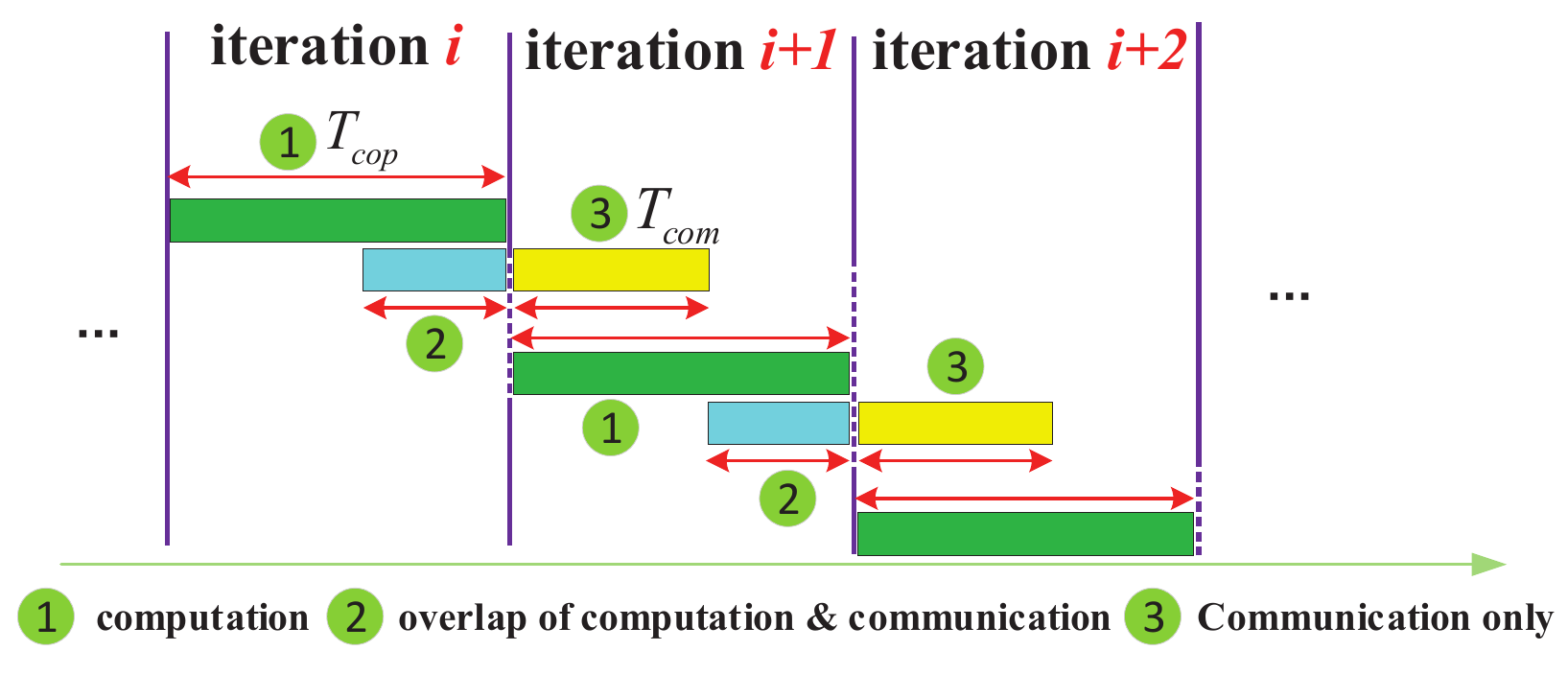}}
	\subfigure[$T_{com}$ = $T_{cop}$]{
		\label{fig:od-performance:b} 
		\includegraphics[width=5cm, height=2.3cm]{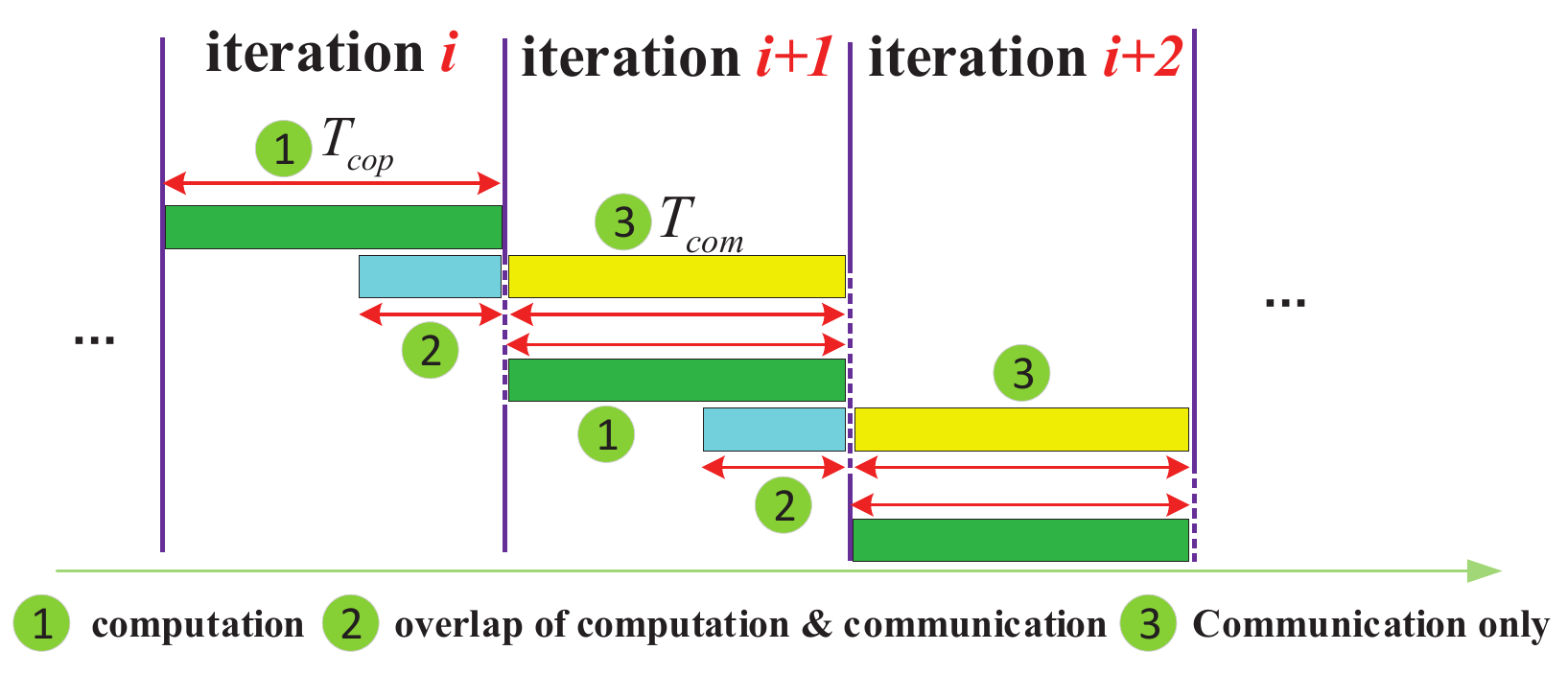}}
	\subfigure[$T_{cop}$ < $T_{com}$]{
		\label{fig:od-performance:c} 
		\includegraphics[width=5cm, height=2.3cm]{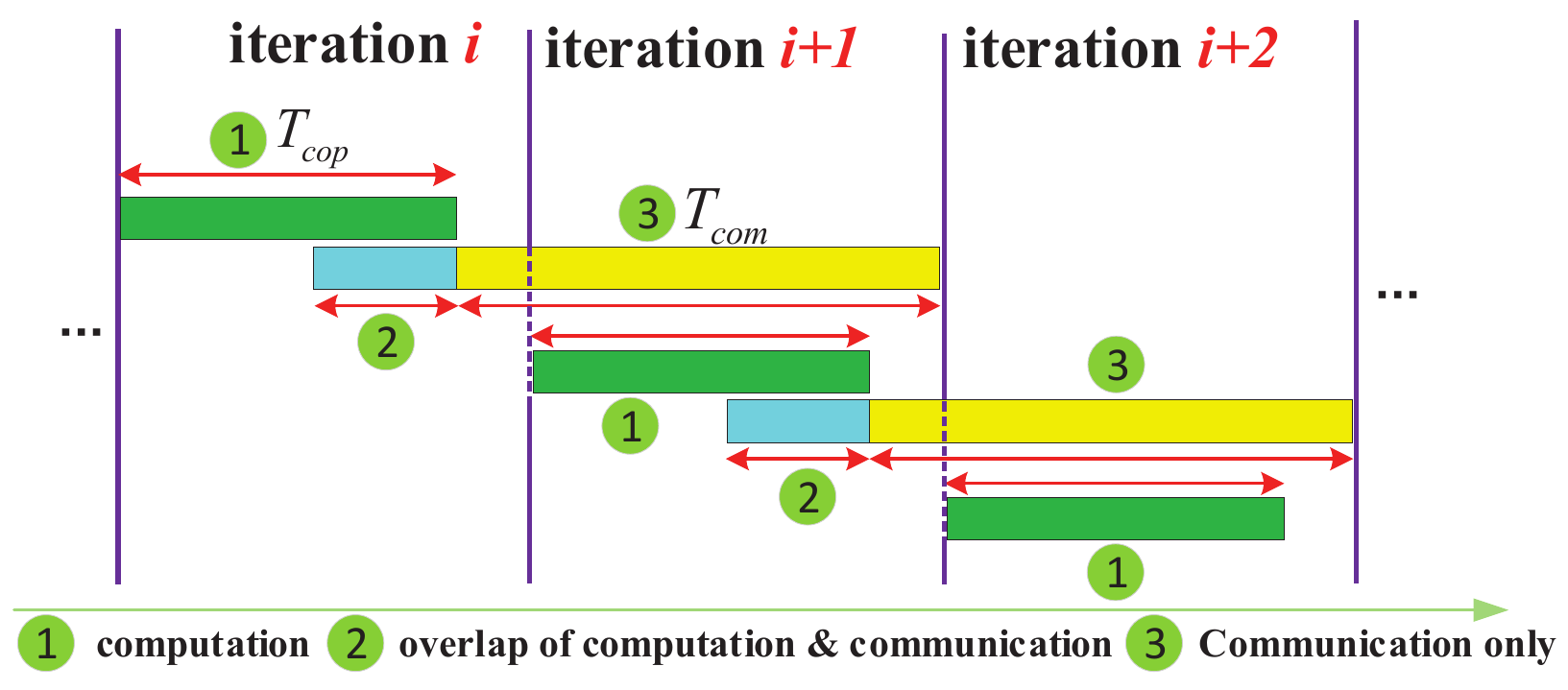}}
	\caption{Three cases of OD-SGD algorithm divided by the value of $T_{cop}$ and $T_{com}$. $T_{cop}$ is the overhead of \textbf{\textit{Phase}} \textcircled{1} while $T_{com}$ represents the time cost of \textbf{\textit{Phase}} \textcircled{3}.}
	\label{fig:od-performance} 
\end{figure}

\subsection{Algorithm Analysis on Performance Improvement}
This section introduces how to evaluate the performance improvement of OD-SGD, which is proposed for improving the distributed training performance and Fig.~{\ref{fig:od-mech}} illustrates its principles briefly. Actually, situation described in Fig.~{\ref{fig:od-mech}} can be further divided into three cases (Fig.~{\ref{fig:od-performance}}) according to the value of $T_{cop}$ and $T_{com}$. We evaluate the training performance improvement rate with Equation~{\ref{eqn:per-rate}}, $T_{org}$ is the time cost for completing one iteration with the SSGD algorithm 
while $T_{new}$ represents the overhead of one iteration under OD-SGD algorithm, which can be calculated by Equation~{\ref{eqn:T-new}}. When $T_{cop}<T_{com}$, $T_{new}$ is the average overhead of two adjacent iterations. It is noteworthy that the communication overhead of SSGD ($T_{com}^{'}$) does not equal that of OD-SGD ($T_{com}$), because the lightweight overheads of the introduced \textbf{\textit{local update}} operations may prolong the communication overhead by postponing the executions of the next \textbf{\textit{Pull}} and \textbf{\textit{Push}} operations. In addition, the degree of network congestion also differs as the communication traffic per second changes.


\begin{equation}
\label{eqn:per-rate}
IMP\_Rate = \frac{T_{org} - T_{new}}{T_{org}};
\end{equation}

\begin{equation}
\label{eqn:T-new}
T_{org} = T_{cop} + T_{com}^{'};
\hspace{3em}
T_{new}=\left\{
\begin{array}{lr}
T_{cop}       & {T_{com} <= T_{cop}}\\
\frac{T_{com} + T_{cop}}{2}     & {T_{cop} < T_{com}}
\end{array} 
\right.
\end{equation}

Equation~{\ref{eqn:per-rate}} and Equation~{\ref{eqn:T-new}} will lead to Equation~{\ref{eqn:per-rate-3}} and we have the next observations based on this new equation:
(1) When $T_{com} <= T_{cop}$, a larger $T_{com}^{'}$ will lead to better performance improvement when OD-SGD is applied because more overhead can be overlapped, the $T_{new}$ equals the $T_{cop}$ and has nothing to do with $T_{com}$.
(2) When $T_{com} > T_{cop}$, the improvement rate is able to reflect the influences of \textbf{\textit{local update}} operations on the communication overhead. Because the value of $T_{cop}$ is fixed, and the difference between $T_{com}$ and $T_{com}^{'}$ will determine the value of \textbf{\textit{IMP\_Rate}}, which results from the additional computational overheads brought about by the \textbf{\textit{local update}} operations. 
(3) The maximum value of performance improvement rate is 50\%, whereas it can not be attained as the value of $T_{com}^{'}$ is always less than $T_{com}$. 


\begin{equation}
\label{eqn:per-rate-3}
IMP\_Rate=\left\{
\begin{array}{lr}
1 - \frac{T_{cop}}{T_{cop}+T_{com}^{'}}       & {T_{com} <= T_{cop}}\\
1 - \frac{T_{com} + T_{cop}}{2(T_{cop}+T_{com}^{'})}     & {T_{cop} < T_{com}}
\end{array} 
\right.
\end{equation}


\subsection{Algorithm Implementation}
In order to implement the OD-SGD, we focus on two things in this section, one of which is to define the local updater, and the other one is to adjust the execution sequences of operations involved in the third stage.

\paragraph{Local Updater} 
Weight delay problem in OD-SGD results from the one-step delay mechanism, the retrieved weights in iteration \textit{i} is used for training task of iteration \textit{i+1}. To deal with the weight delay problem, local workers will update the copied weights with gradients calculated in iteration \textit{i}. Therefore,
OD-SGD algorithm requires a global updater in the parameter server and a local updater in the worker end. The original implementation of MXNet contains only one type of updater in the training process. In the distributed training mode, the updater for updating global weights is maintained in the parameter server, and updater runs in the worker node under the single node training mode. When implementing OD-SGD, updater in the parameter server can be defined by using the original function in MXNet and no modification is needed. We introduce a new function named \textit{set\_updater\_} to define the local updater and users can choose the algorithm through the option $\textit{''--optimizer\_local''}$ when launching the distributed training task. 
Three different algorithms are used in our experiments for the local update operations, namely the SGD, DC-ASGD-c and DC-ASGD-a. DC-ASGD \cite{dcasgd-2017} is proposed to tackle the gradient delay problem in ASGD, DC-ASGD-c and DC-ASGD-a are two different versions of it, corresponding to Equation~{\ref{eqn:DC-ASGD-c-1}}  and Equation~{\ref{eqn:DC-ASGD-a-1}} separately. It is noteworthy that DC-ASGD-a calls for more computation cost than DC-ASGD-c for one update operation.

\begin{equation}
\label{eqn:DC-ASGD-c-1}
\textbf{w}_{t+\tau+1} = \textbf{w}_{t+\tau} - \eta(g(\textbf{w}_{t}) +\lambda*g(\textbf{w}_{t}) \odot g(\textbf{w}_{t}) \odot (\textbf{w}_{t+\tau} - \textbf{w}_{t}))
\end{equation}

\begin{equation}
\label{eqn:DC-ASGD-a-1}
\textbf{w}_{t+\tau+1} = \textbf{w}_{t+\tau} - \eta(g(\textbf{w}_{t}) +(\lambda/sqrt(MeanSquare(t) + \epsilon)) \odot g(\textbf{w}_{t}) \odot g(\textbf{w}_{t}) \odot (\textbf{w}_{t+\tau} - \textbf{w}_{t}))
\end{equation}

\begin{equation}
\label{eqn:mean-square-1}
MeanSquare(t) = m \cdot MeanSquare(t-1) + (1-m) \cdot g(\textbf{w}_{t})^{2}
\end{equation}

\begin{figure}[t]
	\centering
	\begin{minipage}{0.45\textwidth}
		\centering
		\includegraphics[width=5.4cm, height=2.8cm]{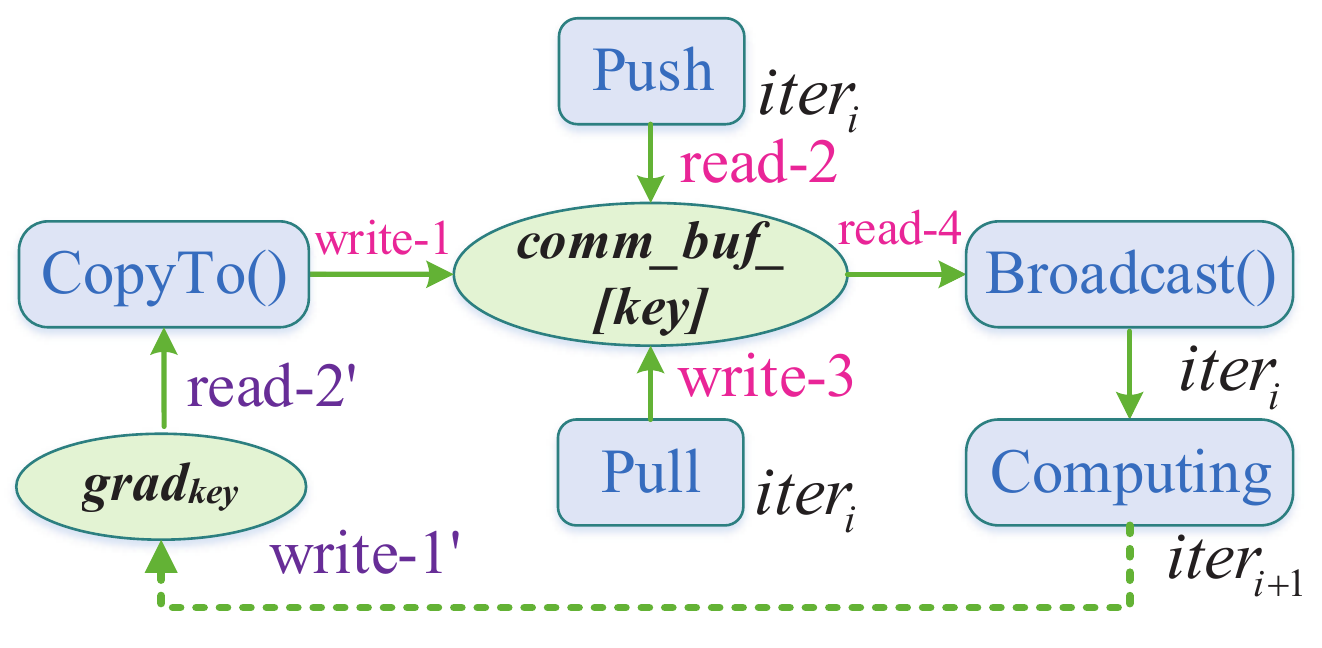}
		\caption{Training procedures of the SSGD. Numbers at the end of \textit{read} and \textit{write} operations describe the order in which they are performed.}
		\label{fig:flow1}
	\end{minipage}	
	\begin{minipage}{0.45\textwidth}
		\centering
		\includegraphics[width=5.4cm, height=2.8cm]{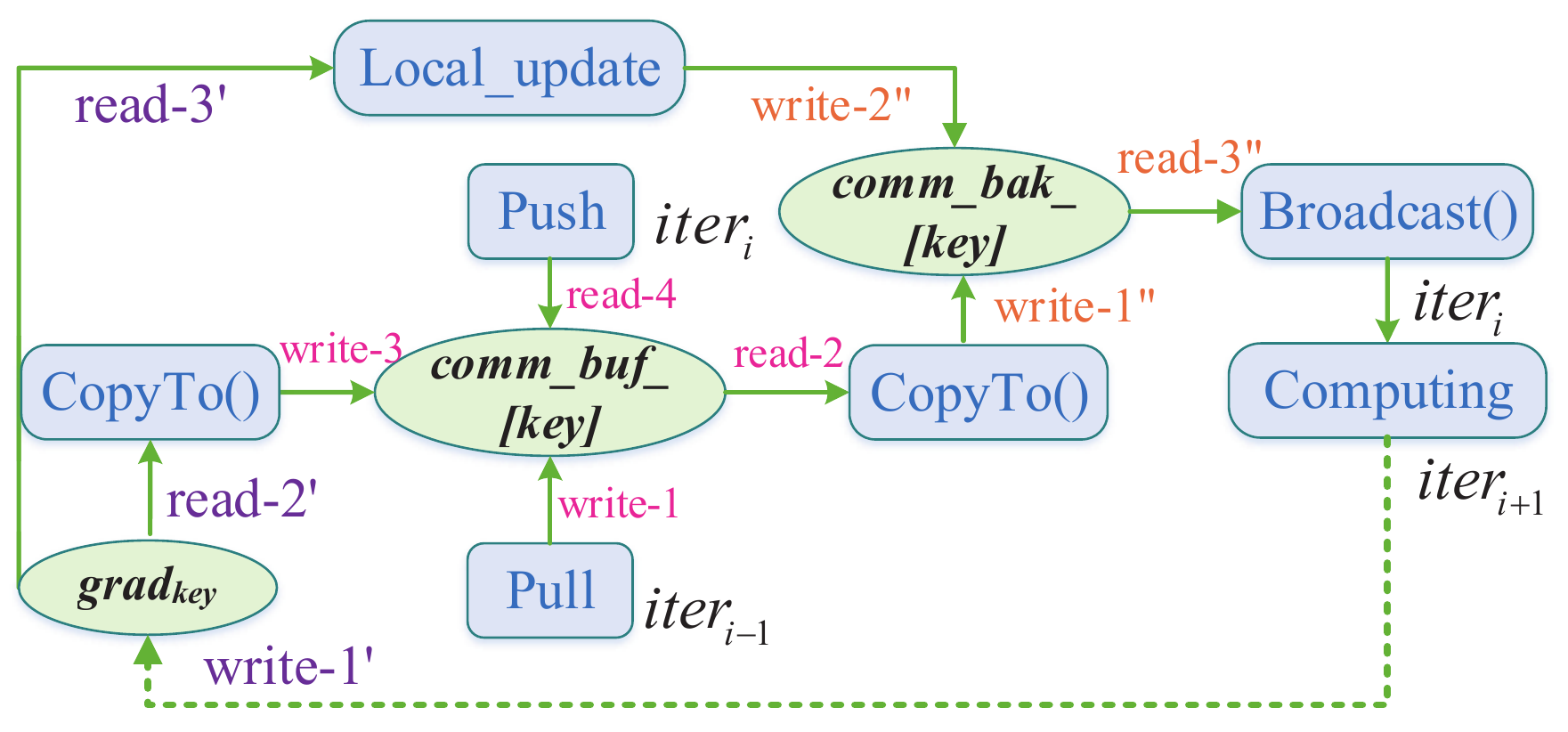}
		\caption{Training procedures of the OD-SGD. Numbers at the end of \textit{read} and \textit{write} operations describe their execution orders.}
		\label{fig:flow2}
	\end{minipage}
\end{figure}

\paragraph{Execution Sequences of Operations} 
In the warm-up stage, the cluster conducts training task under the SSGD mechanism, and Fig.~\ref{fig:flow1} presents the original training procedures of SSGD method. \textit{comm\_buf\_} is the shared variable between \textbf{\textit{Push}} and \textbf{\textit{Pull}} operations. The aggregated local gradient will be copied to \textit{comm\_buf\_[key]} firstly, then \textbf{\textit{Push}} operation will read the gradient 
from \textit{comm\_buf\_[key]} and send it to the parameter server. \textit{key} is the index of the corresponding parameter. After the completion of \textbf{\textit{Push}} operation,  \textbf{\textit{Pull}} operation will pull back the updated weight from the parameter server and store it in the \textit{comm\_buf\_[key]}. Finally the value in \textit{comm\_buf\_[key]} is broadcast to local devices for computing task of the next iteration. 
Fig.~\ref{fig:flow2} demonstrates the training process of OD-SGD under the third stage. 
The \textbf{\textit{Pull}} operation of the previous iteration pulls back the weight and 
keeps it in the \textit{comm\_buf\_[key]}, the \textit{CopyFromTo} function then makes a copy of the value and saves it to \textit{comm\_bak\_[key]}. The aggregated gradient computed in current iteration will overwrite the \textit{comm\_buf\_[key]} for the \textbf{\textit{Push}} operation and be used for \textit{local\_update} process. Ultimately, the updated value in \textit{comm\_bak\_[key]} is broadcast for launching the next iteration task. 
The OD-SGD algorithm involves the above two training processes, we implement the training flow shown in Fig.~\ref{fig:flow2} and it will be switched on after the warm-up stage. 
If anyone is interested in the implementation details, the code is open source on github.

\section{Evaluation}
As is mentioned above, we come up with OD-SGD to improve the distributed deep learning training performance, and we evaluate its performance in this section. We firstly introduce the hardware configurations, datasets and neural networks used in our experiments. Secondly, we demonstrate the performance comparisons with ASGD, SSGD and DC-ASGD, 
test accuracy of all involved neural networks is TOP-1 accuracy.
Thirdly, we assess the performance improvement of OD-SGD under various cluster scales and network bandwidths. 
Finally, we make some sensitivity analysis on warm-up period.

\subsection{Experiments Configuration}

The OD-SGD is implemented in MXNet framework and we conduct the evaluation experiments on two different GPU clusters. One of cluster is equipped with 2 K80 (dual GPUs) Tesla GPUs on each node, interconnected with both Ethernet (10Gbps) and InfiniBand (56Gbps), while the other cluster is equipped with 4 V100 Tesla GPUs on each node, interconnected with InfiniBand (56Gbps) only. The topology of both clusters is the simple star network topology. Machines of the K80 GPU cluster are installed with Red Hat 4.8.3, CUDA 8.0 and cuDNN 6.0 while machines of the V100 GPU cluster are installed with Centos 7.6, CUDA 10.0 and cuDNN 7.4.1. Because of the limited number of nodes, we set only one parameter server throughout our experiments.

\textbf{\textit{Datasets}}. To prove the effectiveness of our algorithm, we utilize three datasets in our experiments: (1) MNIST \cite{lecun1998mnist}, it is a database of handwritten digits, which consists of $60,000$ training examples and $10,000$ test samples. (2) CIFAR-10 \cite{CIFAR-10}, it contains $60,000$ 32$x$32 color images in 10 different classes, $50,000$ pictures for training and $10,000$ for testing. (3) ImageNet ILSVRC2012 \cite{deng2009imagenet}, it is a subset of ImageNet and contains 1.2 million pictures in 1000 categories. 

\textbf{\textit{Neural Networks}}. Neural networks applied in our experiments comes with the MXNet framework, including MobileNet1.0 \cite{mobilenet}, ResNet18\_V2, ResNet-20, ResNet-50 \cite{he2016deep}, ranking from low-complexity (MobileNet1.0) to high-complexity (ResNet-50). MobileNet1.0 is trained with MNIST dataset, ResNet18\_V2 and ResNet-20 correspond to CIFAR-10 while ImageNet is for ResNet-50. For the sake of fairness, all these neural networks are trained from the same randomly initialized model. 

We implement the OD-SGD and DC-ASGD on MXNet platform and set 200 iterations for the warm-up stage. In addition, the SSGD, ASGD have already been implemented, we compare the accuracy of these four algorithms to evaluate our proposed OD-SGD algorithm. 
With or without momentum makes almost no difference to the accuracy, therefore, the momentum of DC-ASGD is set to $0$ in the experiments \cite{dcasgd-2017}. 


\subsection{Experimental Results on MNIST}
Training MobileNet1.0 with MNIST dataset is quick and efficient, it is unnecessary for large-scale distributed deployment. We conduct experiments on MobileNet1.0 to prove that OD-SGD algorithm is applicable to neural network of low-complexity. We perform the training task for 40 epochs, with a mini-batch size of 128, the training task is deployed on 4 worker nodes interconnected with Ethernet. The initial learning rate of the 4 different algorithms is set to 0.05, the momentum for ASGD and DC-ASGD is $0$, while $0.9$ for SSGD and the global updater of OD-SGD. We chose  SGD algorithm as the local update function of OD-SGD, whose learning rate and momentum are the same as SSGD. In addition, we chose the constant version of DC-ASGD  (DC-ASGD-c) for comparison with $\lambda=0.04$.

\makeatletter\def\@captype{figure}\makeatother
\begin{figure}
	\begin{minipage}{0.42\linewidth}
		\centerline{\includegraphics[width=5.2cm, height=3.2cm]{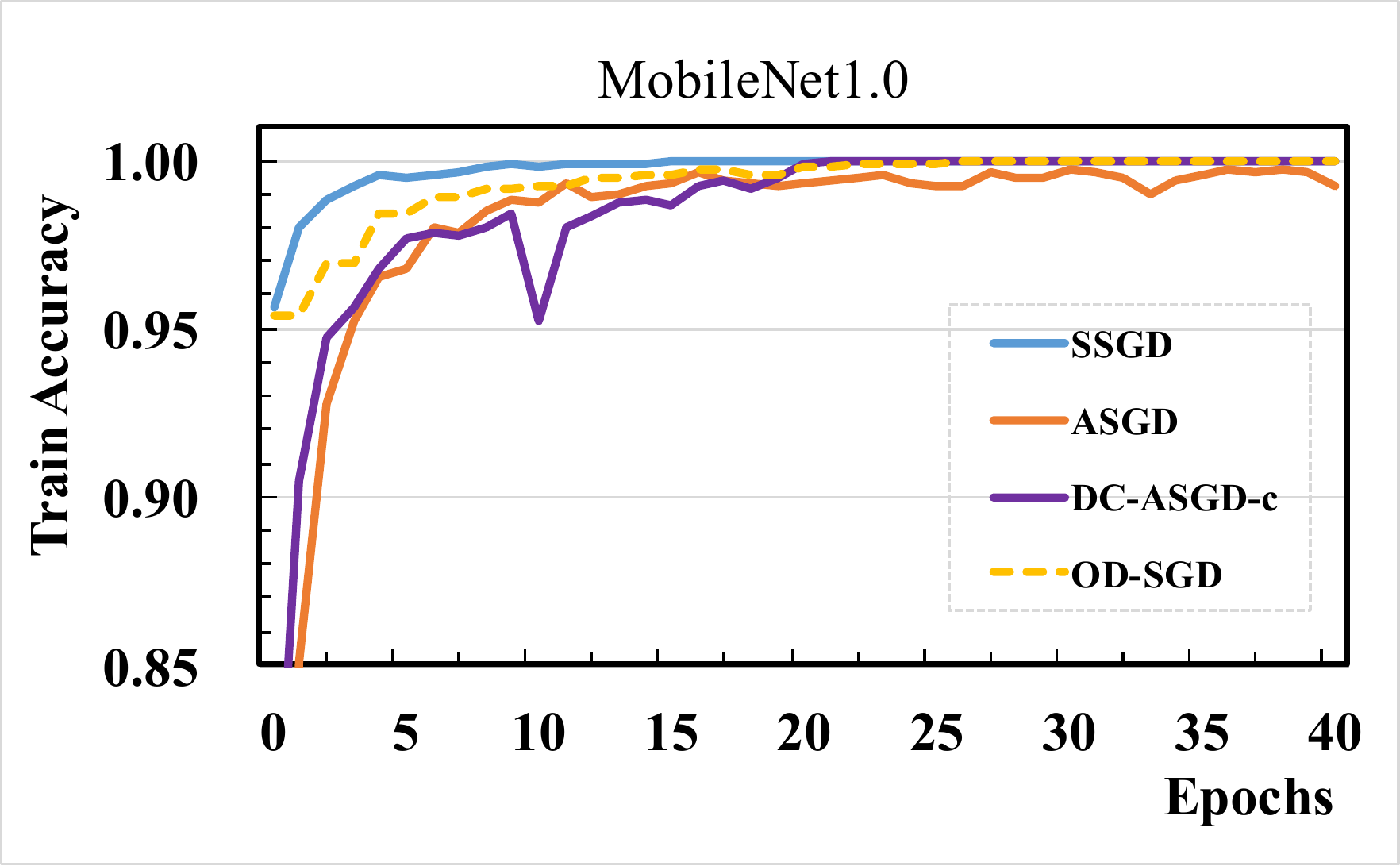}}
		\centerline{(a)}
	\end{minipage}
	\begin{minipage}{0.42\linewidth}
		\centerline{\includegraphics[width=5.2cm, height=3.2cm]{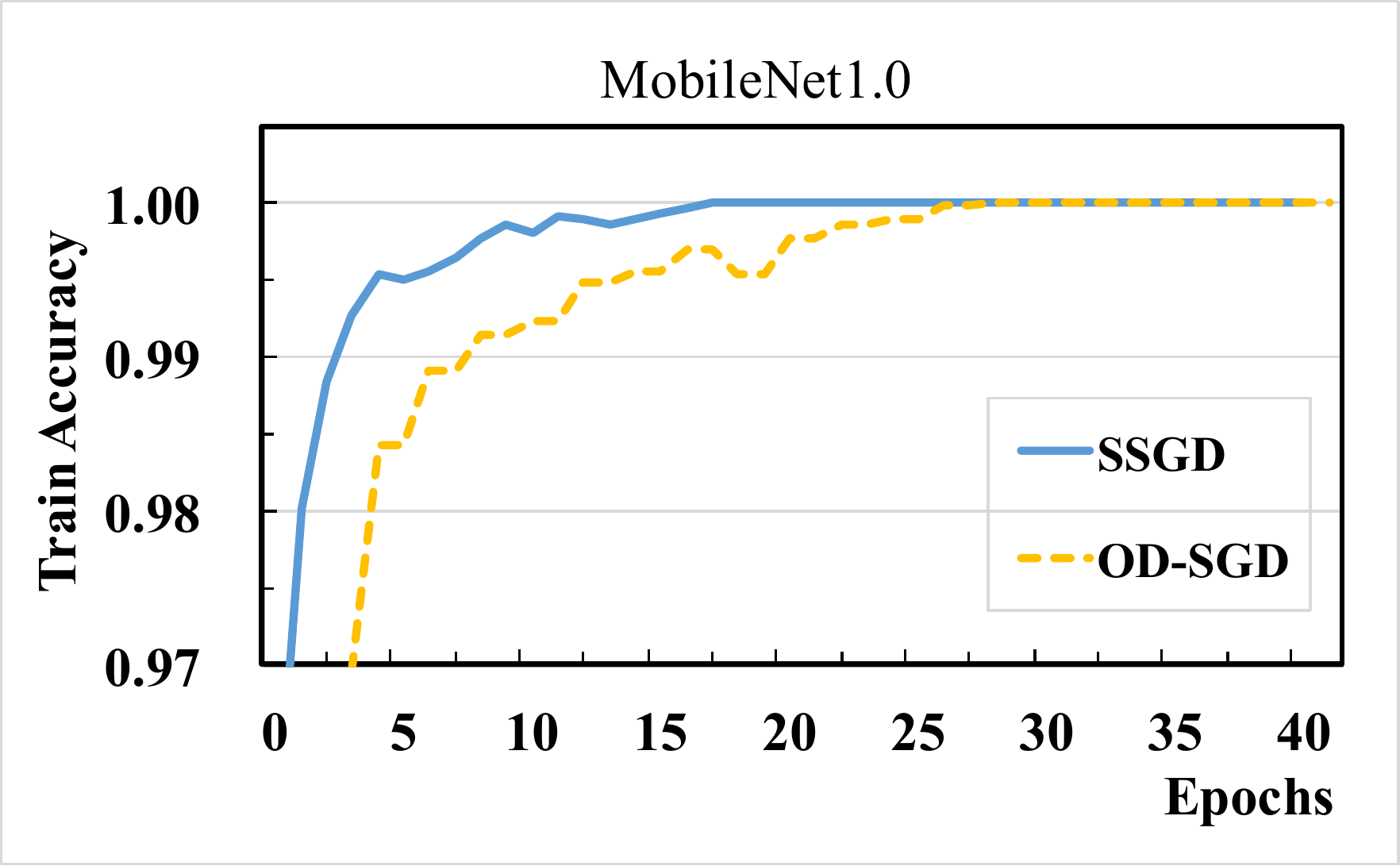}}
		\centerline{(b)}
	\end{minipage}
	\vfill
	\begin{minipage}{0.42\linewidth}
		\centerline{\includegraphics[width=5.2cm, height=3.2cm]{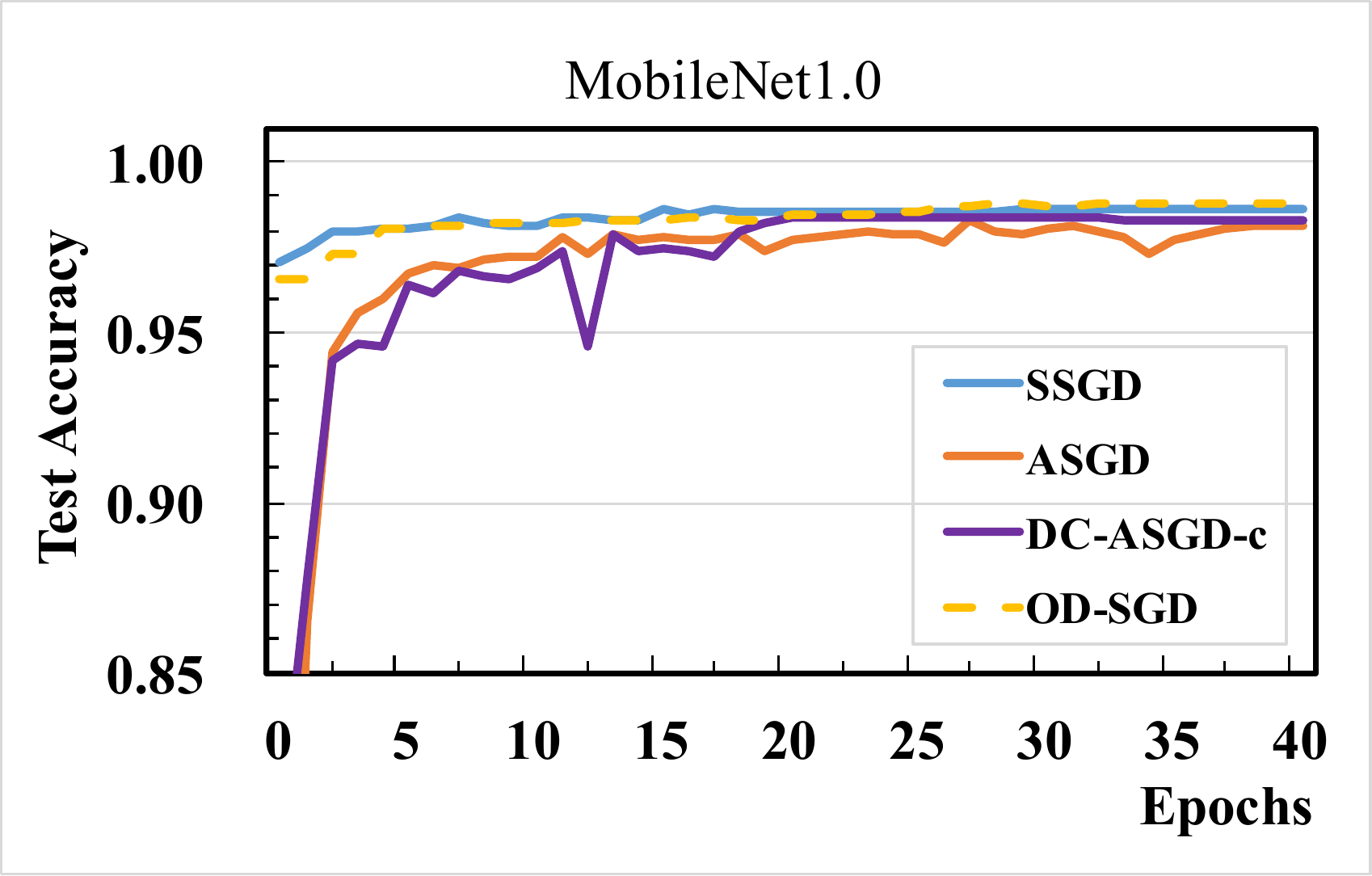}}
		\centerline{(c)}
	\end{minipage}
	\begin{minipage}{0.42\linewidth}
		\centerline{\includegraphics[width=5.2cm, height=3.2cm]{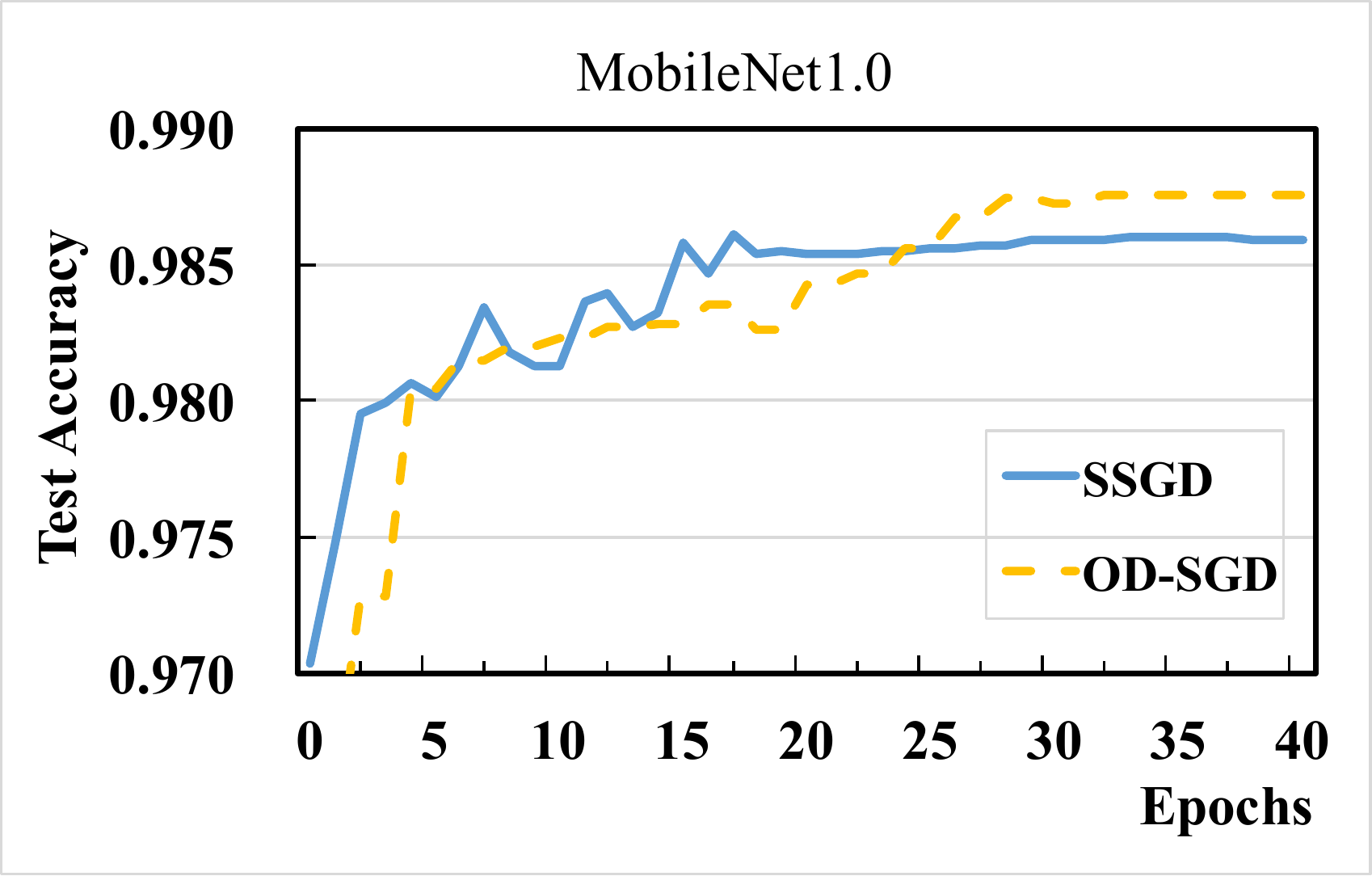}}
		\centerline{(d)}
	\end{minipage}
	\caption{Train accuracy and test accuracy of MobileNet1.0 under different epochs on MNIST data set.}
	\label{fig:mobilenet1}
\end{figure}

In the first place, we demonstrate the learning curves under fixed number of training epochs as shown in Fig.~{\ref{fig:mobilenet1}}. From figure $(a)$ and $(c)$ we can notice that the convergence speed of SSGD and OD-SGD-c is obviously faster than that of ASGD and DC-ASGD. SSGD and OD-SGD share similar curves while the learning patterns of ASGD and DC-ASGD are alike. The better performance of synchronous algorithms (SSGD, OD-SGD) mainly results from the low-complexity of neural network and the computing capacity differences of the workers. 
The low-complexity leads to low computing overhead for each iteration, coupled with the different computation capabilities, the delayed gradients problem is further deteriorated when training with asynchronous algorithms (ASGD, DC-ASGD). Besides, the training tasks for workers are launched at different times, this kind of delay also imposes an negative effect on the delayed gradients problem. Figure $(b)$ and $(d)$ demonstrate the further comparisons between SSGD and OD-SGD. The training accuracy curve of OD-SGD is similar to SSGD while its accuracy is slightly lower than that of SSGD in the first 25 epochs. This results from the fact that OD-SGD utilizes the features of SSGD and ASGD algorithms, the asynchronous mechanism has a negative influence on the accuracy. Nevertheless,  OD-SGD introduces the SGD algorithm for local update to ensure the accuracy.

\makeatletter\def\@captype{figure}\makeatother
\begin{figure}[t]
	\begin{minipage}{0.42\linewidth}
		\centerline{\includegraphics[width=5.6cm, height=3.5cm]{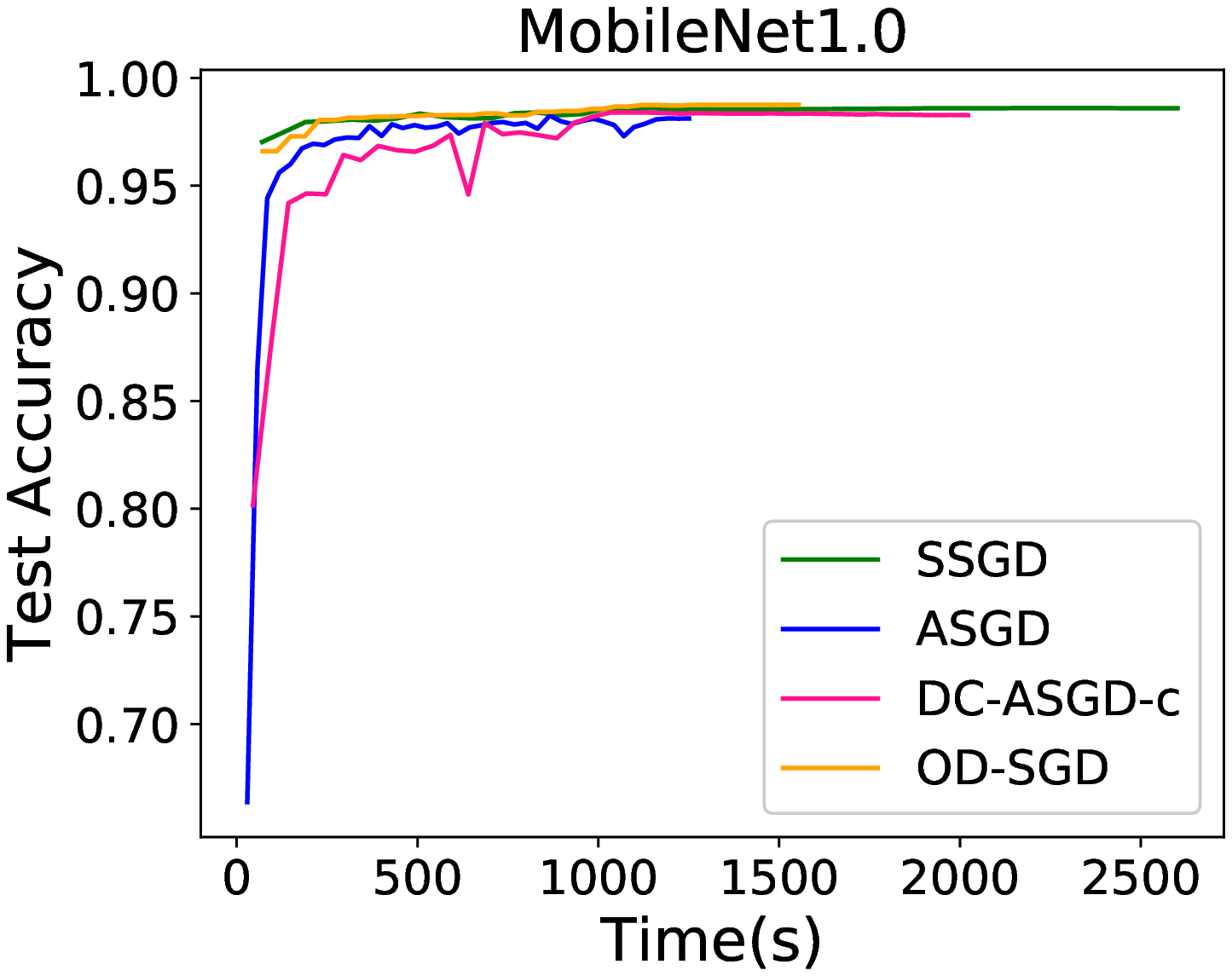}}
		\centerline{(a)}
	\end{minipage}
	\begin{minipage}{0.42\linewidth}
		\centerline{\includegraphics[width=5.6cm, height=3.5cm]{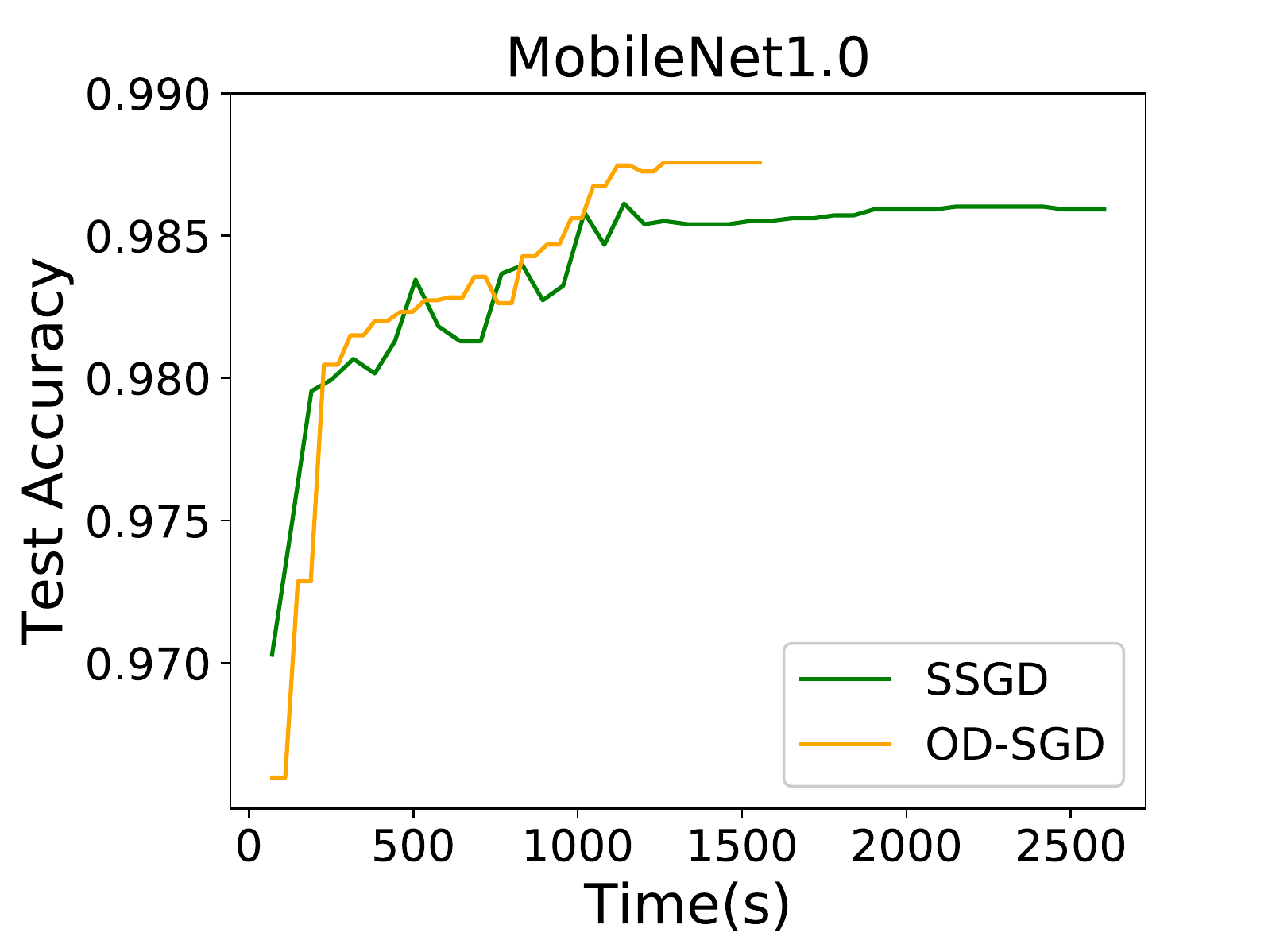}}
		\centerline{(b)}
	\end{minipage}
	\caption{Train accuracy and test accuracy of MobileNet1.0 w.r.t wallclock time on MNIST dataset.}
	\label{fig:mobilenet2}
\end{figure}

We then compare the convergence speed of these algorithms as shown in Fig.~{\ref{fig:mobilenet2}}. As the total training time for 40 epochs does not vary greatly of the algorithms, we present their test accuracy throughout the whole training process. From subgraph $(a)$ we can notice that it takes the longest time for SSGD to complete the training task ($\approx2600s$) due to the synchronous barrier, followed by DC-ASGD, OD-SGD and ASGD. It is obvious that convergence speed of of synchronous algorithms is faster than that of asynchronous algorithms. Moreover, the ASGD achieves faster convergence speed than DC-ASGD algorithm, because the latter brings in extra computation overhead in the parameter server to compensate the delayed gradients. The delayed gradients problem may be alleviated, it also prolongs training time for one iteration, which is more distinct for neural network of low-complexity. 
Subgraph$(b)$ provides further comparisons of convergence speed between SSGD and OD-SGD. It is noteworthy that their curves share similar variation trends, proving that the OD-SGD algorithm has the characteristics of SSGD. Furthermore, the convergence speed of OD-SGD is much faster than SSGD due to the feature of ASGD, it achieves the best test accuracy at about 700 seconds while SSGD fulfills the target at 1200 seconds. Nevertheless, the SSGD algorithm experiences a drop after the peak accuracy and the curve of OD-SGD is more stable.

Table~{\ref{tab:mobilenet-speed}} demonstrates the training speed and classification accuracy on MNIST test set using MobileNet1.0. The following conclusions can be drawn from the table: (1) ASGD algorithm achieves the fastest training speed because there is no synchronous obstacle. DC-ASGD algorithm utilizes the same asynchronous mechanism as ASGD, however, the introduced computational operations slow down its speed to process the training data. Training speed of OD-SGD is faster than DC-ASGD algorithm, because the introduced computational overhead for local update can be partly overlapped by the \textbf{\textit{Push}} communication operation. (2) The training accuracy of OD-SGD can be as high as SSGD, both of their accuracy is $100\%$ after training for 15 epochs. (3) OD-SGD achieves the best test accuracy and its final value is $98.756\%$. In addition, higher training accuracy may not ensure higher test accuracy. The accuracy of DC-ASGD is slightly higher than that of SSGD while its training accuracy is much less. (4) Experiments on MNIST dataset with MobileNet1.0 prove that our proposed OD-SGD algorithm is suitable for training model of low-complexity, which keeps fast training speed and high test accuracy, its performance is slightly better than that of DC-ASGD algorithm. 

\vspace{3ex}
\makeatletter\def\@captype{table}\makeatother
\begin{minipage}{0.48\textwidth}
	\caption{Training speed and classification accuracy on MNIST test set.}
	\label{tab:mobilenet-speed}
	\begin{center}
		\begin{tabular}{c|c|c|c}
			\hline
			\multirow{2}*{Algorithms} & Speed  & Train & Test\\
			& (\textit{samples/s}) & Acc (\%) & Acc (\%)\\
			\hline
			SSGD & 236.63 & 100 & 98.592\\
			\hline
			ASGD & \color{red}{\textbf{505.29}} & 99.731 & 98.252 \\
			\hline
			DC-ASGD & 306.434 & 99.963 & 98.396 \\
			\hline
			OD-SGD & 401.395 & \color{red}{\textbf{100}} & \color{red}{\textbf{98.756}} \\
			\hline
		\end{tabular}
	\end{center}
\end{minipage}	
\hspace{3ex}
\makeatletter\def\@captype{table}\makeatother
\begin{minipage}{0.48\textwidth}
	\caption{Test accuracy on CIFAR-10 after the moving average operation.}
	\label{tab:cifar-avg}
	\begin{center}
		\begin{tabular}{c|c|c|c|c}
			\hline
			\# of  & \multicolumn{4}{|c}{Test Accuracy (\%)} \\
			workers & SSGD & ASGD & DC-ASGD & OD-SGD\\
			\hline
			4 & \color{red}{80.482}& 79.890 & 80.156 & 80.426 \\
			\hline
			8 & 79.673 & 78.232 & 78.985 & \color{red}{79.678} \\
			\hline
			12 & 78.844 & 78.288& 77.659 & \color{red}{78.914} \\
			\hline	
			1 & \multicolumn{4}{|c}{\color{blue}{81.048}}\\
			\hline
		\end{tabular}
	\end{center}
\end{minipage}		
\vspace{3ex}

\subsection{Experimental Results on CIFAR-10}
In this section, we investigate the training performance of those algorithms on CIFAR-10 dataset (without data augmentation), using the neural network ResNet18\_v2. We conduct the training for 160 epochs, with a mini-batch size of 128. 
Training task is deployed on clusters with 4, 8 and 12 worker nodes (In addition to a parameter server node), corresponding to 16, 32 and 48 GPUs, respectively, and we make use of the InfiniBand network for communication process.

We conduct all the training for 160 epochs, with a mini-batch size of 128, and the learning rate is decreased by 10 times after training for 80 and 120 epochs. 
Hyper parameters for DC-ASGD utilize the recommended value ($\eta=0.5$, $\lambda=0.04$ for DC-ASGD-c and $\eta=0.5$, $\lambda=2$, $m=0.95$ for DC-ASGD-a) [61]. ASGD and SSGD share the same learning rates ($\eta_{0}=0.1$ ($M=4$), $\eta_{1}=0.2$ ($M=8$), $\eta_{2}=0.2$ ($M=12$)), $M$ means the number of workers in the training process. Momentums for DC-ASGD and ASGD are $0$ while $0.9$ for SSGD. 
For OD-SGD, SSGD is utilized for the global update operation in the parameter server. DC-ASGD-c and DC-ASGD-a are used for local update operations, corresponding to OD-SGD-c and OD-SGD-a. When training with 4 workers, the DC-ASGD-c is applied, DC-ASGD-a method can also be used while it brings about more additional computational overhead. However, when the cluster scale is increased to 8 or 12 worker nodes, local update operation with DC-ASGD-c cannot achieve the target accuracy and DC-ASGD-a is used. 
We compare the training performance and convergence speed of the above four algorithms (SSGD, ASGD, DC-ASGD-c and OD-SGD). The DC-ASGD-c is chosen for comparison because its computational cost is lower than DC-ASGD-a and its test accuracy is higher than SSGD according to [61].

\makeatletter\def\@captype{figure}\makeatother
\begin{figure}
	\begin{minipage}{0.33\linewidth}
		\centerline{\includegraphics[width=4.8cm, height=3cm]{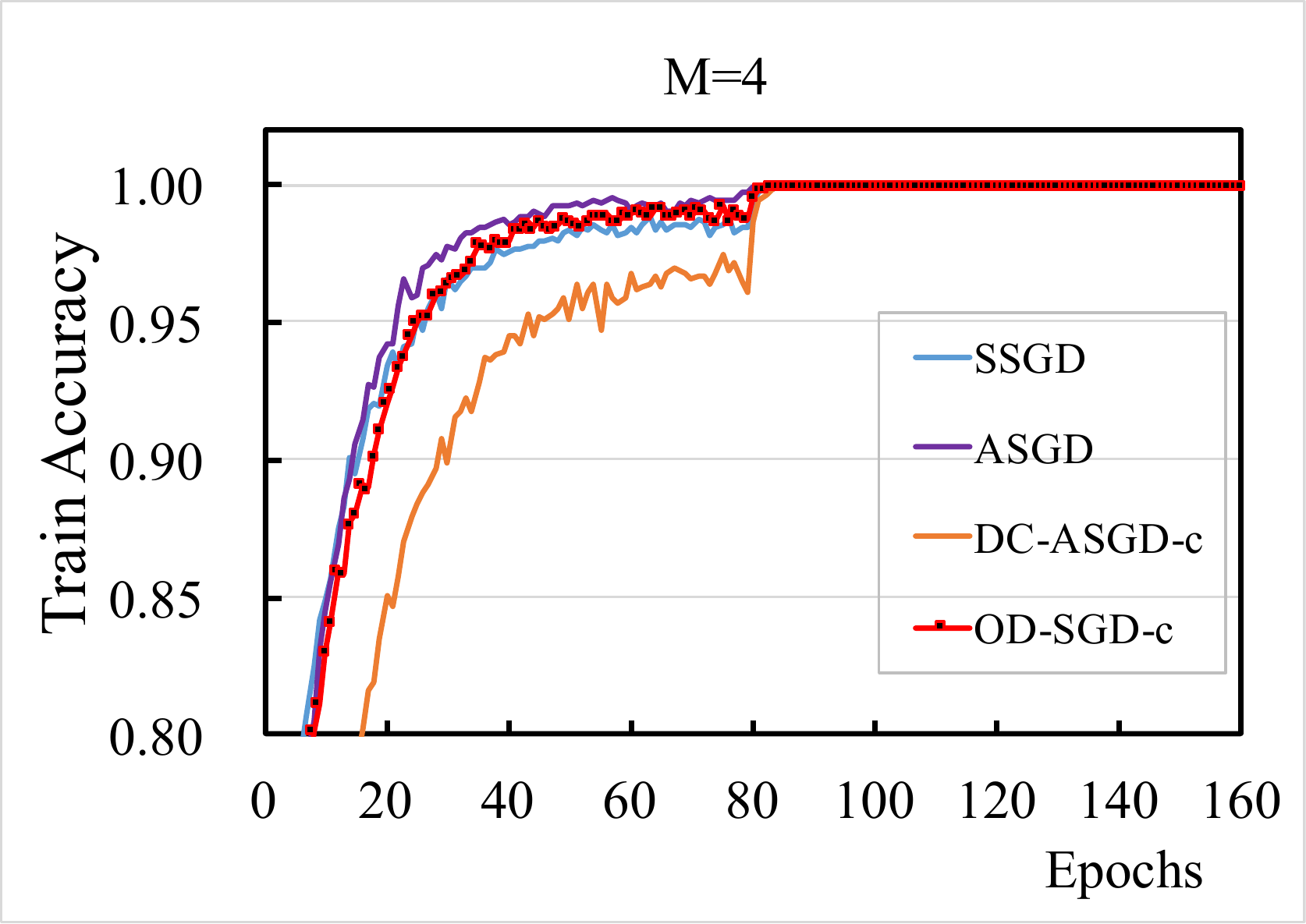}}
		\centerline{(a)}
	\end{minipage}
	\begin{minipage}{0.33\linewidth}
		\centerline{\includegraphics[width=4.8cm, height=3cm]{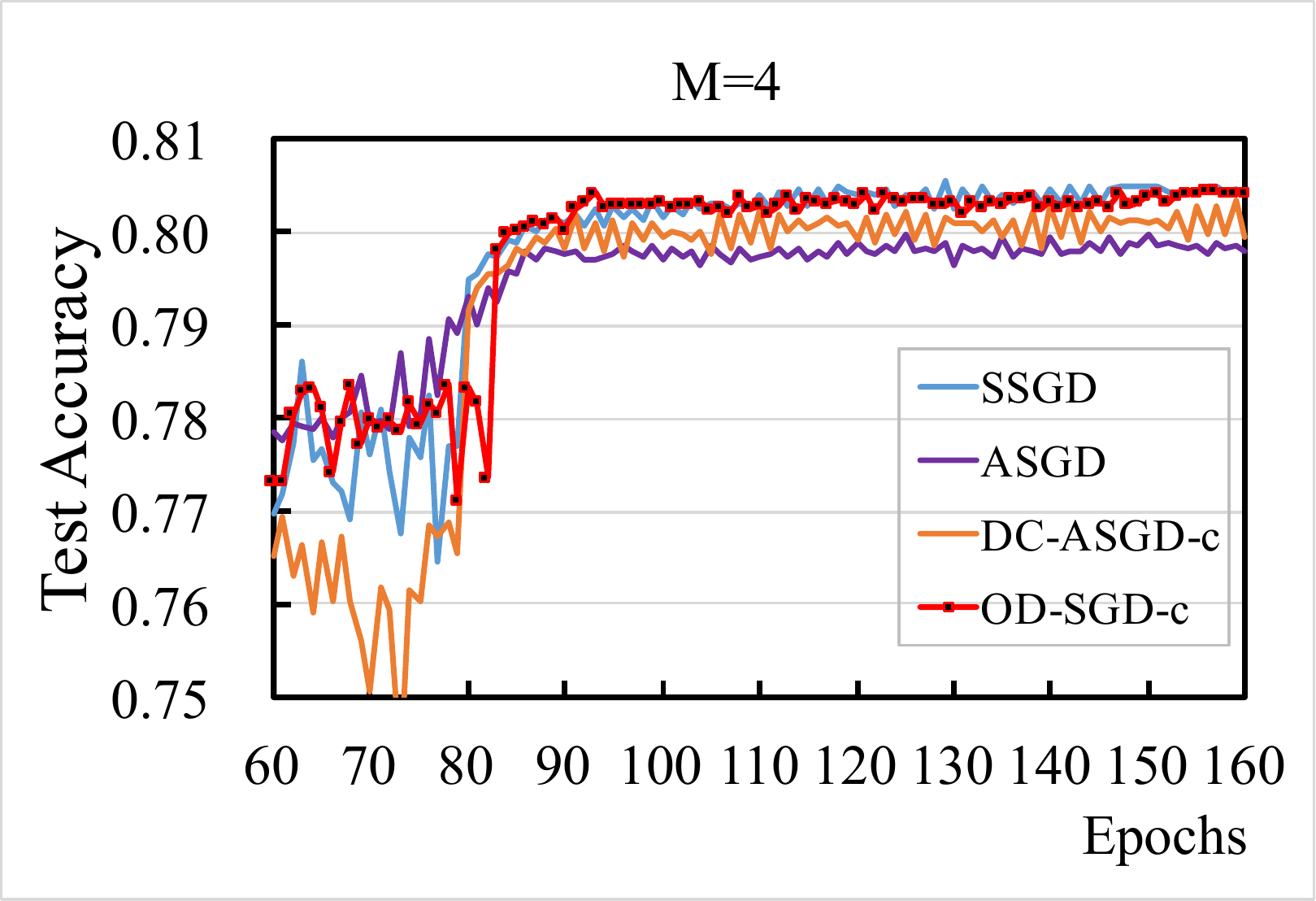}}
		\centerline{(b)}
	\end{minipage}
	\begin{minipage}{0.33\linewidth}
		\centerline{\includegraphics[width=4.8cm, height=3cm]{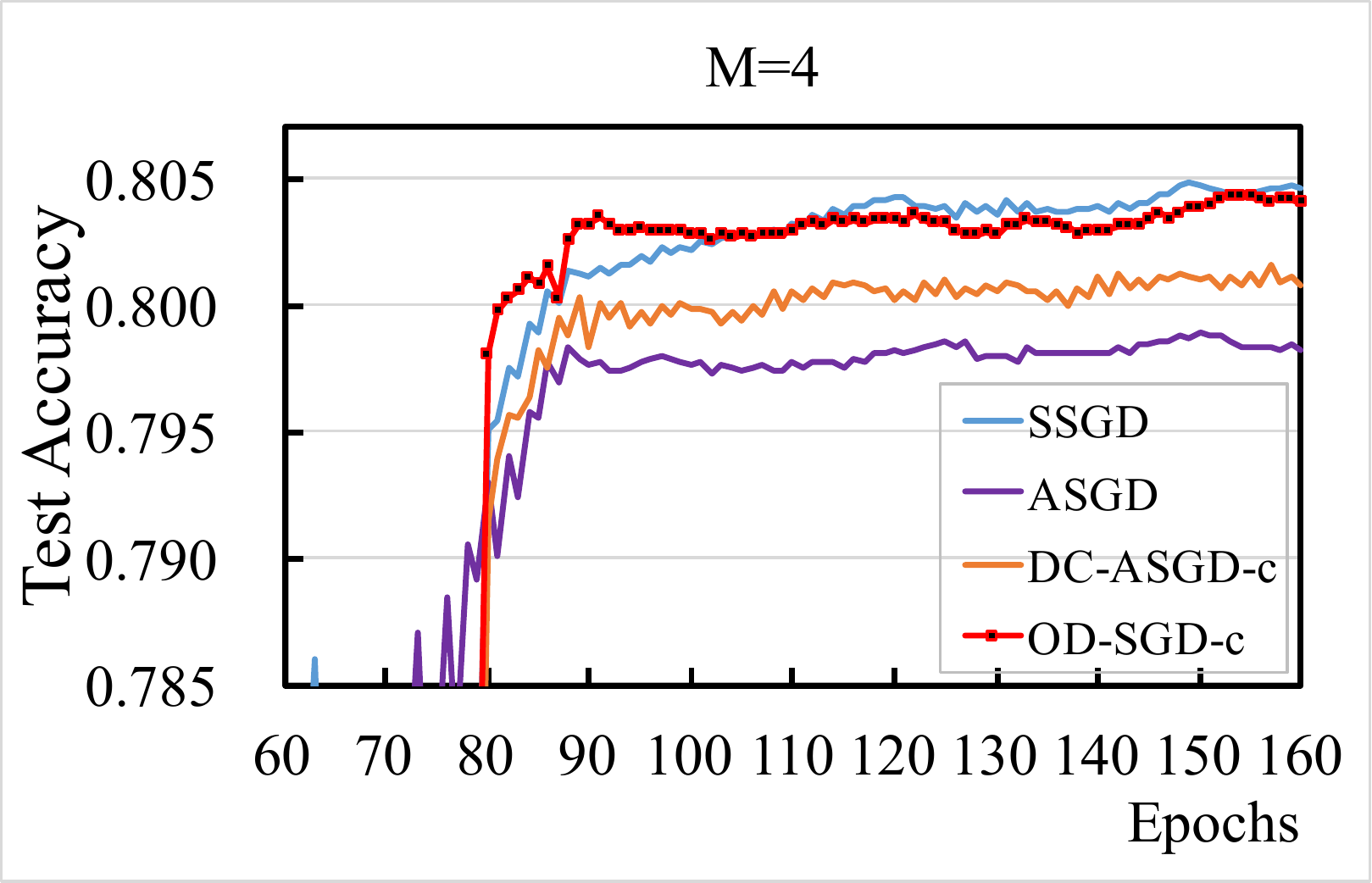}}
		\centerline{(c)}
	\end{minipage}
	\vfill
	\begin{minipage}{0.33\linewidth}
		\centerline{\includegraphics[width=4.8cm, height=3cm]{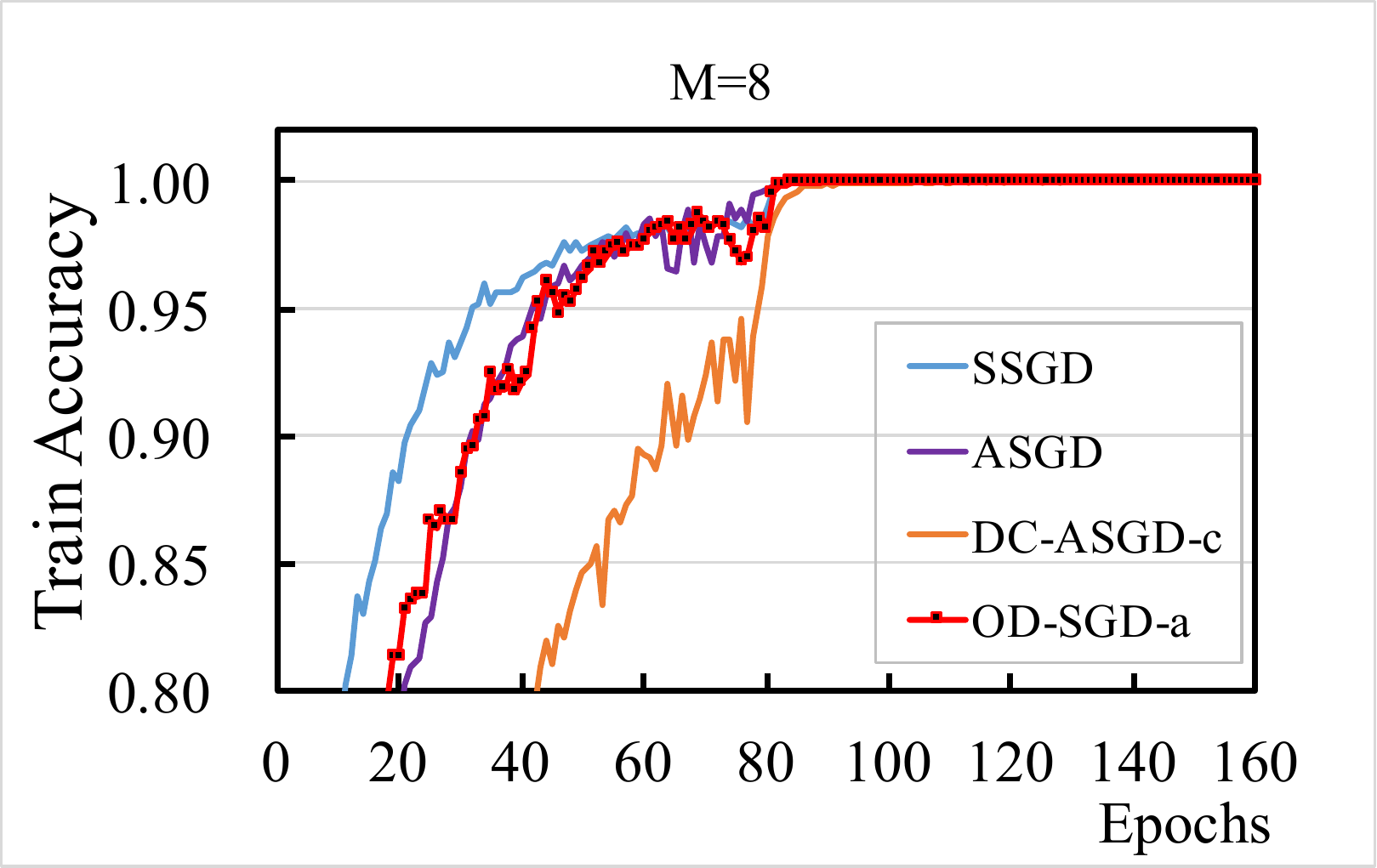}}
		\centerline{(d)}
	\end{minipage}
	\begin{minipage}{0.33\linewidth}
		\centerline{\includegraphics[width=4.8cm, height=3cm]{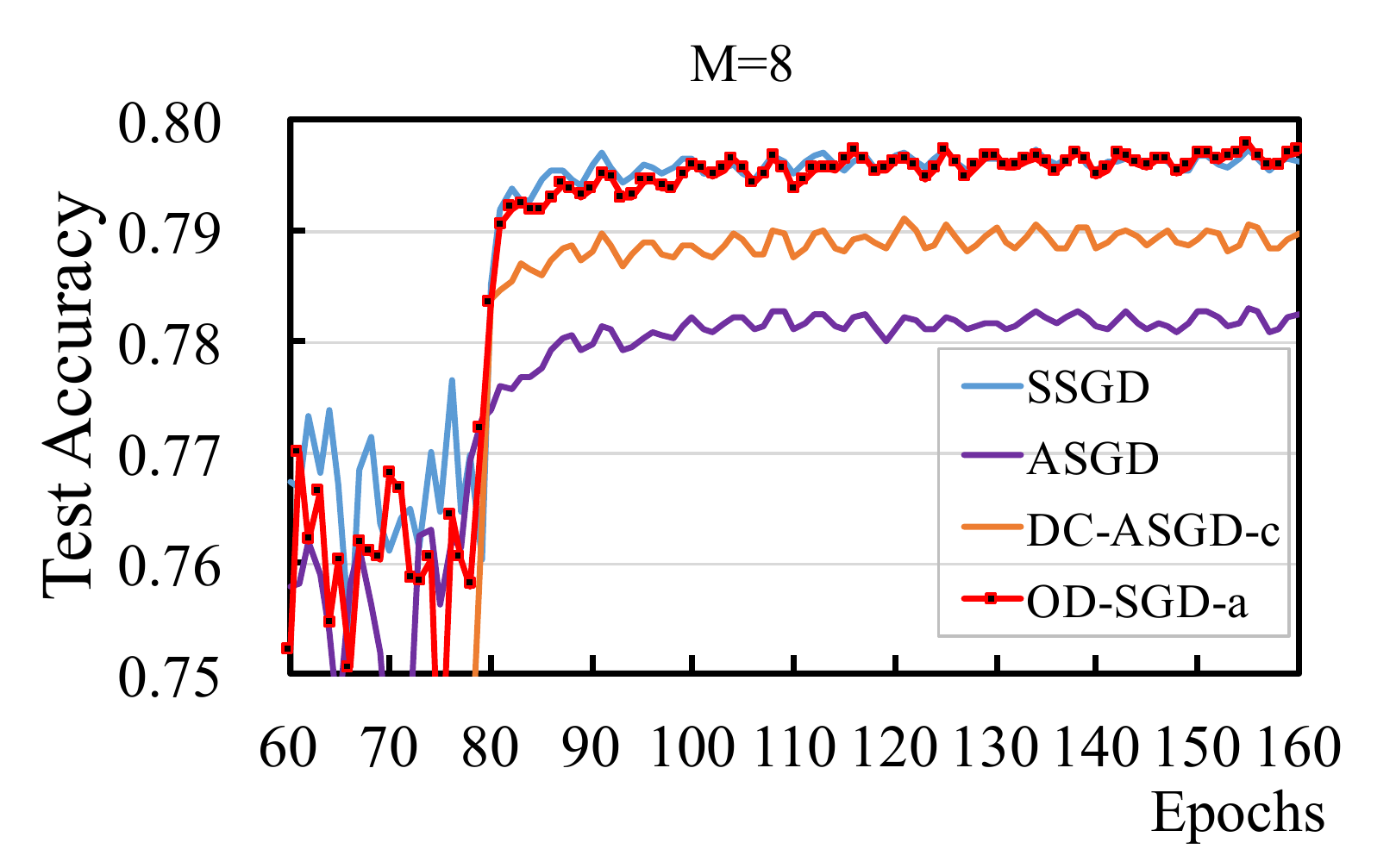}}
		\centerline{(e)}
	\end{minipage}
	\begin{minipage}{0.33\linewidth}
		\centerline{\includegraphics[width=4.8cm, height=3cm]{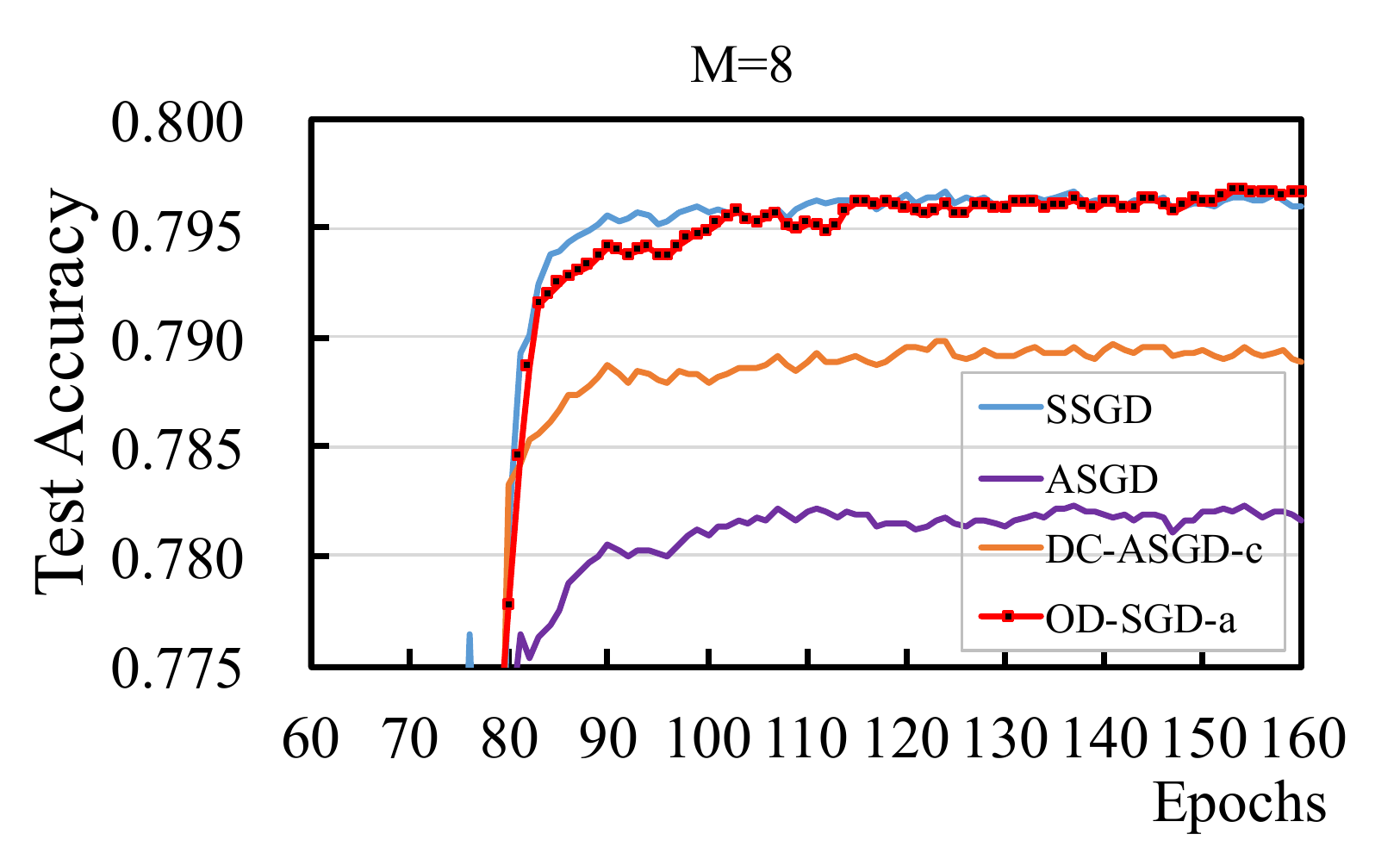}}
		\centerline{(f)}
	\end{minipage}	
	\vfill
	\begin{minipage}{0.33\linewidth}
		\centerline{\includegraphics[width=4.8cm, height=3cm]{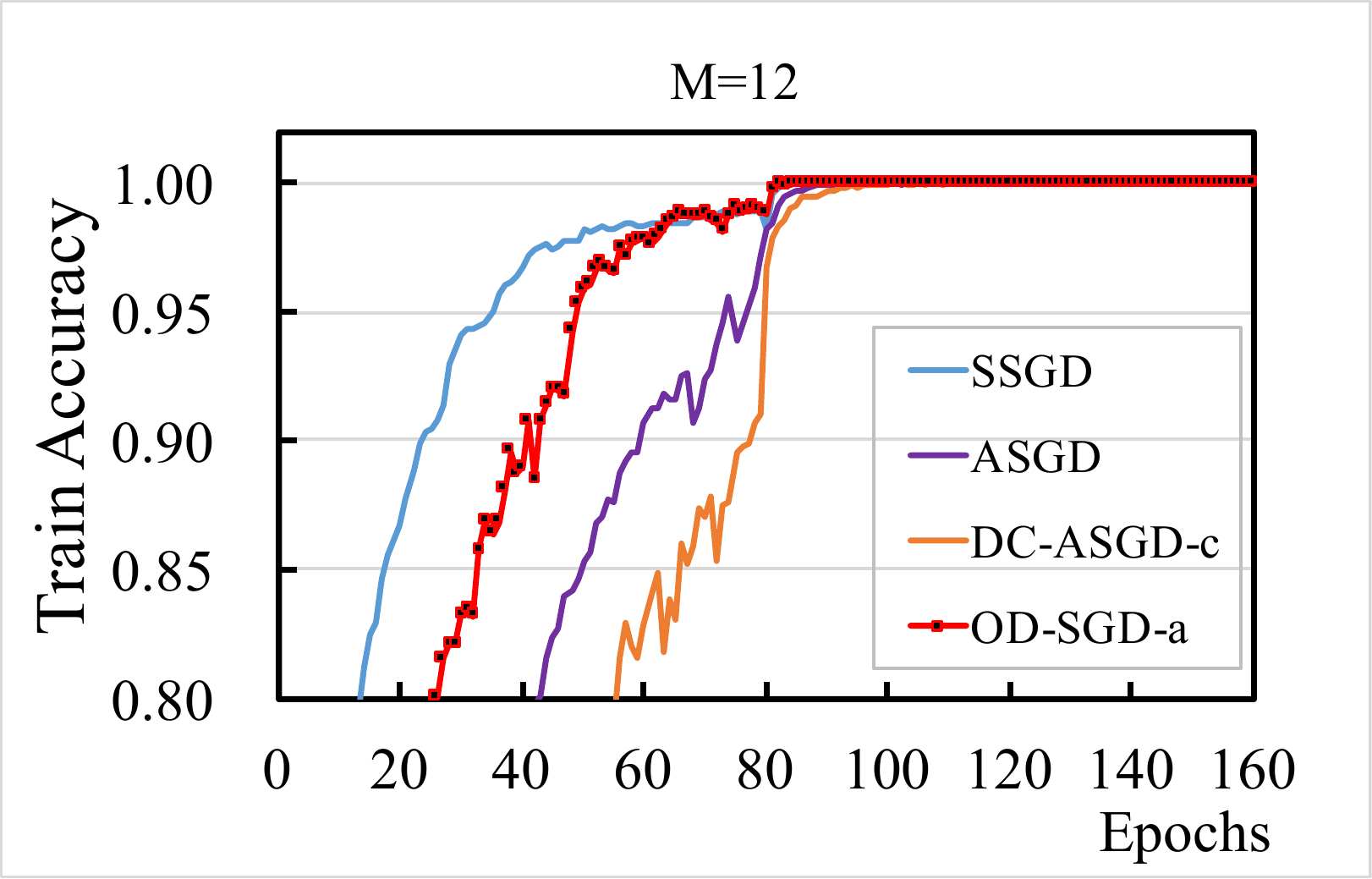}}
		\centerline{(h)}
	\end{minipage}
	\begin{minipage}{0.33\linewidth}
		\centerline{\includegraphics[width=4.8cm, height=3cm]{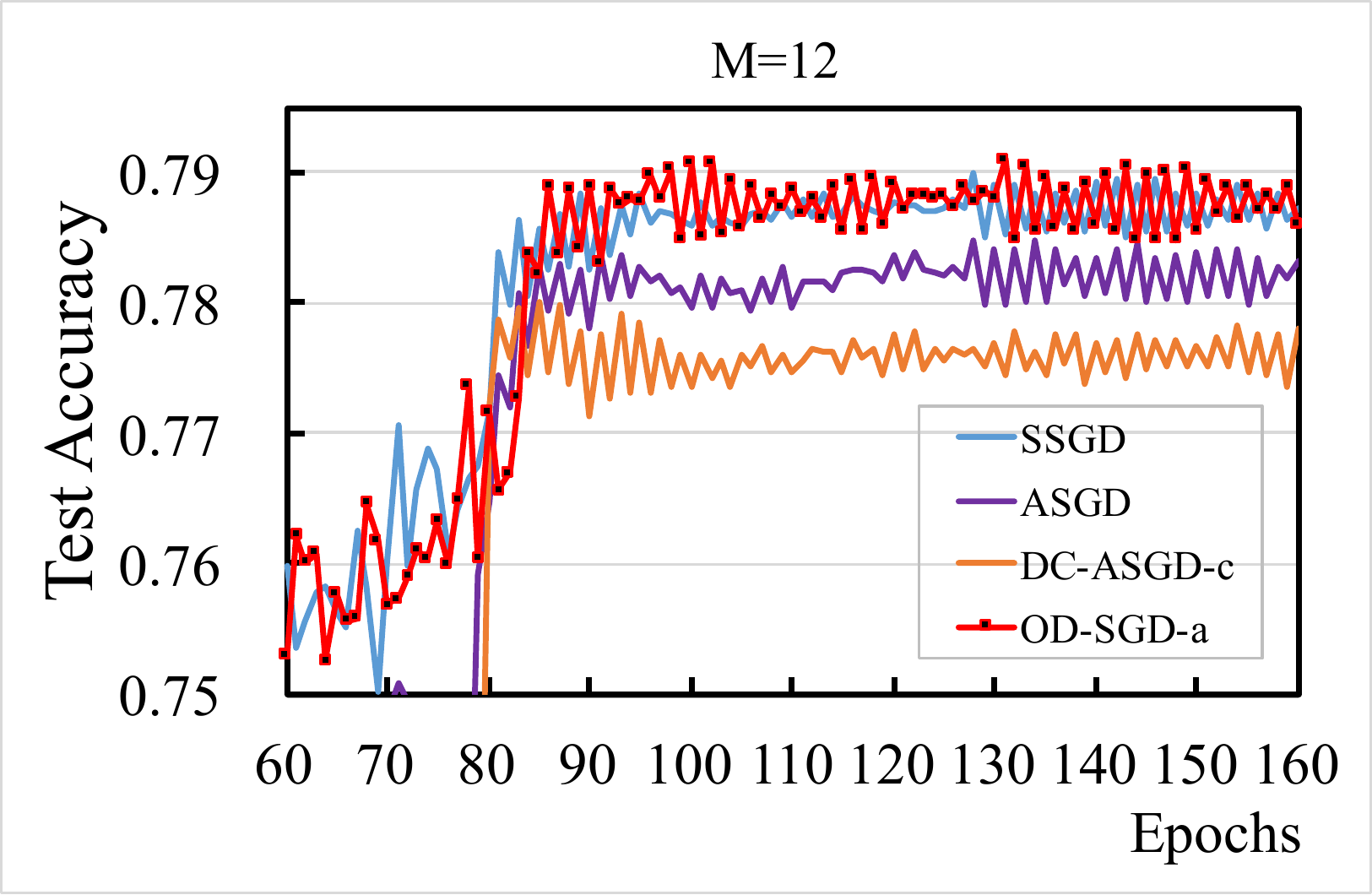}}
		\centerline{(i)}
	\end{minipage}
	\begin{minipage}{0.33\linewidth}
		\centerline{\includegraphics[width=4.8cm, height=3cm]{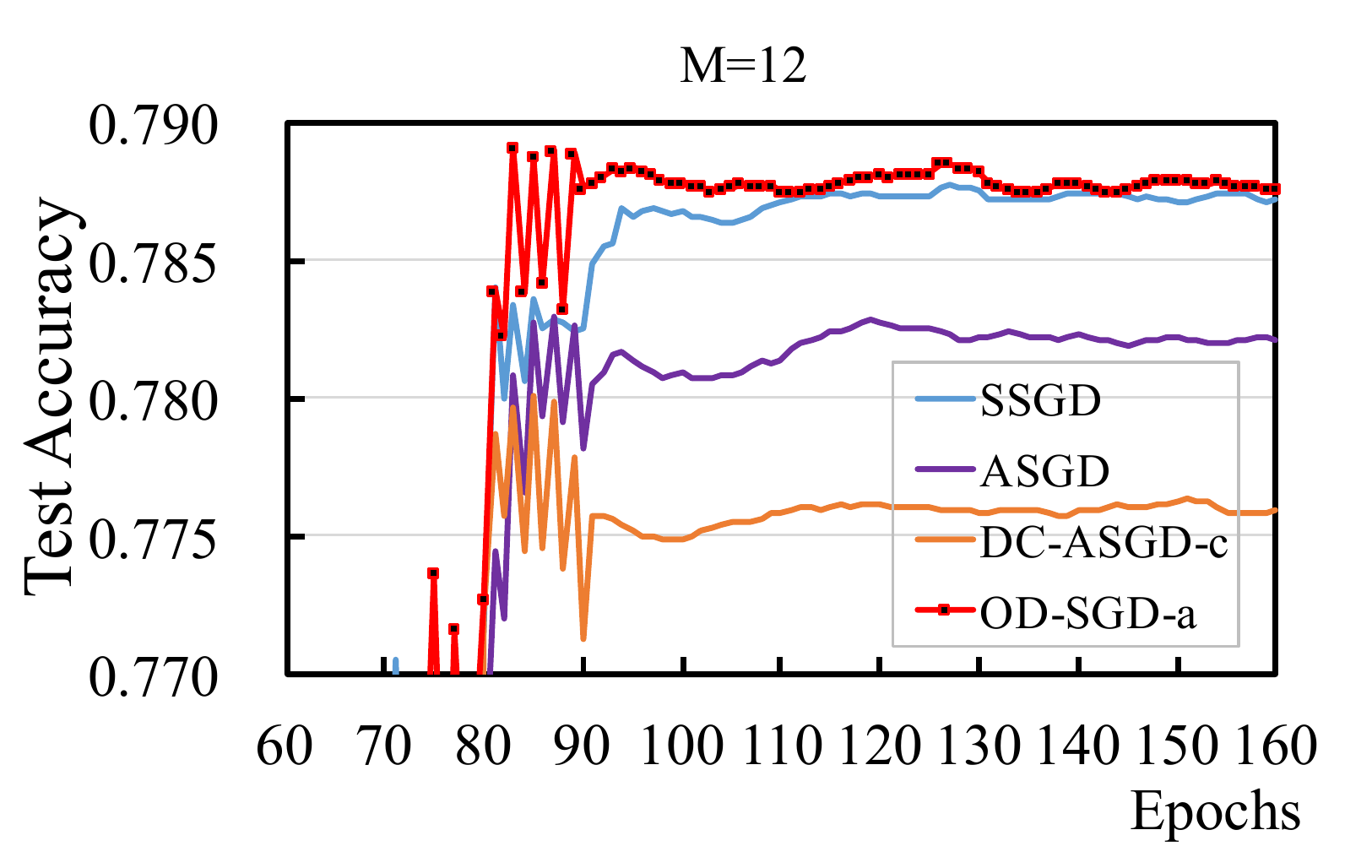}}
		\centerline{(j)}
	\end{minipage}	
	\caption{Train accuracy and test accuracy of ResNet18\_V2 under different epochs on CIFAR-10 dataset.}
	\label{fig:cifar-epoch}
\end{figure}
\vspace{1ex}

Firstly, we present the training accuracy and test accuracy under different epochs (Fig.~{\ref{fig:cifar-epoch}}). 
Subgraphs in the left column demonstrates the training accuracies of different clusters and we have the following observations. (1) All algorithms achieve 100\% training accuracies after training for about 80 epochs, although they experience various convergence curves in the first 80 epochs. (2) The gradient problem deteriorates as the scale of cluster goes up. When $M=4$, the ASGD, SSGD and OD-SGD-c have similar convergence curves, while when $M=12$, there exists obvious differences between their convergence curves. The OD-ASGD-c has the worst training curves in all three subgraphs and we attribute this to the improper hyper-parameters.  Nevertheless, the hyper-parameters involved in OD-SGD-c for local update operation are the same to those of DC-ASGD-c. For the sake of fairness, we conduct experiments on ResNet-20 nerual network later. 

Subgraphs in the middle column displays the test accuracies under different clusters. Convergence curves in subgraph $(b)$, $(e)$ and $(i)$ are based on the raw test accuracy data, from which we draw the following conclusions. (1) OD-SGD (OD-SGD-a and OD-SGD-c) can achieve similar or slightly better test accuracy when compared with SSGD. (2) A larger number of workers leads to more pronounced vibration of test accuracy. When $M=4$, the test accuracy curves of SSGD and OD-SGD-c are stable while the curves of ASGD and OD-SGD-c vibrate slightly. However, the accuracy curves of all algorithms vibrate significantly When $M=12$. (3) A larger cluster scale expand the test accuracy gap between different algorithm. Because a larger cluster size leads to worse gradient delay problem. It is noteworthy that when $M=12$, the test accuracy of OD-SGD-c is less than ASGD. Therefore, the compensation strategy based on ASGD may not be applicable when the batch size is large enough, extra optimization method is needed.  

\makeatletter\def\@captype{figure}\makeatother
\begin{figure}
	\begin{minipage}{0.33\linewidth}
		\centerline{\includegraphics[width=4.8cm, height=3cm]{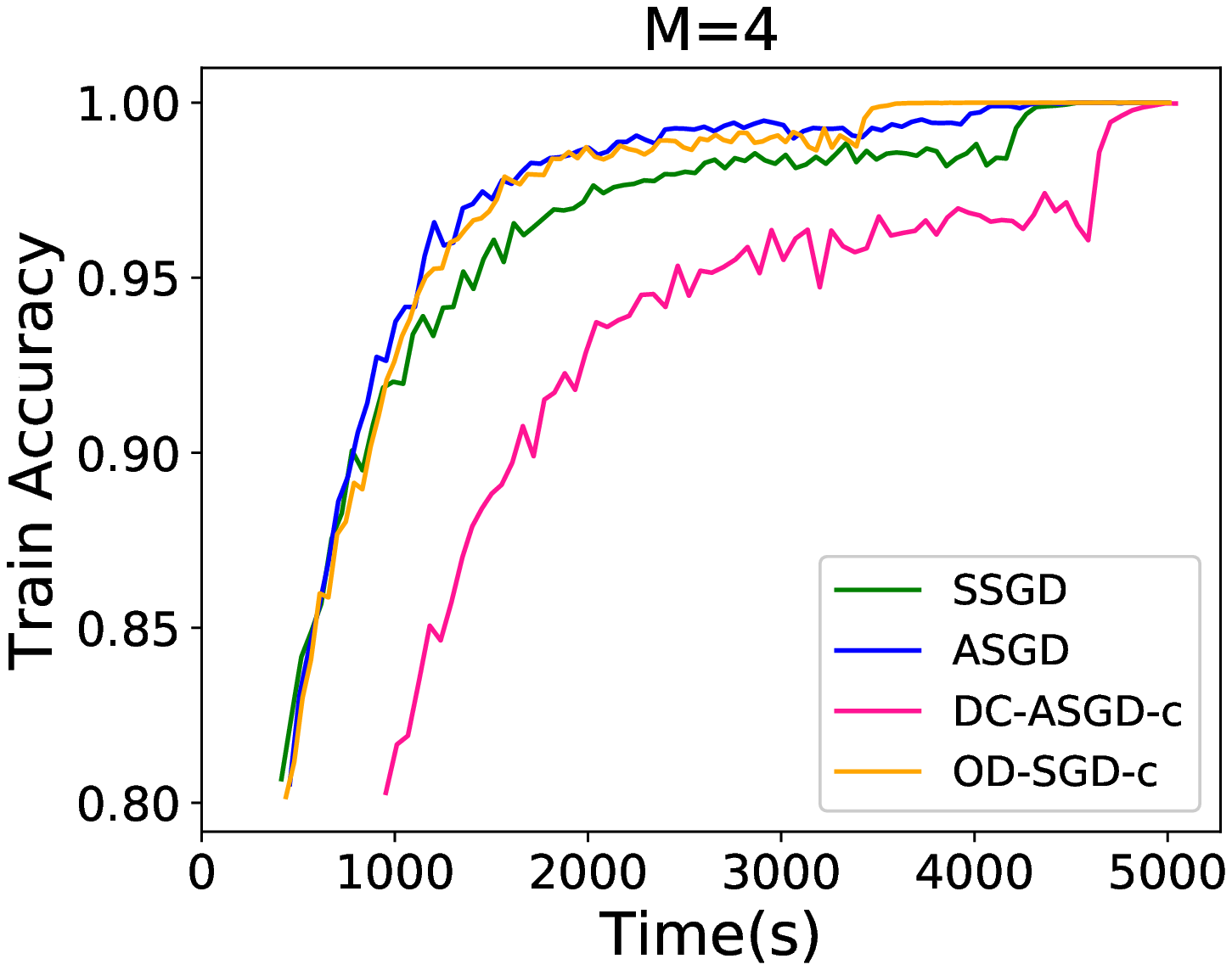}}
		\centerline{(a)}
	\end{minipage}
	\begin{minipage}{0.33\linewidth}
		\centerline{\includegraphics[width=4.8cm, height=3cm]{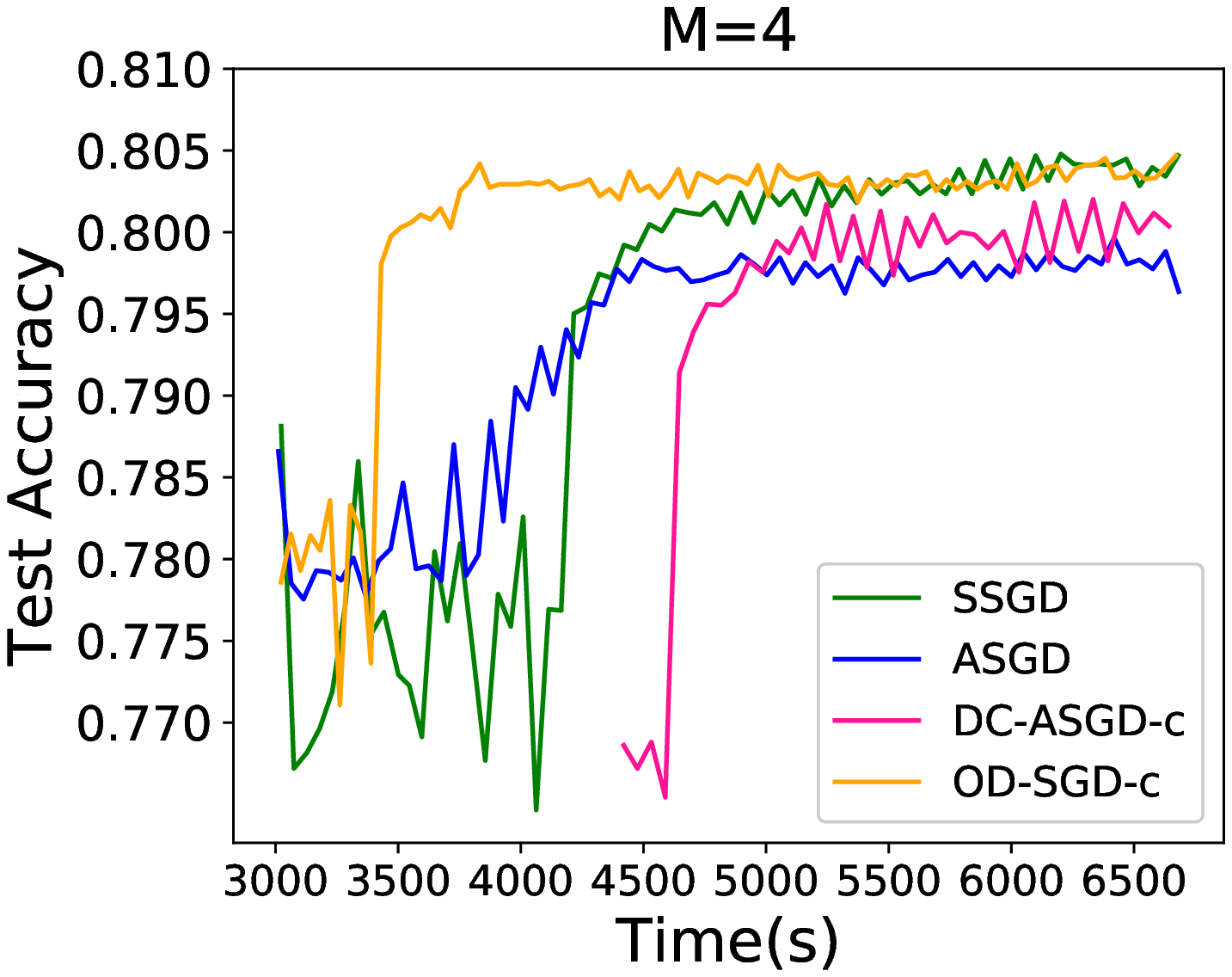}}
		\centerline{(b)}
	\end{minipage}
	\begin{minipage}{0.33\linewidth}
		\centerline{\includegraphics[width=4.8cm, height=3cm]{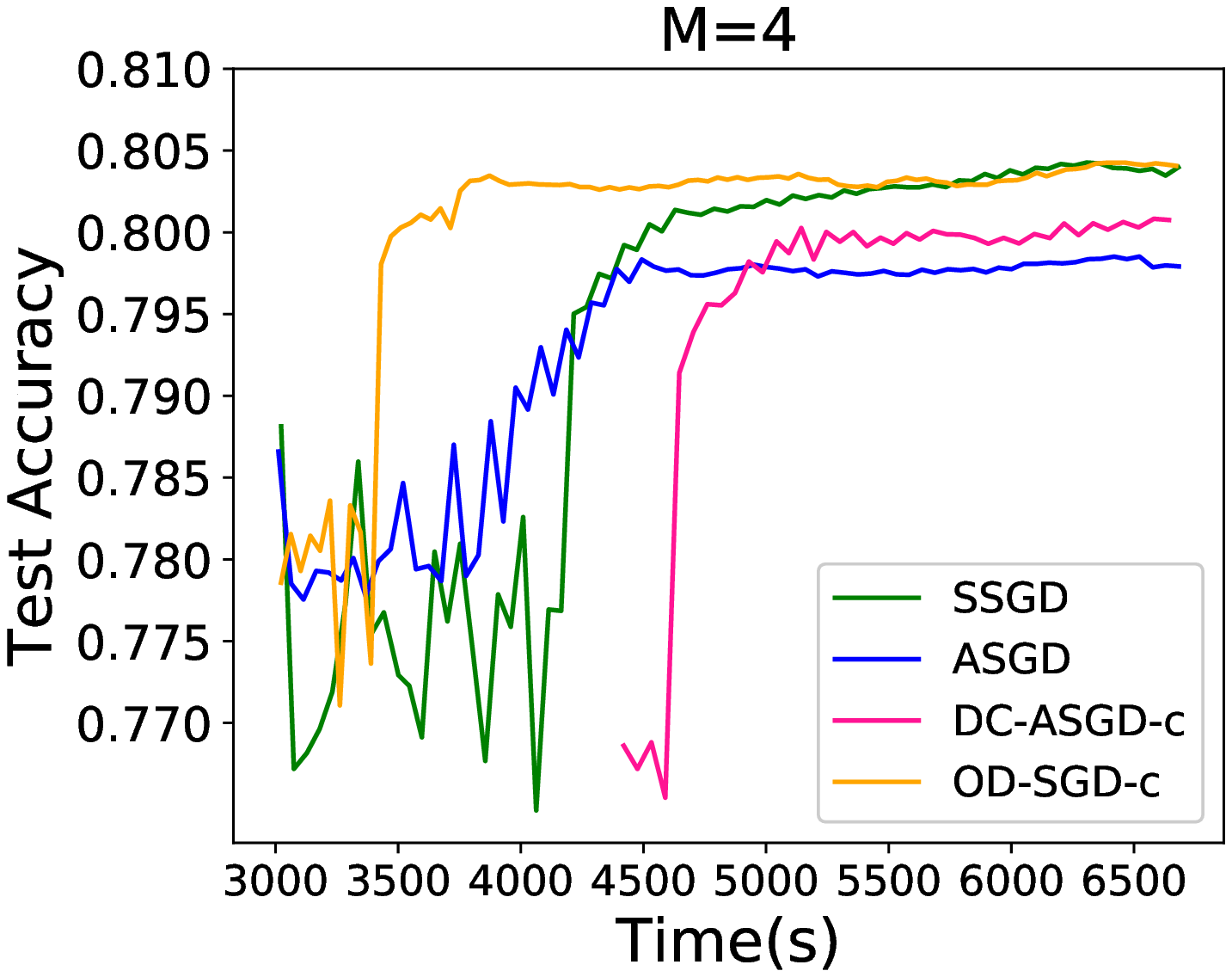}}
		\centerline{(c)}
	\end{minipage}
	\vfill
	\begin{minipage}{0.33\linewidth}
		\centerline{\includegraphics[width=4.8cm, height=3cm]{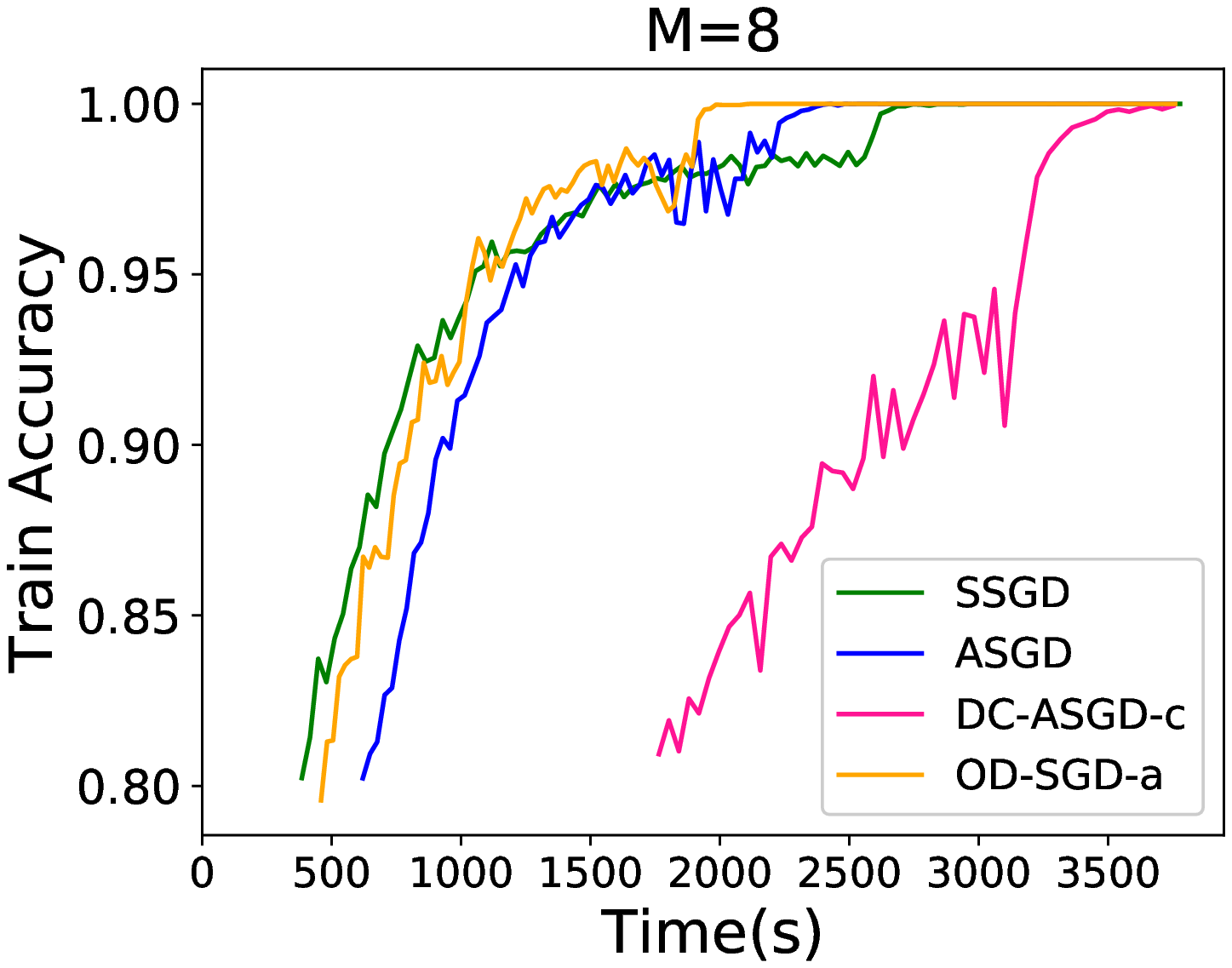}}
		\centerline{(d)}
	\end{minipage}
	\begin{minipage}{0.33\linewidth}
		\centerline{\includegraphics[width=4.8cm, height=3cm]{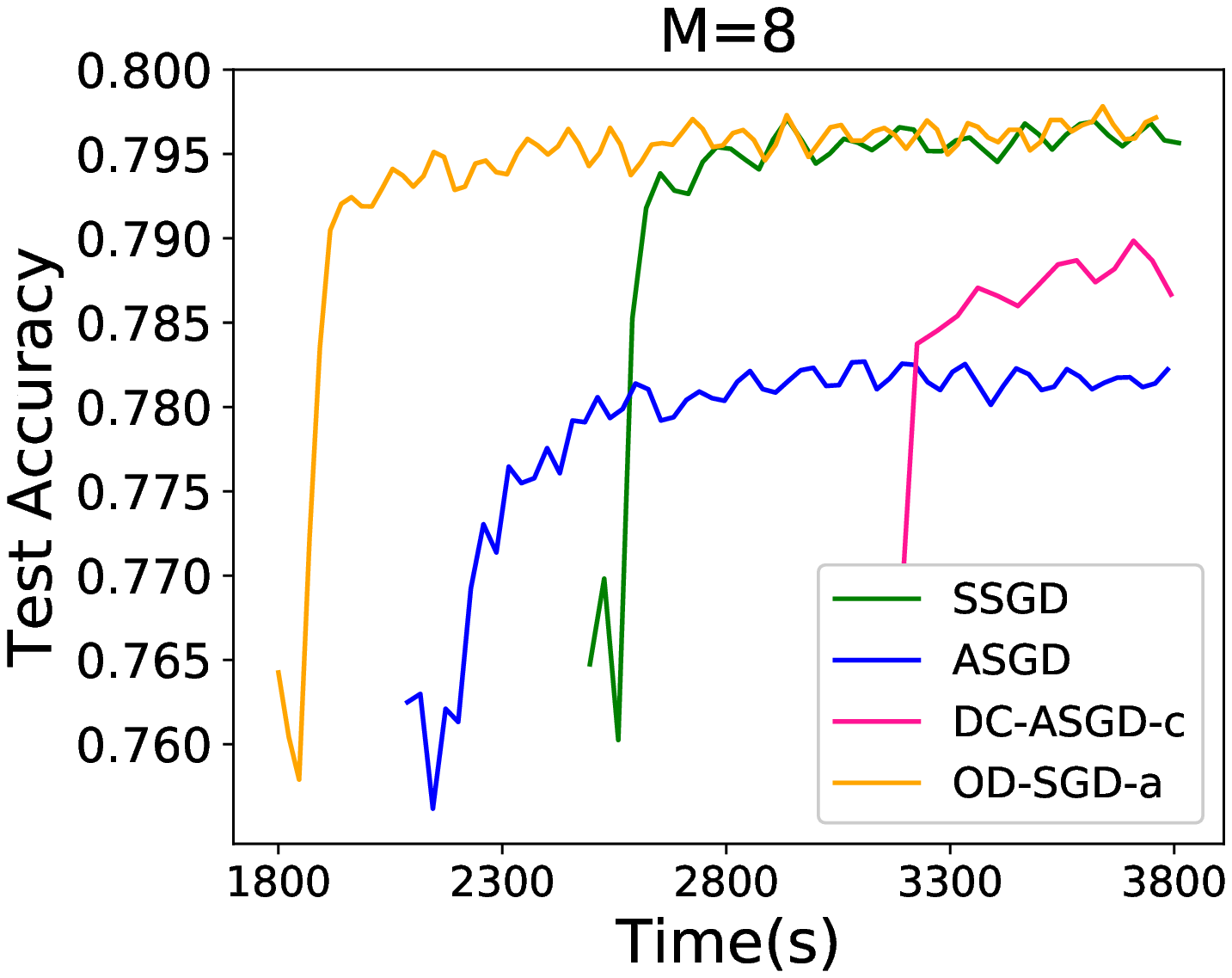}}
		\centerline{(e)}
	\end{minipage}
	\begin{minipage}{0.33\linewidth}
		\centerline{\includegraphics[width=4.8cm, height=3cm]{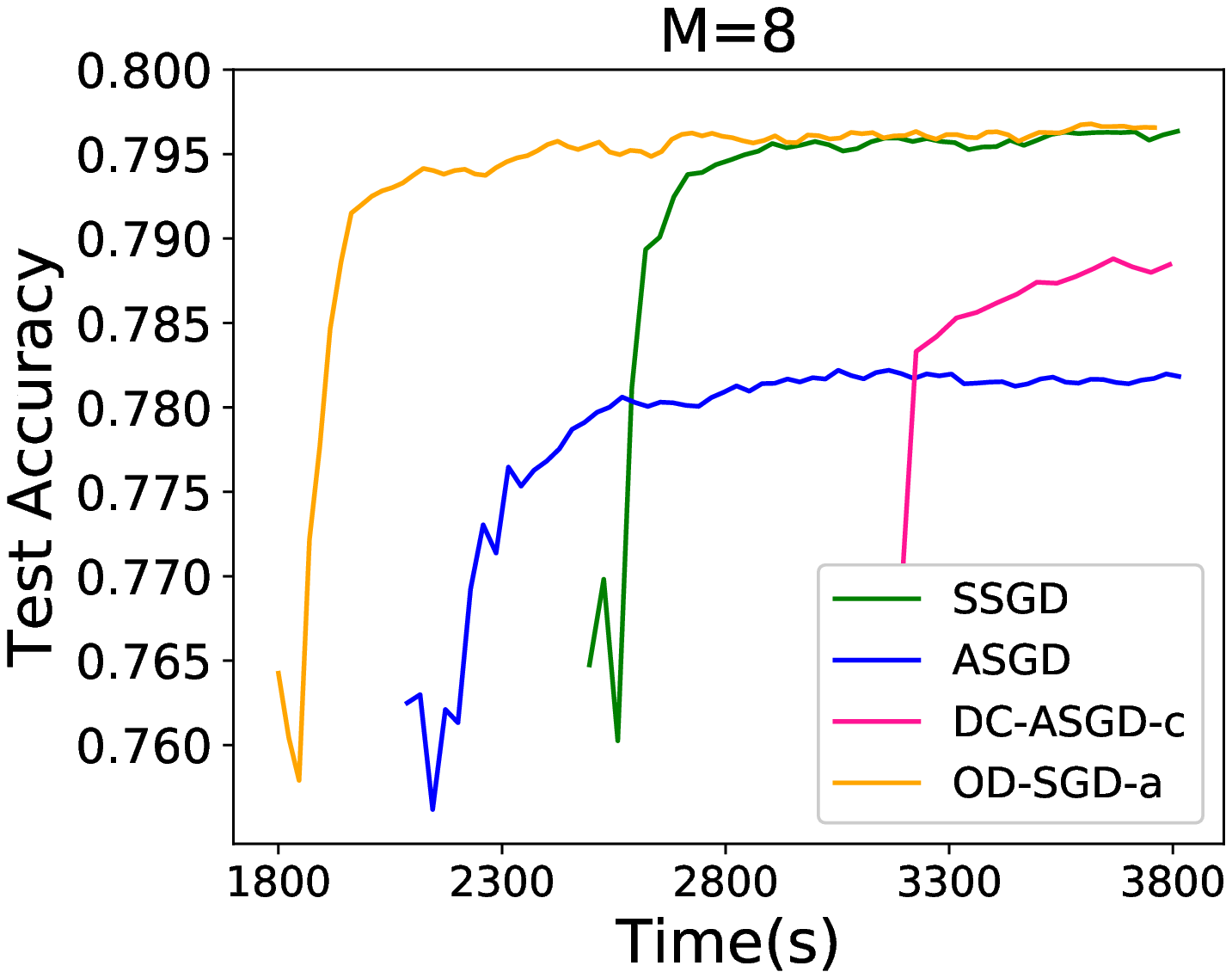}}
		\centerline{(f)}
	\end{minipage}	
	\vfill
	\begin{minipage}{0.33\linewidth}
		\centerline{\includegraphics[width=4.8cm, height=3cm]{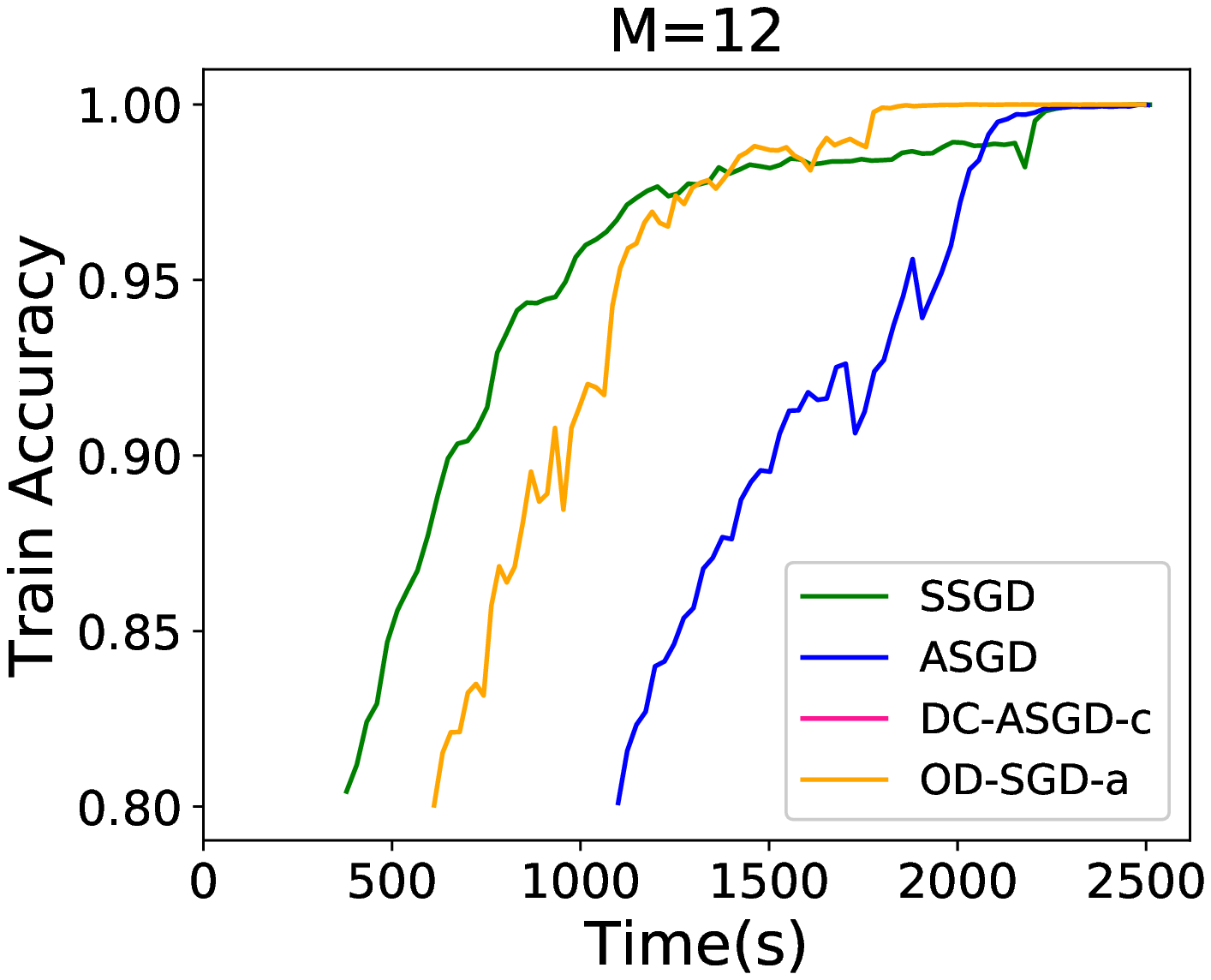}}
		\centerline{(h)}
	\end{minipage}
	\begin{minipage}{0.33\linewidth}
		\centerline{\includegraphics[width=4.8cm, height=3cm]{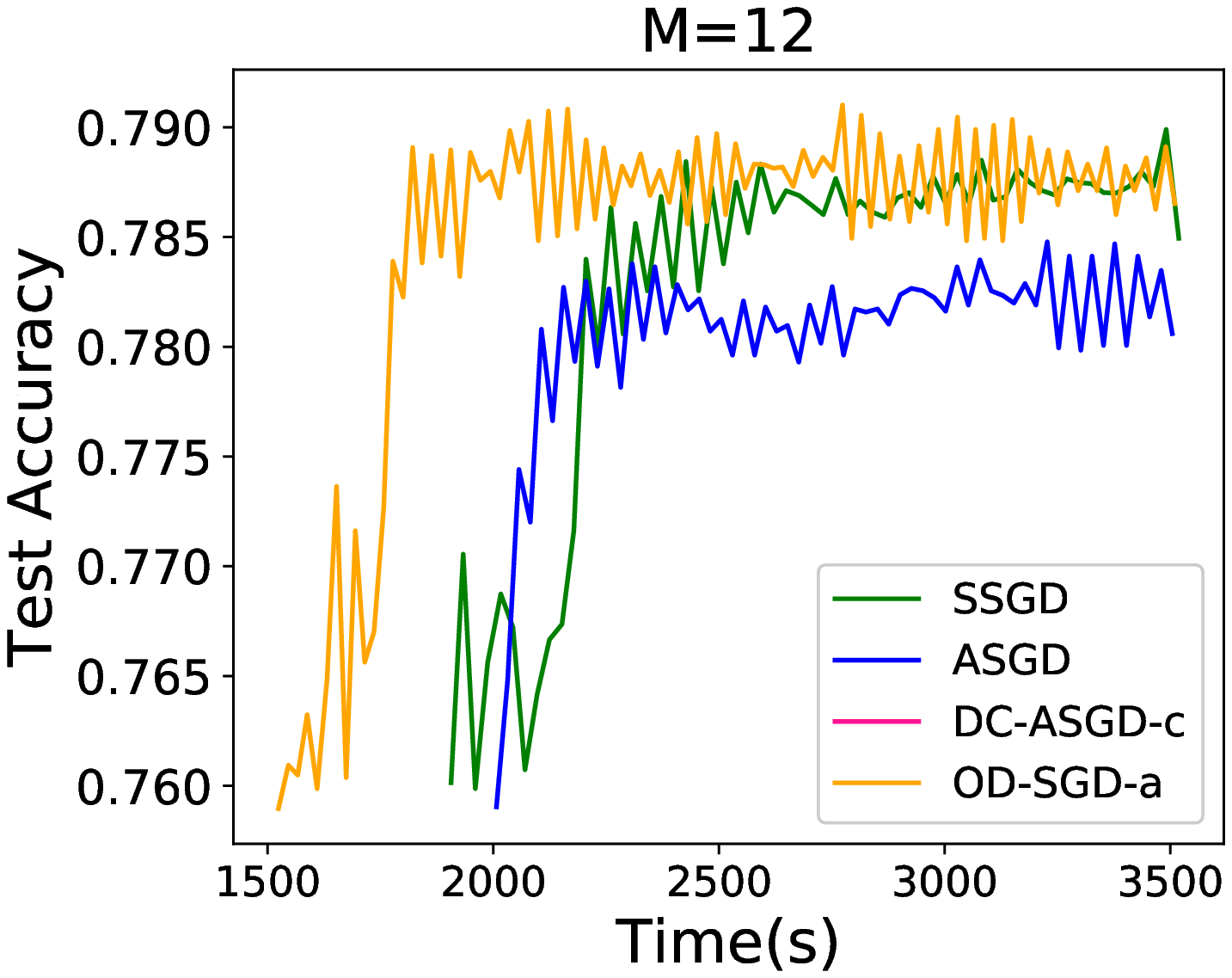}}
		\centerline{(i)}
	\end{minipage}
	\begin{minipage}{0.33\linewidth}
		\centerline{\includegraphics[width=4.8cm, height=3cm]{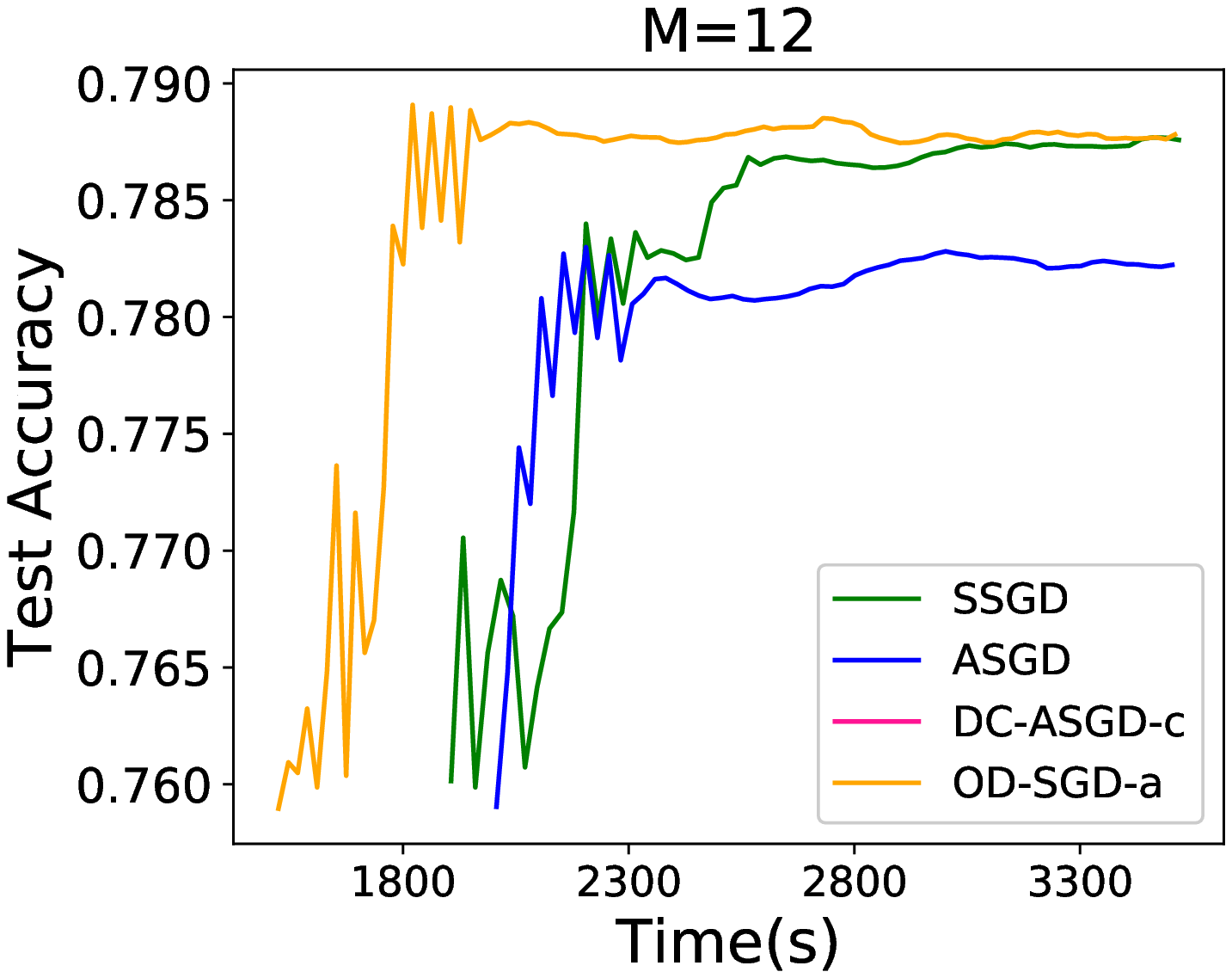}}
		\centerline{(j)}
	\end{minipage}	
	\caption{Train accuracy and test accuracy of ResNet18\_V2 w.r.t wallclock time on CIFAR-10 dataset}
	\label{fig:cifar-time}
\end{figure}

Subgraphs in the right column also illustrates the test accuracies. To get rid of the vibrations, we use moving average to smooth the data after the 91th epoch, with window size of 5. For instance, the test accuracy of the 92th epoch is the average value of the 90th, 91th,92th, 93th and 94th epochs. Curves of these subgraphs are based on the average test accuracies, which provide clearer trends throughout the training process. From these subgraphs we can apparently notice that OD-SGD is able to guarantee the test accuracy under different clusters, proving its effectiveness in distributed training.

Table~\ref{tab:cifar-avg} displays the test accuracy after smoothing the data with moving average, from which we have the following observations. (1) Test accuracies of all algorithms decrease as the cluster size goes up in general. Take SSGD as an example, when $M=4$, the test accuracy is 80.482\%, which decreases to 78.844\% when $M=12$. This phenomenon is reasonable, ASGD and DC-ASGD suffers from the gradient delay problem which deteriorates as the cluster size goes up. The mini-batch size of each local worker keeps unchanged as the cluster size increase, and larger cluster size will lead to larger batch size. The enlarged batch size imposes negative influences on the test accuracy of neural network, which accounts for the accuracy decrease of SSGD and OD-SGD. (2) Training on a single local worker achieves the best test accuracy. (3) OD-SGD can achieve similar test accuracy when compared with SSGD. When $M=4$, its accuracy is 0.056\% lower than SSGD. Nevertheless, its test accuracy is higher than that of SSGD when $M=8$ and $M=12$.

The convergence speed of different algorithms is shown in Fig.~{\ref{fig:cifar-time}}. From these subgraphs we have the following observations. (1) OD-SGD-c and OD-SGD-a achieve the best training performance, which run faster than ASGD. (2) Although the convergence point of ASGD is lower than that of SSGD, its convergence speed is faster than that of SSGD. (3) The training speed of SSGD is slowed down by its global barrier, but it still runs faster than DC-ASGD-c. The additional computational overhead brought in by DC-ASGD-c significantly affect its throughput in the training process. When $M=12$, the training speed of OD-SGD-c is too slow to show the training results in the subgraphs.
(4) Our proposed OD-SGD achieves both good accuracy and speed. On the one hand, its convergence speed is faster than that of ASGD. On the other hand, its convergence point is as good as, or even slightly better than that of SSGD.

\makeatletter\def\@captype{figure}\makeatother
\begin{figure}
	\begin{minipage}{0.42\linewidth}
		\centerline{\includegraphics[width=6.4cm, height=4cm]{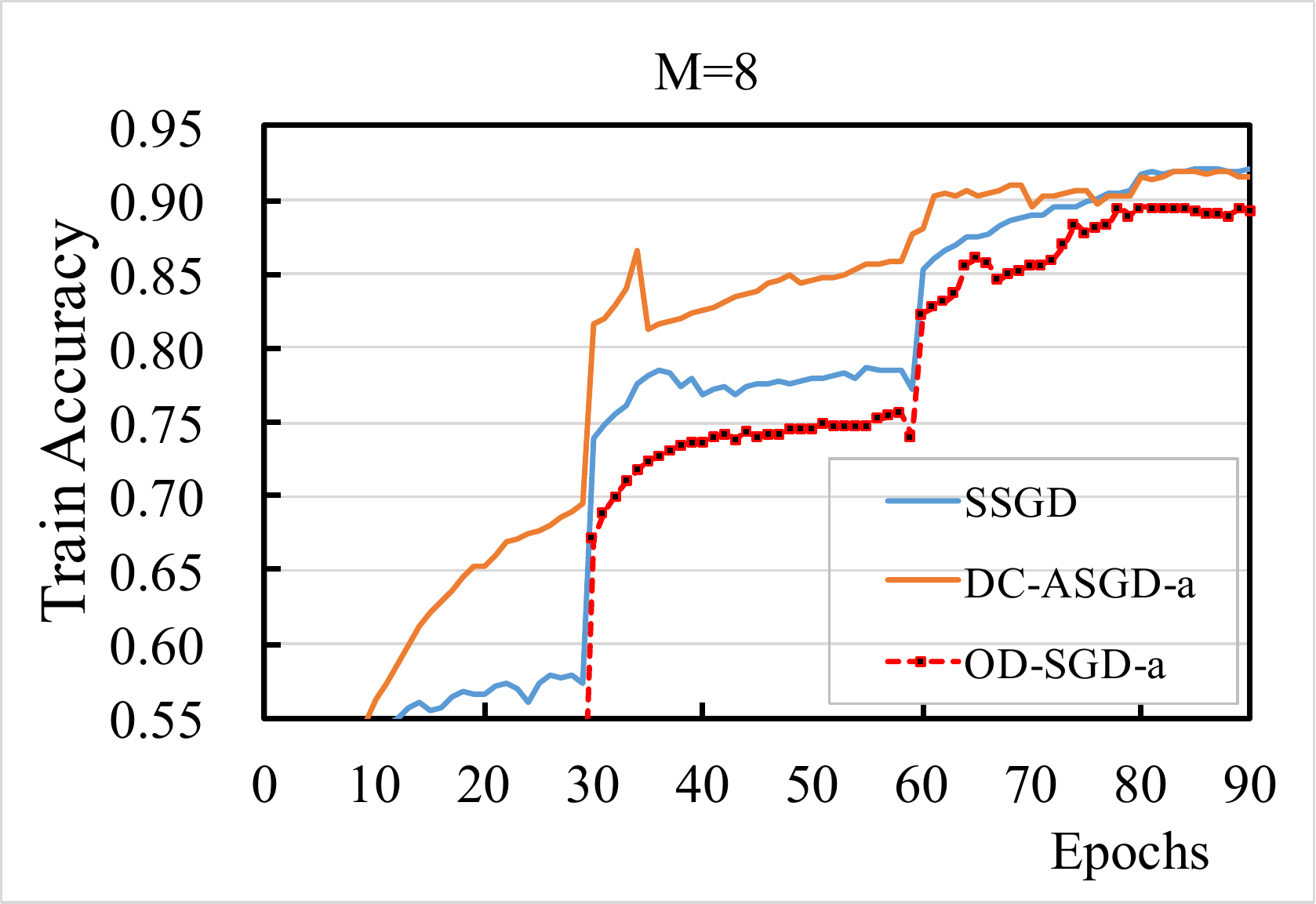}}
		\centerline{(a)}
	\end{minipage}
	\begin{minipage}{0.42\linewidth}
		\centerline{\includegraphics[width=6.4cm, height=4cm]{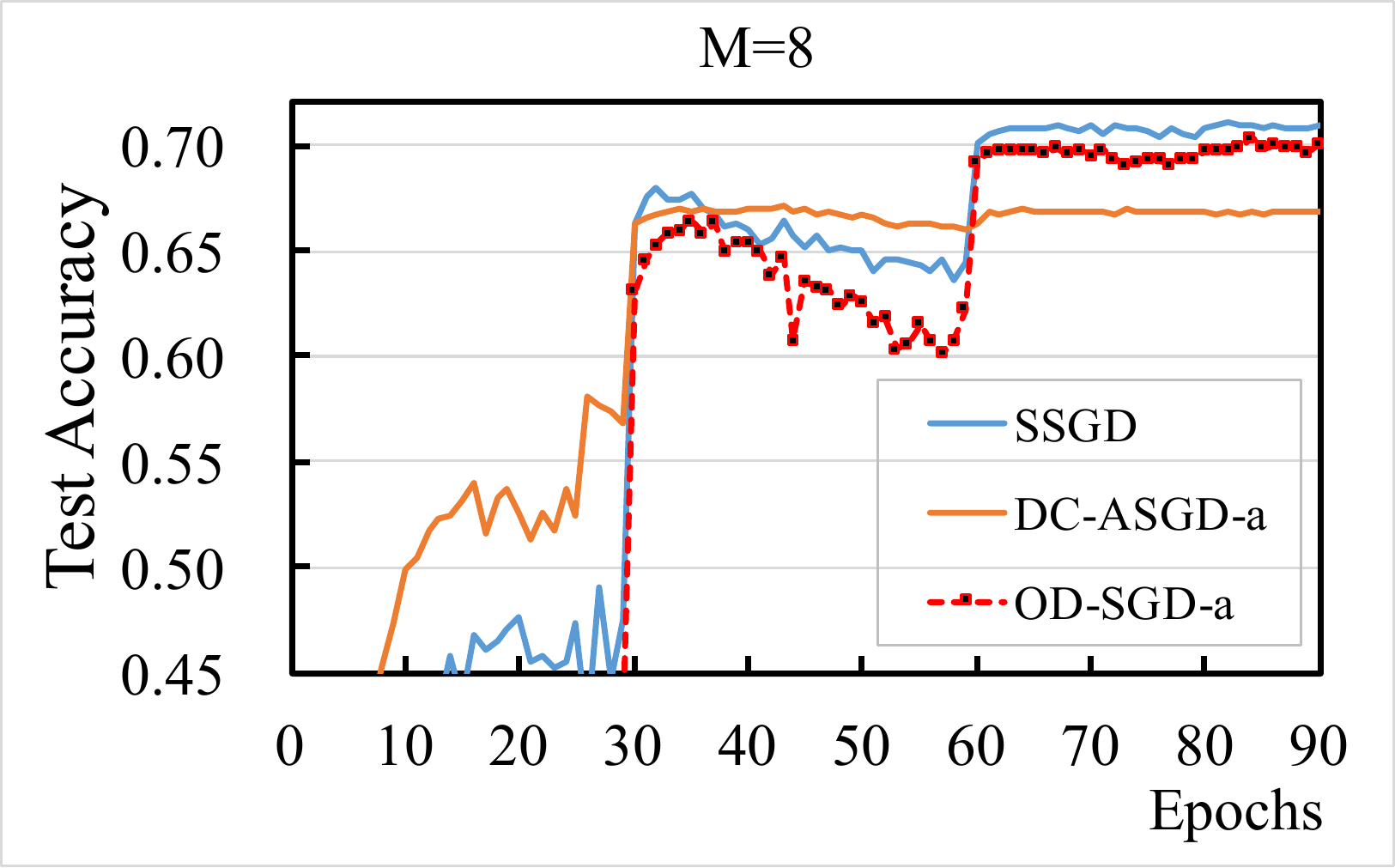}}
		\centerline{(b)}
	\end{minipage}
	\caption{Train accuracy and test accuracy of ResNet-50 w.r.t epochs on ImageNet}
	\label{fig:imagenet-e}
\end{figure}

\makeatletter\def\@captype{figure}\makeatother
\begin{figure}[b]
	\begin{minipage}{0.42\linewidth}
		\centerline{\includegraphics[width=6.4cm, height=4cm]{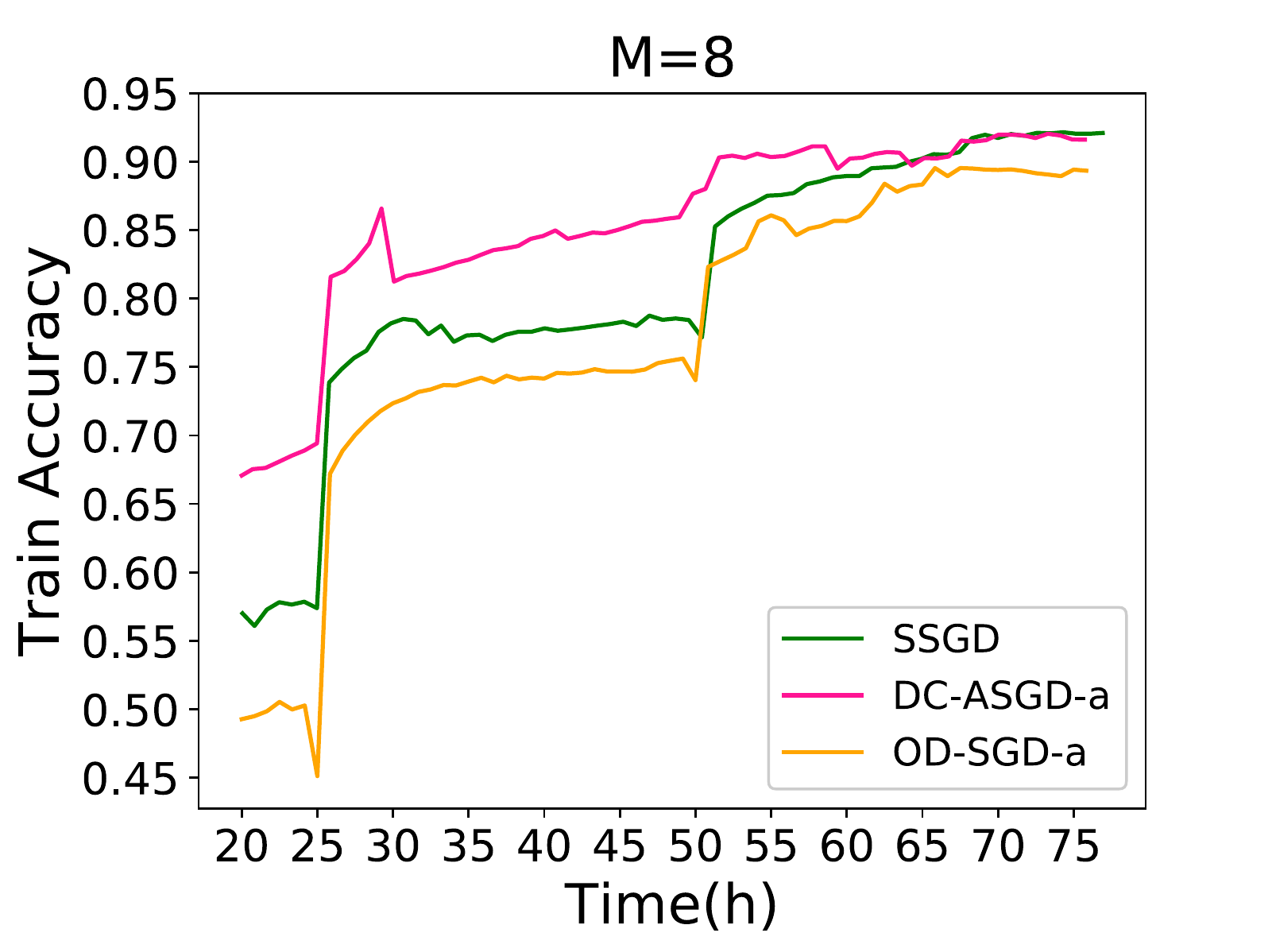}}
		\centerline{(a)}
	\end{minipage}
	\begin{minipage}{0.42\linewidth}
		\centerline{\includegraphics[width=6.4cm, height=4cm]{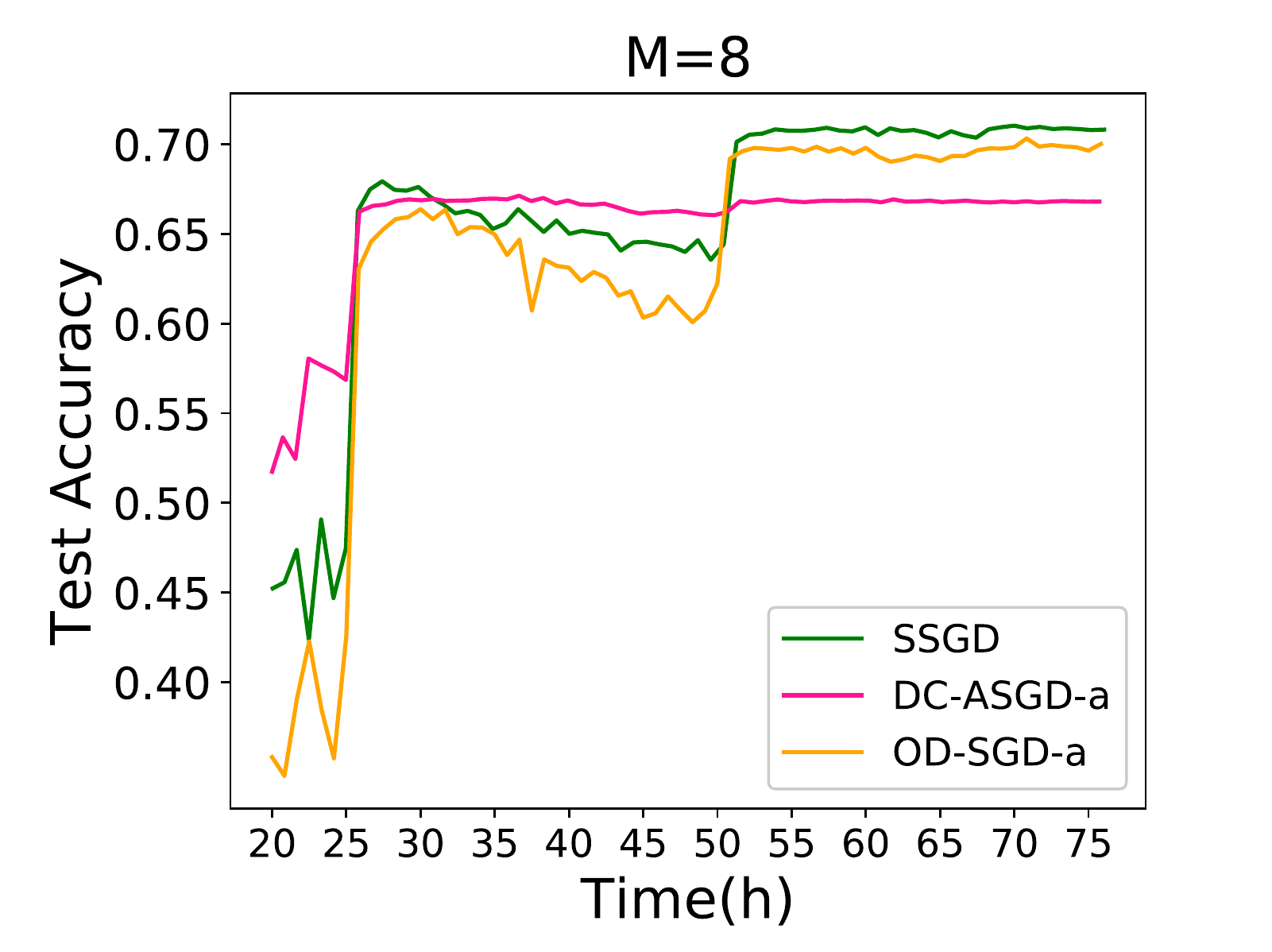}}
		\centerline{(b)}
	\end{minipage}
	\caption{Train accuracy and test accuracy of ResNet-50 w.r.t wallclock time on ImageNet}
	\label{fig:imagenet-t}
\end{figure}

\subsection{Experimental Results on ImageNet}
For the sake of verifying our algorithm on large scale neural network, we conduct experiments on ImageNet dataset using ResNet-50 \cite{he2016deep}. In this large scale experiment, we evaluate the performance of SSGD, DC-ASGD-a and OD-SGD-a. For all algorithms in this experiment, we conduct the training task for 90 epochs (without data augmentation) on 8 workers which are interconnected by InfiniBand network, with a mini-batch size of 256, and the learning rate is reduced by 10 times after training for 30, 60 and 80 epochs according to the practice in \cite{goyal2017accurate}.
The initial learning rate for SSGD is $0.1$ and the momentum value is $0.9$, considering the batch size (2048) and linear scaling rule, we adopt the linear WP strategy to adjust the learning rate to $0.8$ after training for 5 epochs. The hyper parameters of DC-ASGD-a follows the configure in \cite{dcasgd-2017} ($\eta=0.1$, $\lambda=2$, $m=0$), the momentum value is $0$. For OD-SGD-a, the global update in the parameter server follows the configuration of SSGD while local updates in the workers follow the configuration of OD-AGSD-a.

Table~{\ref{tab:imagenet}} displays the test accuracy while Fig.~{\ref{fig:imagenet-e}} presents the train accuracy and test accuracy on ImageNet with regard to the epochs, from which we have the following observations: (1) After training for the same number of epochs, DC-ASGD-a always achieves higher training accuracy and its final value is similar to that of SSGD. On the contrary, training accuracy of OD-SGD-a is lower throughout the training process and the final accuracy is around 90\%. (2) SSGD can obtain the best test accuracy (70.773\%) and the accuracy of OD-SGD-a (69.847\%) is slightly lower. Actually, we do not spare efforts to find the suitable hyper parameters for OD-SGD-a and a suitable configuration may bring about similar accuracy to SSGD. Nevertheless, DC-ASGD-a (66.772\%) is not able to get a good accuracy and the accuracy increase at epoch 60 is significantly lower. We attribute this to the large batch size and the configuration in \cite{dcasgd-2017} is not applicable to large batch size (512 vs 2048) here. (3) Curves of SSGD and OD-SGD-a have similar patterns, proving that OD-SGD-a utilizes the feature of SSGD. 

The convergence speed is illustrated in Fig.~{\ref{fig:imagenet-t}}, from which we can notice the  speed of those algorithms are similar and this result is reasonable. The limited training speed mainly results from the low computing power of K80 GPU, each local worker can just handle about 50 samples per second. In addition, SSGD contains a synchronous barrier and DC-ASGD-a calls for frequent additional computation operations in the parameter server node. 
As to OD-SGD-a, 
the introduced local computation overheads in worker nodes prolong the communication overhead more or less. 
All these algorithms take more than 75 hours to complete the training process.

The long training time results from the low computation capability of K80 GPU and we conduct some  supplementary experiments on a V100 GPU cluster. Because of the limited resource, we are only available to 4 worker nodes for parallel training. In the supplementary experiments, we adopt the data augmentation due to the high computing power of V100 GPU and the polynomial (power=2) decay LR policy is used to adjust the learning rate. We compare the training performance of SSGD and OD-SGD, the mini-batch size of each node is 512 (V100 GPU has larger memory, 16G) with momentum=0.9 and weight decay=0.0001, the WP period for increasing the learning rate is 5 epochs. 

\begin{figure}[t]
	\centering
	\begin{minipage}{0.48\textwidth}
		\centering
		\includegraphics[width=5.6cm, height=3.5cm]{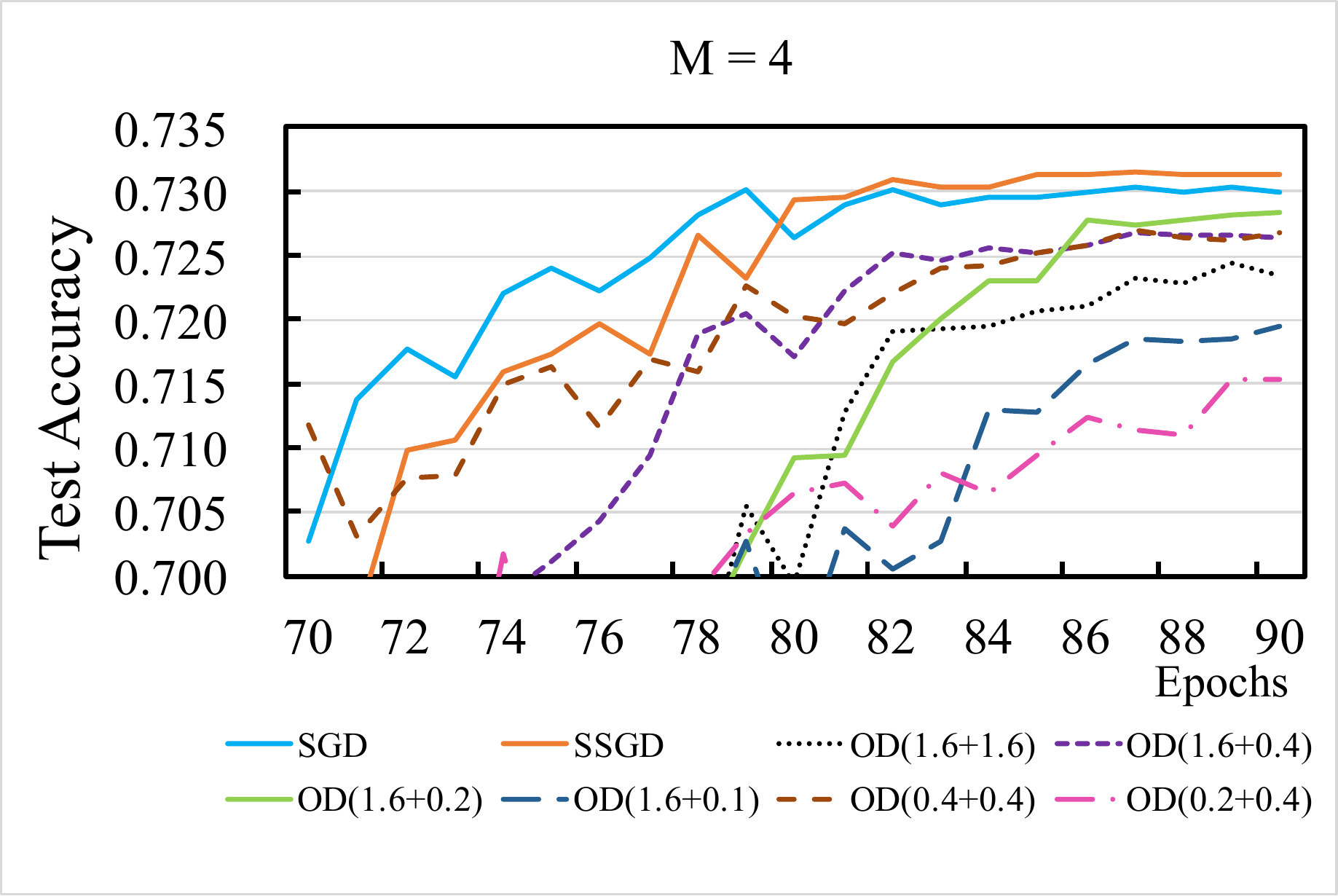}
		\caption{Test accuracy of ResNet-50 w.r.t epochs on ImageNet. \textit{SGD} means training on a single node, \textbf{OD(1.6+0.4)} means the global learning rate is 1.6 and the local learning rate for each worker is 0.4.}
		\label{fig:epoch-train} 
	\end{minipage}
	\begin{minipage}{0.48\textwidth}
		\centering
		\includegraphics[width=5.4cm, height=3.5cm]{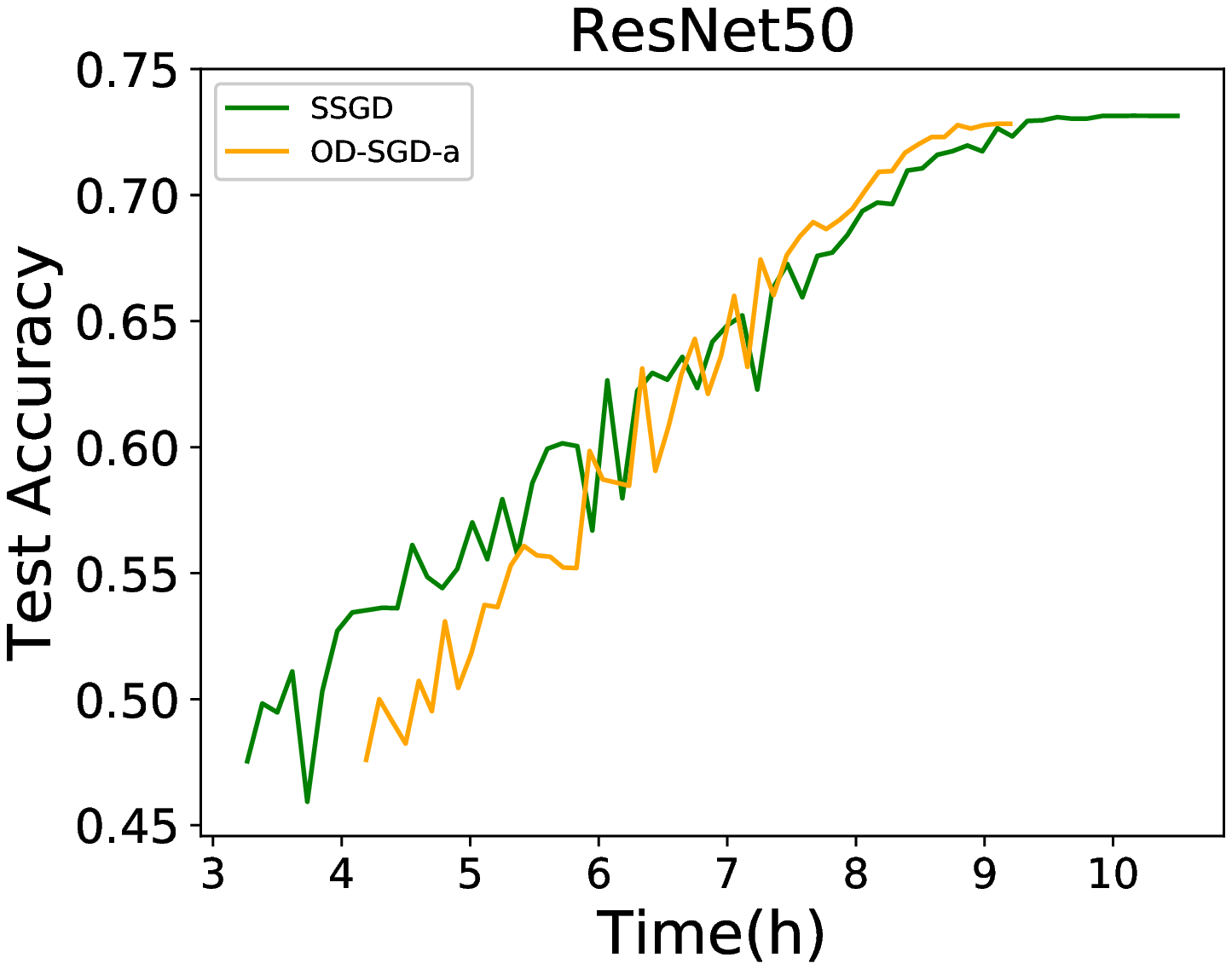}
		\caption{Test accuracy w.r.t wallclock time on ImageNet, the mini-batch size is 512.}
		\label{fig:time-train} 
	\end{minipage}
\end{figure}

\vspace{3ex}
\makeatletter\def\@captype{table}\makeatother
\begin{minipage}{0.3\textwidth}
	\caption{Test accuracy on ImageNet with ResNet-50 (K80 GPU cluster).}
	\label{tab:imagenet}
	\begin{center}
		\begin{tabular}{c|c}
			\hline
			\multirow{2}*{Algorithms}  & Test  \\
			& Accuracy (\%) \\
			\hline
			SSGD & \color{red}{70.773} \\
			\hline
			DC-ASGD-a & 66.772 \\
			\hline	
			OD-SGD-a & 69.847\\
			\hline
		\end{tabular}
	\end{center}
\end{minipage}	
\hspace{2ex}
\makeatletter\def\@captype{table}\makeatother
\begin{minipage}{0.64\textwidth}
	\centering
	\caption{Test accuracy on ImageNet with ResNet-50 (V100 GPU cluster).}
	\label{tab:imagenet-tune}
	\begin{center}
		\begin{tabular}{c|c|c|c|c}
			\hline
			Algorithm & SGD  &SSGD & OD(1.6+1.6) & OD(1.6+0.4)\\ 
			\hline
			Accuracy(\%) &73.041 & \color{red}{73.145}& 72.439 &72.671 \\
			\hline
			Algorithm & OD(1.6+0.2) & OD(1.6+0.1) & OD(0.4+0.4) &OD(0.2+0.4) \\
			\hline
			Accuracy(\%)& \textbf{72.829} & 71.952 & 72.675 & 71.543\\
			\hline
		\end{tabular}
	\end{center}
\end{minipage}	
\vspace{3ex}

Fig.~\ref{fig:epoch-train} demonstrates the test accuracy of SGD, SSGD and OD-SGD-a. The accuracy of sequential SGD is the baseline and its learning rate is 0.4 with no WP. The initial learning rate of SSGD is 1.6 and we present accuracies of OD-SGD-a under different initial learning rates, as shown in Table~\ref{tab:imagenet-tune}. From Fig.~\ref{fig:epoch-train} we have the following observations: (1) When M = 4, the test accuracy for SSGD is 73.145\%, which is better than sequential SGD 73.041\%. This test performance lift might arise from the regularization effect brought by the variance introduced by parallel training. (2) OD-SGD-a achieves the best test accuracy when the global learning rate is 1.6 and local learning rate is 0.2, which is about 0.316\% and 0.212\% lower than that of SSGD and SGD respectively. (3) It is necessary to search for the proper hyper-parameters like learning rate for OD-SGD via grid search, this figure shows that improper learning rates will lead to obvious accuracy drop. However, it is very time consuming for large-scale neural network to search for the best configuration. In addition to learning rate, the hyper-parameters of DC-ASGD-a should also be taken into account. Fig.~\ref{fig:time-train} displays the convergence speed of SSGD and OD-SGD-a on the V100 GPU cluster. The sequential SGD takes about 35.6 hours to complete the training task and the convergence curve is not shown in this figure. The training time of SSGD is 10.5 hours while OD-SGD-a takes aboout 9.2 hours to accomplish the training workload, which achieves 12.38\% improvement in training speed with only 0.317\% convergence accuracy drop. Therefore, OD-SGD can be used for training models of large-scale.

\makeatletter\def\@captype{figure}\makeatother
\begin{minipage}{0.48\linewidth}
	\centerline{\includegraphics[width=6cm, height=3.8cm]{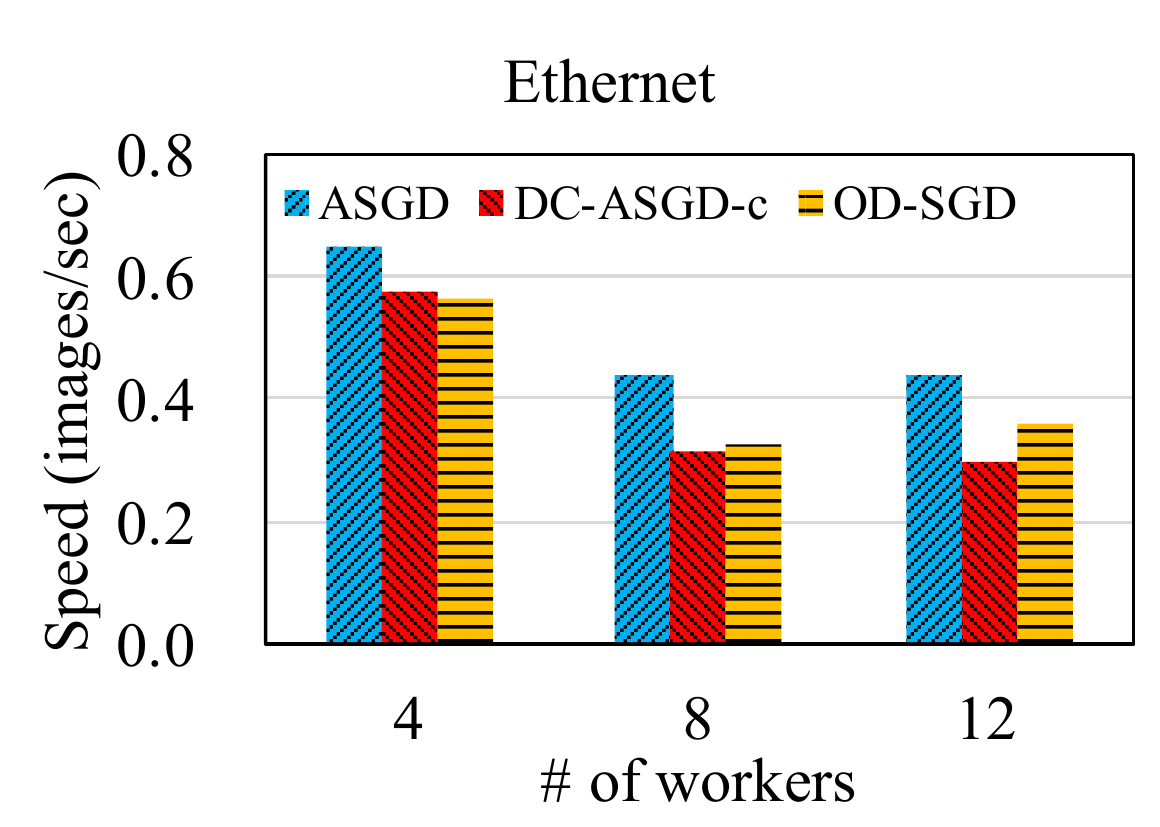}}
	\caption{Performance improvement with Ethernet}
	\label{fig:eth-speed}
\end{minipage}
\begin{minipage}{0.48\linewidth}
	\centerline{\includegraphics[width=6cm, height=3.8cm]{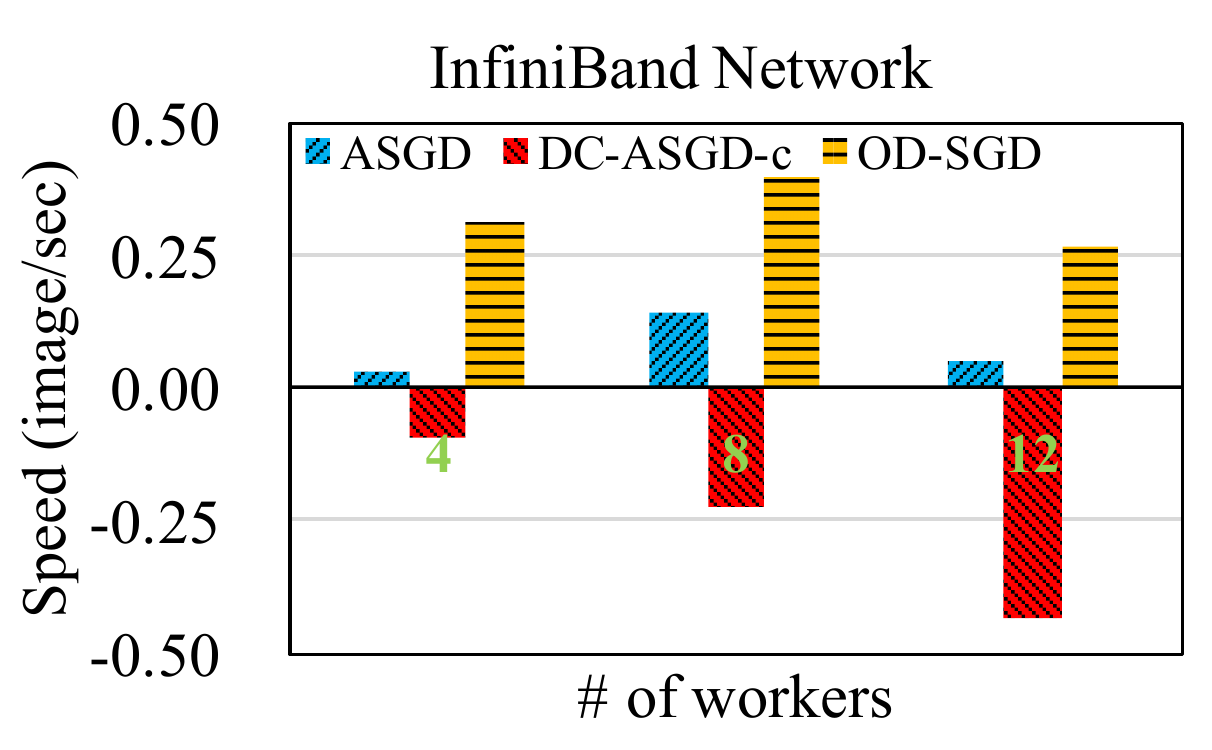}}
	\caption{Performance improvement with IB network}
	\label{fig:ib-speed}
\end{minipage}

\subsection{Training Performance under Different Network Bandwidths and Cluster Sizes}

In this section, we evaluate the impacts of network bandwidth and cluster scale on training speed.
Network bandwidth and cluster size are two main factors that affect the communication overhead in distributed training process. Therefore, we are analyzing the influences of communication overhead on training performance.

We conduct the training task of ResNet18\_V2 with CIFAR-10 again on clusters with different scales,  while the communication process is completed through Ethernet instead of InfiniBand network. For the sake of saving the expenditure of renting machines, we do not run training task to the end when the Ethernet is used because of the slow training speed.
Table~\ref{tab:cifar-speed-com} presents the throughputs of all algorithms with various number of worker nodes. The throughput of SSGD is chosen as the baseline and we evaluate the performance improvement according to Equation~\ref{eqn:speed-growth}. Fig.~{\ref{fig:eth-speed}} and Fig.~{\ref{fig:ib-speed}} illustrate the improvement ratio under Ethernet and InfiniBand network. From the table and figures, we have the following observations. (1) The throughput of each local worker decreases apparently as the cluster size goes up. This results from the increase of communication overhead, leading to growth of overhead for each iteration. (2) OD-SGD achieves training performance improvements with both Ethernet or InfiniBand network. (3) When the Ethernet is used for communication, ASGD has the best performance and OD-SGD is slightly better than DC-ASGD when $M=12$. (4) When InfiniBand network is applied, the performance improvement ratio of OD-SGD is significantly higher than ASGD and OD-SGD. Furthermore, the DC-ASGD imposes negative impacts on the training speed and a larger cluster size leads to worse performance. Because the application of InfiniBand network significantly decrease the communication overhead and the additional computation overhead introduced by OD-SGD makes the bottleneck. The parameter server needs to conduct more computation operations with a larger cluster size.

\begin{equation}
\label{eqn:speed-growth}
GR\_Rate_{od-sgd} = \frac{Speed_{od-sgd} - Speed_{ssgd}}{Speed_{ssgd}} * 100\%;
\hspace{1em}
GR\_Rate_{dc-sgd} = \frac{Speed_{dc-sgd} - Speed_{ssgd}}{Speed_{ssgd}} * 100\%
\end{equation}

\makeatletter\def\@captype{table}\makeatother
\begin{minipage}{0.95\textwidth}
	\caption{Training speed comparisons between SSGD, OD-SGD and DC-ASGD under different clusters. The neural network and dataset are ResNet18\_V2 and CIFAR-10 separately. \textbf{Eth} is short for Ethernet while \textbf{IB} means InfiniBand network.}
	\label{tab:cifar-speed-com}
	\begin{center}
		\begin{tabular}{c|c|c|c|c|c}
			\hline
			\# of & Network  & SSGD & ASGD & DC-ASGD & OD-SGD \\
			Workers & Type & (samples/s) & (samples/s) & (samples/s) & (samples/s)\\
			\hline
			\multirow{2}*{4} & Eth & 95.785 & 157.887 & 151.011 & 149.517\\
			& \color{blue}{IB} & \color{blue}{236.78} & \color{blue}{244.19} & \color{blue}{215.03} & \color{blue}{309.782} \\
			\hline
			\multirow{2}*{8} & Eth & 74.685 & 107.35  & 98.26  & 98.76 \\ 
			& \color{blue}{IB} & \color{blue}{193.43} & \color{blue}{220.63 } & \color{blue}{149.83 } & \color{blue}{270.08 }\\
			\hline
			\multirow{2}*{12} & Eth & 57.62  & 83.00 & 74.82  & 78.37\\  
			& \color{blue}{IB} & \color{blue}{155.47 } & \color{blue}{163.40 } & \color{blue}{88.47 } & \color{blue}{196.94 }\\ 
			\hline
		\end{tabular}
	\end{center}
\end{minipage}

\subsection{Warm-up Sensitivity Analysis}

\begin{figure}[b]
	\begin{minipage}{0.42\linewidth}
		\centerline{\includegraphics[width=6.4cm, height=4cm]{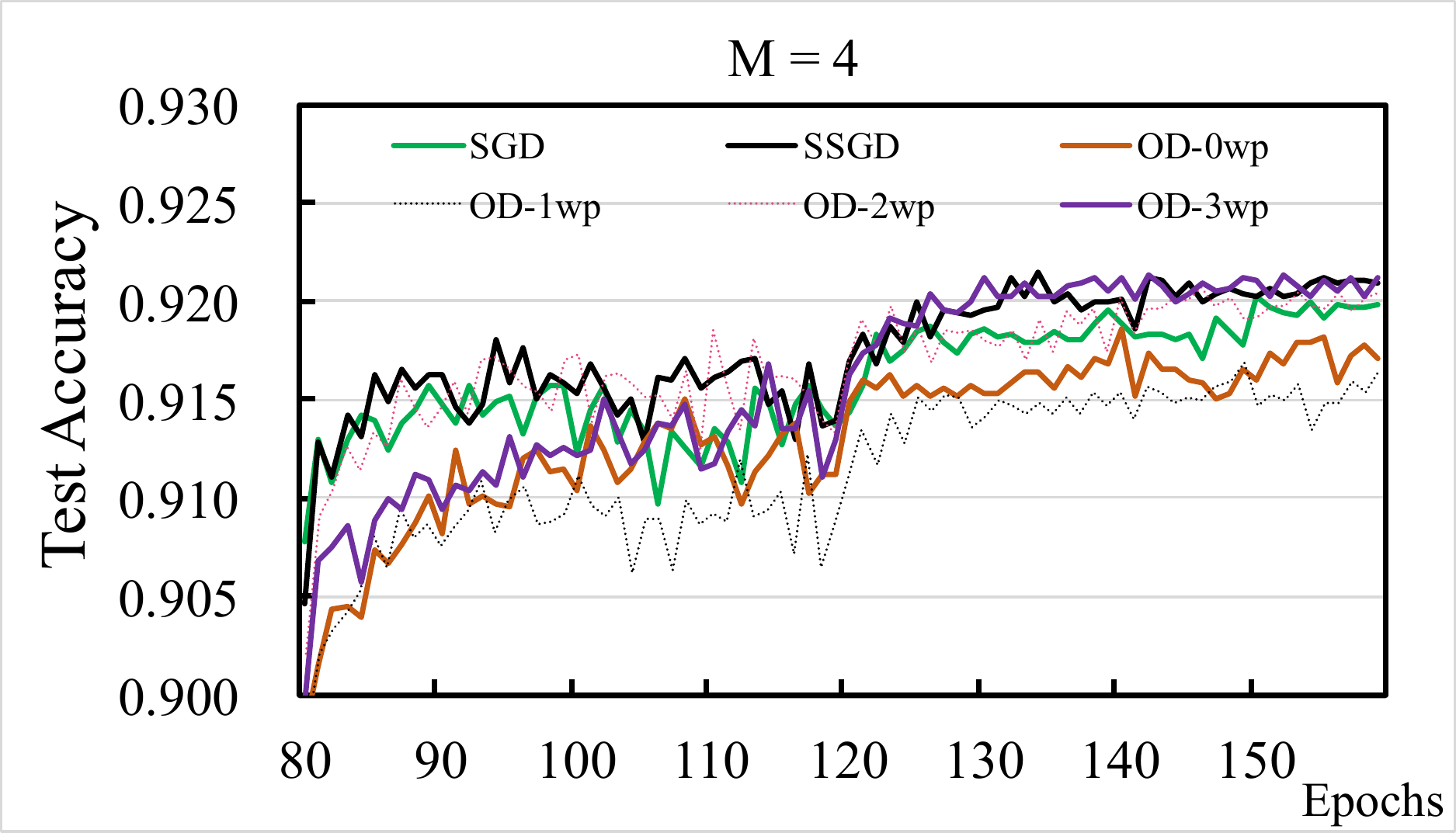}}
		\centerline{(a)}
	\end{minipage}
	\hspace{3ex}
	\begin{minipage}{0.42\linewidth}
		\centerline{\includegraphics[width=6.4cm, height=4cm]{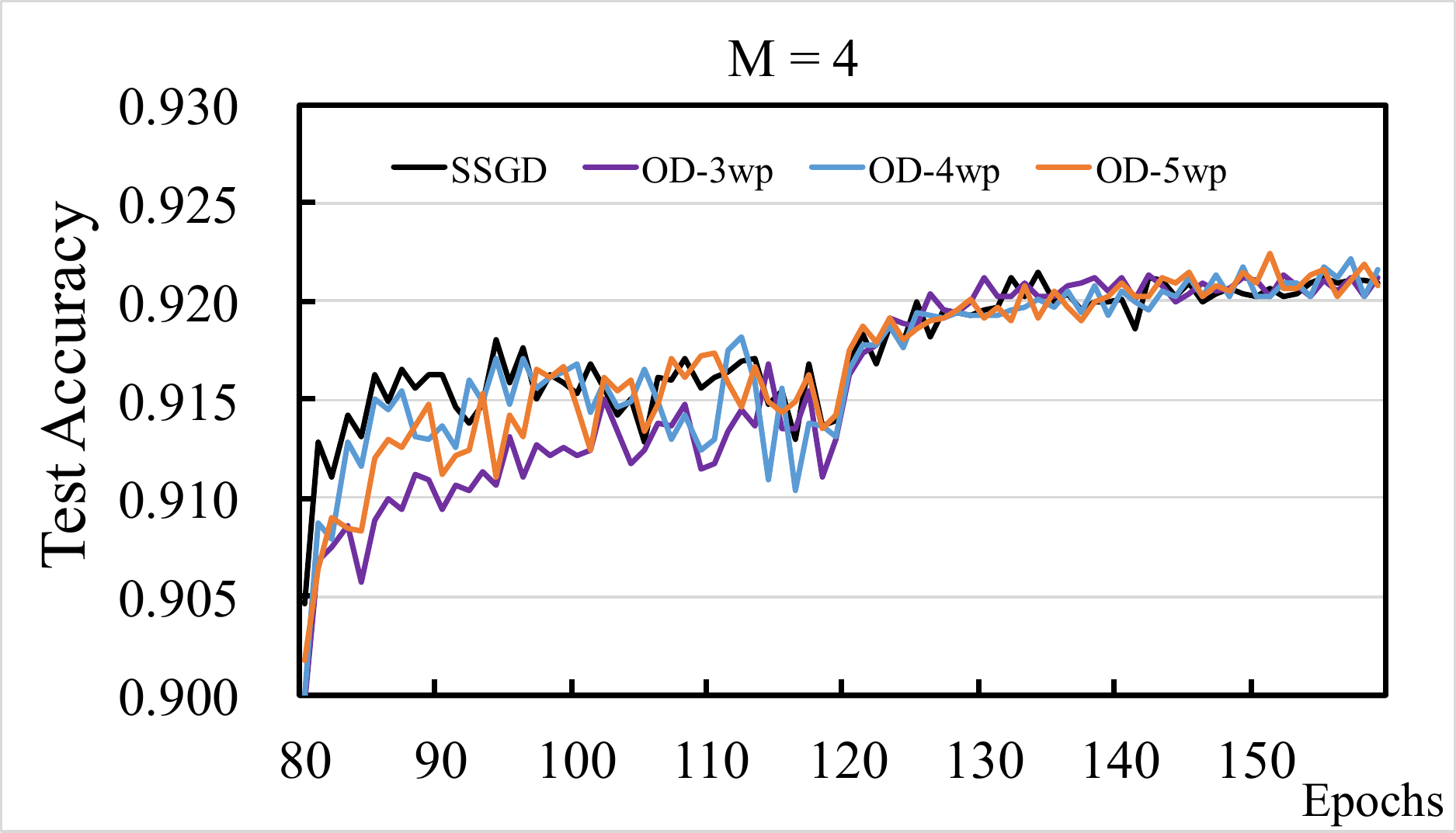}}
		\centerline{(b)}
	\end{minipage}
	\caption{Test accuracy of ResNet-20 w.r.t different warm-up epochs on CIFAR-10, convergence curves of SGD and SSGD are also included.}
	\label{fig:warmup-a}
\end{figure}

The warm-up stage introduced in OD-SGD is to make the weights more stable before the third stage, thus achieving a better convergence point. To analyze the influence from warm-up stage, we conduct some training tasks on CIFAR-10 using the ResNet-20 model, using the V100 GPU cluster with 4 worker nodes. The warm-up period for OD-SGD-c (DC-ASGD-c is used for local update operation) is set 0 epoch, 1 epoch, 2 epochs, 3 epochs, 4 epochs and 5 epochs separately. The mini-batch size for a single node is 128, learning rates of SGD and SSGD are 0.1 and 0.4, the global learning for OD-SGD-c is 0.4 while the local learning rate is 0.5, no WP strategy is used. 

Fig.~\ref{fig:warmup-a} demonstrates the test accuracy curves of SGD, SSGD and OD-SGD-c under various warm-up epochs, the test accuracy is shown in Table~\ref{tab:resnet20-epoch}. From the figure and table we have the following observations: (1) Test accuracy of SSGD is slightly better than that of sequential SGD, which is the same as training ResNet-50 with ImageNet. (2) when the warm-up stage are 0 epoch and 1 epoch, the convergence accuracy is obviously lower than that of SGD, proving that warm-up stage in OD-SGD-c is necessary. (3) When the warm-up stage is 2 epochs, test accuracy of OD-SGD-c is slightly higher than that of SGD while lower than SSGD. (4) When warm-up stage is 3 or more epochs, test accuracy of OD-SGD-c is comparable to that of SSGD (Fig.~\ref{fig:warmup-a}(b)), we can also notice OD-SGD-c even achieves about 0.05\% improvement in convergence accuracy according to Table~\ref{tab:resnet20-epoch}. 

\vspace{3ex}
\makeatletter\def\@captype{table}\makeatother
\begin{minipage}{0.9\textwidth}
	\centering
	\caption{Test accuracy on CIFAR-10 with ResNet-20 (V100 GPU cluster), different epochs are used for OD-SGD-c in the warm-up stage, \textbf{OD-1wp} means the warm-up stage is 1 epoch.}
	\label{tab:resnet20-epoch}
	\begin{center}
		\begin{tabular}{c|c|c|c|c|c|c|c|c}
			\hline
			Algorithm & SGD  &SSGD & OD-0wp & OD-1wp & OD-2wp & OD-3wp & OD-4wp & OD-5wp\\ 
			\hline
			Accuracy(\%) & 92.027 & \textbf{92.125} & 91.780 & 91.643 & \textbf{92.050} & \textbf{92.136} & \textbf{92.177}& \color{red}{92.183}\\
			\hline
		\end{tabular}
	\end{center}
\end{minipage}	
\vspace{3ex}

Based on the observations above, we conduct the training task of ResNet-50 again using the V100 GPU cluster (4 worker nodes) and ImageNet dataset, the warm-up period is set to 1 epoch. Considering the conclusion (section 5.4) that hyper-parameters impose obvious influences on the convergence accuracy, we modify the mini-batch size from 512 to 128 as the configuration ($\eta=0.1$, $\lambda=2$, $m=0$) mentioned in \cite{dcasgd-2017} is more suitable for batch size 128 (32 per GPU card). Nevertheless, the training speed is also greatly reduced as the batch size goes down. The learning rate of SSGD is 0.4 while the global learning of OD-SGD-a is 0.4 and the local learning rate is 0.1.

\begin{figure}[b]
	\centering
	\begin{minipage}{0.48\textwidth}
		\centering
		\includegraphics[width=5.6cm, height=3.5cm]{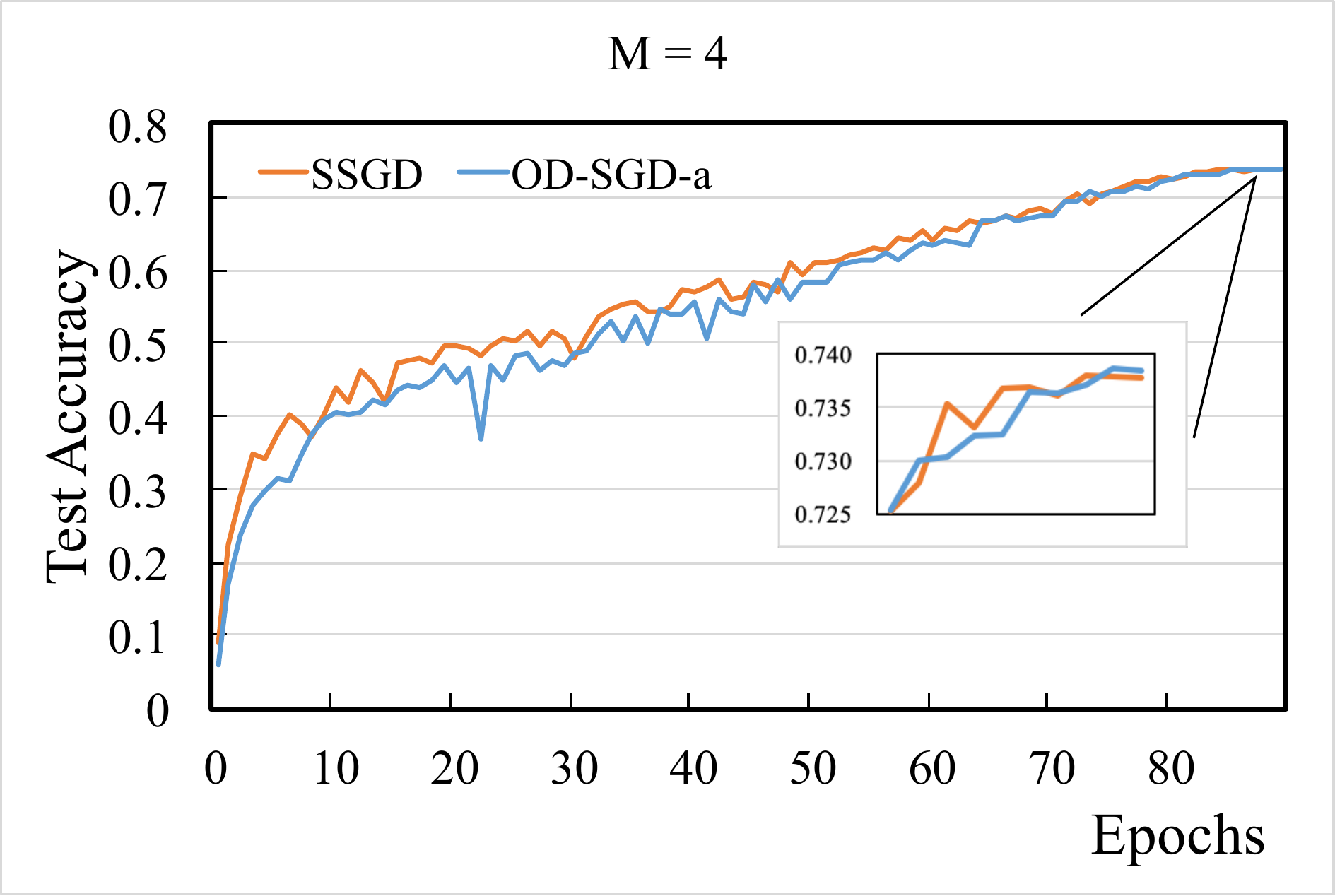}
		\caption{Test accuracy of ResNet-50 w.r.t epochs on ImageNet, the mini-batch size is 128 and the warm-up period is 1 epoch.}
		\label{fig:epoch-128} 
	\end{minipage}
	\begin{minipage}{0.48\textwidth}
		\centering
		\includegraphics[width=5.4cm, height=3.5cm]{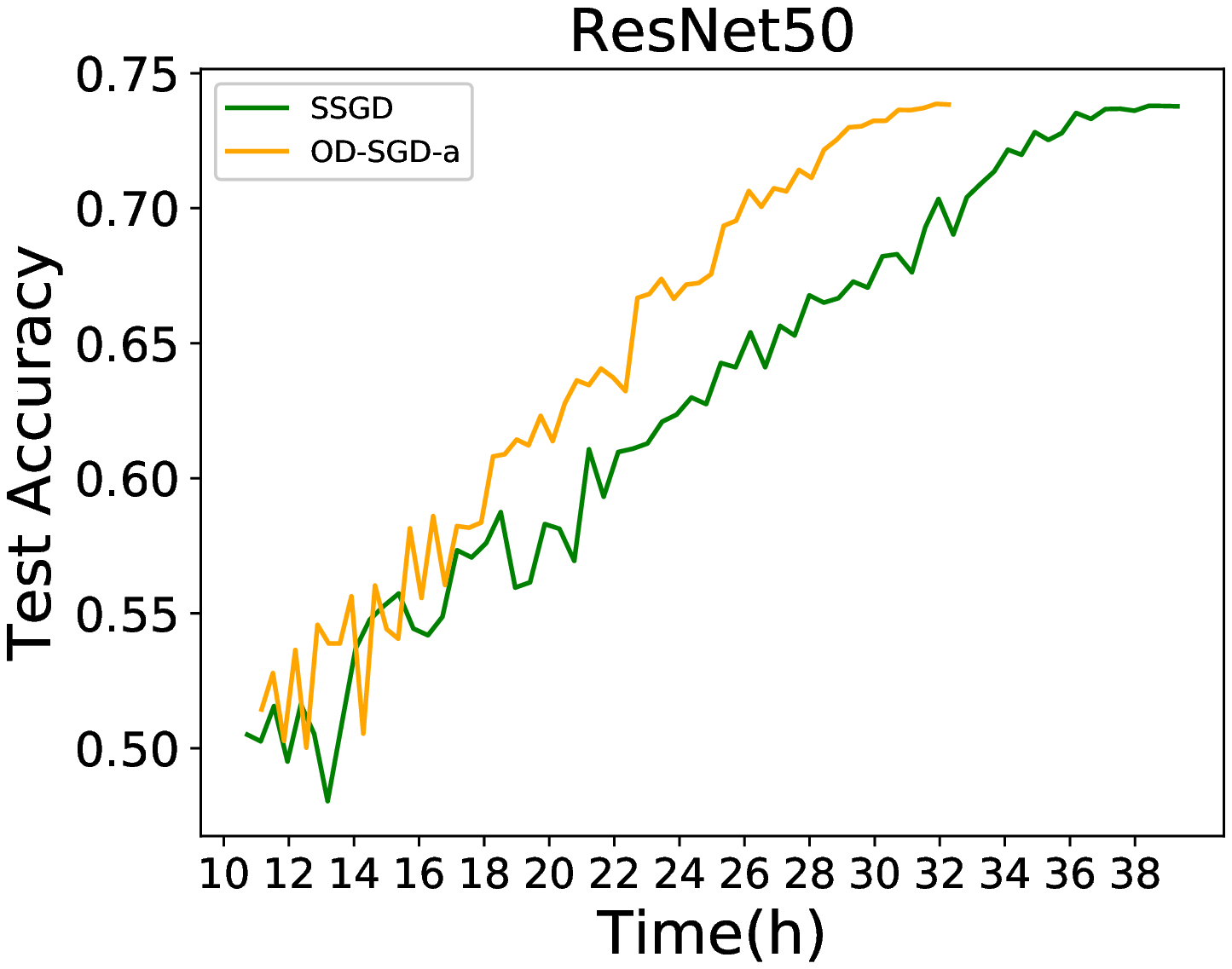}
		\caption{Test accuracy w.r.t wallclock time on ImageNet, the mini-batch size is 128.}
		\label{fig:time-128} 
	\end{minipage}
\end{figure}

Fig.~\ref{fig:epoch-128} presents the test accuracy of ResNet-50 with regard to epochs, from which we have the following observations: (1) SSGD have better convergence rate than OD-SGD-a and its test accuracy is obviously higher than that of OD-SGD-a in the first 65 epochs. (2) It is noteworthy that OD-SGD-a achieves higher convergence accuracy than SSGD after training for 90 epochs, and their test accuracy is 73.864\% and 73.786\% respectively. Fig.~\ref{fig:time-128} demonstrates the test accuracy with regard to time. Although SSGD has better convergence rate, its training speed is much slower than that of OD-SGD-a. With the same amount of training time, test accuracy of OD-SGD-a is significantly higher. It takes 39.30 hours for SSGD to complete the 90 epochs training task while 32.27 hours for OD-SGD-a, leading to 17.89\% improvement in training speed.

Based on our experiments, the warm-up stage is necessary for OD-SGD and the length of warm-up period is much more like a empirical trick, which is difficult to provide an accurate number of epochs for this stage from a theoretical perspective. The required least number of epochs in this stage differs for different neural networks.

\section{Related work}

This section describes the optimization research landscape of distributed deep learning training in recent years and we summary some works from four fields: overlapping communication and computation, reducing computation overhead, reducing the number of communication times, and reducing the communication overhead for each iteration.

\subsection{Overlapping Communication and Computation}
Several similar works like OD-SGD have been conducted previously, which improves distributed training performance via overlapping communication and computation. 
A. Agarwal et al. \cite{rebutal3-2011distributed} analyze the convergence of gradient-based optimization algorithm, and results show that for smooth stochastic problems, delays due to asynchrony are asymptotically negligible. This algorithm is based on asynchronous mechanism and different nodes suffer from various delays according to their distances from the master node. However, OD-SGD utilizes the features of ASGD and SSGD, all local workers suffer from the same level of delay and it is suitable for more than smooth problems. Delay Compensated ASGD (DC-ASGD) proposed by Shuxin Zheng et al. \cite{dcasgd-2017} is based on ASGD and the compensation operations are conducted in the master node to improve the convergence point. Different from DC-ASGD, compensation operations of OD-SGD are executed on each local worker node and there is a global synchronization in the parameter server, mechanism of OD-SGD combines the features of ASGD and SSGD. M. Assran et al. \cite{rebutal2-2018stochastic} also come up with the Stochastic Gradient Push (SGP) method for Peer-to-Peer structure based distributed training. A node will block if it does not receive messages from its in-neighbors after $\tau$	iterations. OD-SGD differs from SGP in that it applies to both Parameter Server and Peer-to-Peer structures while SGP aims at Peer-to-Peer structure.


\subsection{Reducing Computation overhead}
A series of optimizations have been conducted to lower the computation time. 
One of the critical strategy is improving the computing capacity of hardware, 
which is also the main driving force behind the momentum that deep learning 
has gained \cite{awan2017s}. The GPU series ranges from Tesla K80 to Tesla T4 \cite{cuda-gpu}, and the TPU series ranges from TPUv1 to TPUv3 \cite{tpu}, the computing power is extremely improved. 
Besides, reducing the overhead for accessing main memory also contributes to computation overhead reduction, Dadiannao \cite{chen2014dadiannao} greatly accelerates the computing speed by introducing more on-chip storage, leading to much fewer accesses to main memory, while Prime \cite{chi2016prime} takes advantage of Processing-in-memory (PIM) to reduce the memory access overhead. Low-precision computation is 
also used for lowering the time cost of deep learning, and it can be implemented in terms 
of hardware or software. For the perspective of hardware, S. Gupta et al. designs
an energy-efficient hardware accelerator to implement the low-precision (16-bit) neural network training with stochastic rounding \cite{gupta2015deep}, and the neural network accelerator named outlier-aware accelerator (OLAccel) is based on 4-bit multiply-accumulate (MAC) 
units \cite{olaccel} etc. As to the term of software, C. R. Aberger et al. \cite{de2018high} introduce a training algorithm called High-Accuracy Low-Precision (HALP) and 
Micikevicius et al. \cite{micikevicius2017mixed} describe a methodology (Mixed Precision Training, MPT) for neural network training using half-precision floating point numbers. 

\subsection{Reducing Communication Times}
Reducing communication times means a reduction in the number of iterations, and 
for fixed epochs, the way to reduce the iteration number is increasing the batch size. 
Goyal et al. \cite{goyal2017accurate} firstly adopt the warming up scheme and linear scaling rule to increase 
the batch size, and successfully train ResNet-50 ImageNet model in 1 hour.
You et al. design the LARS \cite{lars} to ensure the 
training accuracy as the batch size goes up greatly, and they further increase the 
batch size from 8K to 32K. 
Peng Sun et al. \cite{sun2019optimizing} 
increase the batch size to 64K and they complete the training of AlexNet ImageNet model in 1.5 minutes over 512 Volta GPUs. As a matter of fact, a series of research has been done to 
increase the batch size \cite{smith2017don, codreanu2017scale, akiba2017extremely, jia2018highly},  
and the most important thing in this field 
is keeping the model accuracy as the batch size goes up. 
Therefore, for any strategy to improve the training performance, we should prevent the accuracy from dropping. 


\subsection{Reducing Communication Overhead}
As the scale of cluster increases, the high communication overhead makes the 
bottleneck of distributed machine learning, especially for centralized deployment 
mode based frameworks such as MXNet \cite{chen2015mxnet} and Tensorflow \cite{abadi2016tensorflow}. 
The primary method for decreasing the communication overhead is reducing the 
communication traffic. Wei Wen et al. \cite{wen2017terngrad} utilize ternary gradients to largely 
cut down the communication workload. 
Dan Alistarh et al. \cite{alistarh2017qsgd} designs the 
Quantized SGD (QSGD), which is a family of compression schemes with 
convergence guarantees and good practical performance. Mingchao Yu 
et al. \cite{yu2018gradiveq} propose a gradient vector quantization technique with the name GradiVeQ 
by using the strong linear correlations between CNN gradients. 
In addition to these quantization strategies, Baidu \cite{baidu} proposes the ring-based all-reduce algorithm and the PowerAI Distributed Deep Learning system of IBM \cite{cho2017powerai} also mentioned a new all-reduce algorithm. 
In response to the disadvantage of the original version of ring-based all reduced algorithm, 
Horovod \cite{sergeev2018horovod} introduces the gradient fusion method to all-reduce algorithm, reducing tensor fragmentation and improving bandwidth utilization. 


\section{Conclusions}
In this article, we proposed a novel algorithm named One-Step Delay SGD (OD-SGD), which combines the features of SSGD and ASGD, to improve the distributed deep learning training performance. We have evaluated OD-SGD on MNIST, CIFAR-10 and ImageNet datasets, and compare its performance with the ASGD, SSGD and DC-ASGD. Experimental results demonstrate that OD-SGD can obtain similar or slightly better accuracy than SSGD with a much faster convergence speed, which is even faster than ASGD. When applied to large-scale neural network, OD-SGD experiences a lightly drop in convergence accuracy, which calls for further improvement. Therefore, as for future work, we plan to design a proper compensation algorithm for local update operation. Further, we intend to extend the delay from one step to several steps to achieve higher communication and computation overlap ratio for performance improvement. Our work are open source on github  (\url{https://github.com/CynthiaProtector/OD-SGD}).


\bibliographystyle{unsrt}  

\bibliography{reference}

\begin{thebibliography}{10}

\bibitem{mpcasgd}
Matthias Langer, Ashley Hall, Zhen He, and Wenny Rahayu.
\newblock Mpca sgd—a method for distributed training of deep learning models
  on spark.
\newblock {\em IEEE Transactions on Parallel and Distributed Systems},
  29(11):2540--2556, 2018.

\bibitem{GPU}
CUDA GPUs.
\newblock https://developer.nvidia.com/cuda-gpus, 2019.

\bibitem{jouppi2017datacenter}
Norman~P Jouppi, Cliff Young, Nishant Patil, David Patterson, Gaurav Agrawal,
  Raminder Bajwa, Sarah Bates, Suresh Bhatia, Nan Boden, Al~Borchers, et~al.
\newblock In-datacenter performance analysis of a tensor processing unit.
\newblock In {\em Proceedings of the 44th Annual International Symposium on
  Computer Architecture}, pages 1--12. ACM, 2017.

\bibitem{deng2009imagenet}
Jia Deng, Wei Dong, Richard Socher, Li-Jia Li, Kai Li, and Li~Fei-Fei.
\newblock Imagenet: A large-scale hierarchical image database.
\newblock In {\em Computer Vision and Pattern Recognition, 2009. CVPR 2009.
  IEEE Conference on}, pages 248--255. IEEE, 2009.

\bibitem{CIFAR-10}
The CIFAR-10 dataset.
\newblock https://www.cs.toronto.edu/~kriz/cifar.html, 2019.

\bibitem{he2016deep}
Kaiming He, Xiangyu Zhang, Shaoqing Ren, and Jian Sun.
\newblock Deep residual learning for image recognition.
\newblock In {\em Proceedings of the IEEE conference on computer vision and
  pattern recognition}, pages 770--778, 2016.

\bibitem{chilimbi2014project}
Trishul~M Chilimbi, Yutaka Suzue, Johnson Apacible, and Karthik Kalyanaraman.
\newblock Project adam: Building an efficient and scalable deep learning
  training system.
\newblock In {\em OSDI}, volume~14, pages 571--582, 2014.

\bibitem{xing2015petuum}
Eric~P Xing, Qirong Ho, Wei Dai, Jin~Kyu Kim, Jinliang Wei, Seunghak Lee, Xun
  Zheng, Pengtao Xie, Abhimanu Kumar, and Yaoliang Yu.
\newblock Petuum: A new platform for distributed machine learning on big data.
\newblock {\em IEEE Transactions on Big Data}, 1(2):49--67, 2015.

\bibitem{moritz2015sparknet}
Philipp Moritz, Robert Nishihara, Ion Stoica, and Michael~I Jordan.
\newblock Sparknet: Training deep networks in spark.
\newblock {\em arXiv preprint arXiv:1511.06051}, 2015.

\bibitem{zinkevich2010parallelized}
Martin Zinkevich, Markus Weimer, Lihong Li, and Alex~J Smola.
\newblock Parallelized stochastic gradient descent.
\newblock In {\em Advances in neural information processing systems}, pages
  2595--2603, 2010.

\bibitem{lin2017deep}
Yujun Lin, Song Han, Huizi Mao, Yu~Wang, and William~J Dally.
\newblock Deep gradient compression: Reducing the communication bandwidth for
  distributed training.
\newblock {\em arXiv preprint arXiv:1712.01887}, 2017.

\bibitem{chen2016revisiting}
Jianmin Chen, Xinghao Pan, Rajat Monga, Samy Bengio, and Rafal Jozefowicz.
\newblock Revisiting distributed synchronous sgd.
\newblock {\em arXiv preprint arXiv:1604.00981}, 2016.

\bibitem{goyal2017accurate}
Priya Goyal, Piotr Doll{\'a}r, Ross Girshick, Pieter Noordhuis, Lukasz
  Wesolowski, Aapo Kyrola, Andrew Tulloch, Yangqing Jia, and Kaiming He.
\newblock Accurate, large minibatch sgd: training imagenet in 1 hour.
\newblock {\em arXiv preprint arXiv:1706.02677}, 2017.

\bibitem{chen2015mxnet}
Tianqi Chen, Mu~Li, Yutian Li, Min Lin, Naiyan Wang, Minjie Wang, Tianjun Xiao,
  Bing Xu, Chiyuan Zhang, and Zheng Zhang.
\newblock Mxnet: A flexible and efficient machine learning library for
  heterogeneous distributed systems.
\newblock {\em arXiv preprint arXiv:1512.01274}, 2015.

\bibitem{pytorch}
PyTorch.
\newblock https://pytorch.org/features, 2019.

\bibitem{jia2014caffe}
Yangqing Jia, Evan Shelhamer, Jeff Donahue, Sergey Karayev, Jonathan Long, Ross
  Girshick, Sergio Guadarrama, and Trevor Darrell.
\newblock Caffe: Convolutional architecture for fast feature embedding.
\newblock In {\em Proceedings of the 22nd ACM international conference on
  Multimedia}, pages 675--678. ACM, 2014.

\bibitem{you2018imagenet}
Yang You, Zhao Zhang, Cho-Jui Hsieh, James Demmel, and Kurt Keutzer.
\newblock Imagenet training in minutes.
\newblock In {\em Proceedings of the 47th International Conference on Parallel
  Processing}, page~1. ACM, 2018.

\bibitem{dcasgd-2017}
Shuxin Zheng, Qi~Meng, Taifeng Wang, Wei Chen, Nenghai Yu, Zhi-Ming Ma, and
  Tie-Yan Liu.
\newblock Asynchronous stochastic gradient descent with delay compensation.
\newblock In {\em Proceedings of the 34th International Conference on Machine
  Learning-Volume 70}, pages 4120--4129. JMLR. org, 2017.

\bibitem{lecun1998mnist}
Yann LeCun.
\newblock The mnist database of handwritten digits.
\newblock {\em http://yann. lecun. com/exdb/mnist/}, 1998.

\bibitem{mobilenet}
Andrew~G Howard, Menglong Zhu, Bo~Chen, Dmitry Kalenichenko, Weijun Wang,
  Tobias Weyand, Marco Andreetto, and Hartwig Adam.
\newblock Mobilenets: Efficient convolutional neural networks for mobile vision
  applications.
\newblock {\em arXiv preprint arXiv:1704.04861}, 2017.

\bibitem{rebutal3-2011distributed}
Alekh Agarwal and John~C Duchi.
\newblock Distributed delayed stochastic optimization.
\newblock In {\em Advances in Neural Information Processing Systems}, pages
  873--881, 2011.

\bibitem{rebutal2-2018stochastic}
Mahmoud Assran, Nicolas Loizou, Nicolas Ballas, and Michael Rabbat.
\newblock Stochastic gradient push for distributed deep learning.
\newblock {\em arXiv preprint arXiv:1811.10792}, 2018.

\bibitem{awan2017s}
Ammar~Ahmad Awan, Khaled Hamidouche, Jahanzeb~Maqbool Hashmi, and Dhabaleswar~K
  Panda.
\newblock S-caffe: Co-designing mpi runtimes and caffe for scalable deep
  learning on modern gpu clusters.
\newblock In {\em Proceedings of the 22nd ACM SIGPLAN Symposium on Principles
  and Practice of Parallel Programming}, pages 193--205. ACM, 2017.

\bibitem{cuda-gpu}
Cuda GPUs.
\newblock https://developer.nvidia.com/cuda-gpus, 2019.

\bibitem{tpu}
TPU.
\newblock
  https://www.nextplatform.com/2018/05/10/tearing-apart-googles-tpu-3-0-ai-coprocessor/,
  2019.

\bibitem{chen2014dadiannao}
Yunji Chen, Tao Luo, Shaoli Liu, Shijin Zhang, Liqiang He, Jia Wang, Ling Li,
  Tianshi Chen, Zhiwei Xu, Ninghui Sun, et~al.
\newblock Dadiannao: A machine-learning supercomputer.
\newblock In {\em Proceedings of the 47th Annual IEEE/ACM International
  Symposium on Microarchitecture}, pages 609--622. IEEE Computer Society, 2014.

\bibitem{chi2016prime}
Ping Chi, Shuangchen Li, Cong Xu, Tao Zhang, Jishen Zhao, Yongpan Liu, Yu~Wang,
  and Yuan Xie.
\newblock Prime: A novel processing-in-memory architecture for neural network
  computation in reram-based main memory.
\newblock In {\em ACM SIGARCH Computer Architecture News}, volume~44, pages
  27--39. IEEE Press, 2016.

\bibitem{gupta2015deep}
Suyog Gupta, Ankur Agrawal, Kailash Gopalakrishnan, and Pritish Narayanan.
\newblock Deep learning with limited numerical precision.
\newblock In {\em International Conference on Machine Learning}, pages
  1737--1746, 2015.

\bibitem{olaccel}
Eunhyeok Park, Dongyoung Kim, and Sungjoo Yoo.
\newblock Energy-efficient neural network accelerator based on outlier-aware
  low-precision computation.
\newblock In {\em 2018 ACM/IEEE 45th Annual International Symposium on Computer
  Architecture (ISCA)}, pages 688--698. IEEE, 2018.

\bibitem{de2018high}
Christopher De~Sa, Megan Leszczynski, Jian Zhang, Alana Marzoev, Christopher~R
  Aberger, Kunle Olukotun, and Christopher R{\'e}.
\newblock High-accuracy low-precision training.
\newblock {\em arXiv preprint arXiv:1803.03383}, 2018.

\bibitem{micikevicius2017mixed}
Paulius Micikevicius, Sharan Narang, Jonah Alben, Gregory Diamos, Erich Elsen,
  David Garcia, Boris Ginsburg, Michael Houston, Oleksii Kuchaiev, Ganesh
  Venkatesh, et~al.
\newblock Mixed precision training.
\newblock {\em arXiv preprint arXiv:1710.03740}, 2017.

\bibitem{lars}
Yang You, Igor Gitman, and Boris Ginsburg.
\newblock Large batch training of convolutional networks.
\newblock {\em arXiv preprint arXiv:1708.03888}, 2017.

\bibitem{sun2019optimizing}
Peng Sun, Wansen Feng, Ruobing Han, Shengen Yan, and Yonggang Wen.
\newblock Optimizing network performance for distributed dnn training on gpu
  clusters: Imagenet/alexnet training in 1.5 minutes.
\newblock {\em arXiv preprint arXiv:1902.06855}, 2019.

\bibitem{smith2017don}
Samuel~L Smith, Pieter-Jan Kindermans, Chris Ying, and Quoc~V Le.
\newblock Don't decay the learning rate, increase the batch size.
\newblock {\em arXiv preprint arXiv:1711.00489}, 2017.

\bibitem{codreanu2017scale}
Valeriu Codreanu, Damian Podareanu, and Vikram Saletore.
\newblock Scale out for large minibatch sgd: Residual network training on
  imagenet-1k with improved accuracy and reduced time to train.
\newblock {\em arXiv preprint arXiv:1711.04291}, 2017.

\bibitem{akiba2017extremely}
Takuya Akiba, Shuji Suzuki, and Keisuke Fukuda.
\newblock Extremely large minibatch sgd: training resnet-50 on imagenet in 15
  minutes.
\newblock {\em arXiv preprint arXiv:1711.04325}, 2017.

\bibitem{jia2018highly}
Xianyan Jia, Shutao Song, Wei He, Yangzihao Wang, Haidong Rong, Feihu Zhou,
  Liqiang Xie, Zhenyu Guo, Yuanzhou Yang, Liwei Yu, et~al.
\newblock Highly scalable deep learning training system with mixed-precision:
  Training imagenet in four minutes.
\newblock {\em arXiv preprint arXiv:1807.11205}, 2018.

\bibitem{abadi2016tensorflow}
Mart{\'\i}n Abadi, Ashish Agarwal, Paul Barham, Eugene Brevdo, Zhifeng Chen,
  Craig Citro, Greg~S Corrado, Andy Davis, Jeffrey Dean, Matthieu Devin, et~al.
\newblock Tensorflow: Large-scale machine learning on heterogeneous distributed
  systems.
\newblock {\em arXiv preprint arXiv:1603.04467}, 2016.

\bibitem{wen2017terngrad}
Wei Wen, Cong Xu, Feng Yan, Chunpeng Wu, Yandan Wang, Yiran Chen, and Hai Li.
\newblock Terngrad: Ternary gradients to reduce communication in distributed
  deep learning.
\newblock In {\em Advances in neural information processing systems}, pages
  1509--1519, 2017.

\bibitem{alistarh2017qsgd}
Dan Alistarh, Demjan Grubic, Jerry Li, Ryota Tomioka, and Milan Vojnovic.
\newblock Qsgd: Communication-efficient sgd via gradient quantization and
  encoding.
\newblock In {\em Advances in Neural Information Processing Systems}, pages
  1709--1720, 2017.

\bibitem{yu2018gradiveq}
Mingchao Yu, Zhifeng Lin, Krishna Narra, Songze Li, Youjie Li, Nam~Sung Kim,
  Alexander Schwing, Murali Annavaram, and Salman Avestimehr.
\newblock Gradiveq: Vector quantization for bandwidth-efficient gradient
  aggregation in distributed cnn training.
\newblock In {\em Advances in Neural Information Processing Systems}, pages
  5123--5133, 2018.

\bibitem{baidu}
Bringing HPC~Techniques to~Deep~Learning.
\newblock http://andrew.gibiansky.com, 2017.

\bibitem{cho2017powerai}
Minsik Cho, Ulrich Finkler, Sameer Kumar, David Kung, Vaibhav Saxena, and
  Dheeraj Sreedhar.
\newblock Powerai ddl.
\newblock {\em arXiv preprint arXiv:1708.02188}, 2017.

\bibitem{sergeev2018horovod}
Alexander Sergeev and Mike Del~Balso.
\newblock Horovod: fast and easy distributed deep learning in tensorflow.
\newblock {\em arXiv preprint arXiv:1802.05799}, 2018.

\end{thebibliography}
\end{document}